\documentclass[11pt]{article}

\usepackage[final]{acl}

\usepackage{times}
\usepackage{latexsym}

\usepackage[T1]{fontenc}
\usepackage{amssymb}
\usepackage{amsmath}

\usepackage[utf8]{inputenc}

\usepackage{microtype}

\usepackage{inconsolata}

\usepackage{graphicx}
\usepackage{enumitem}
\usepackage{todonotes}
\usetikzlibrary{tikzmark}
\usepackage{array}
\usepackage{booktabs}
\usepackage{colortbl}
\usepackage{arydshln}
\usepackage{placeins}

\definecolor{tticblue}{RGB}{0, 94, 184}
\definecolor{placeholderctx}{RGB}{31, 91, 154}
\definecolor{realctx}{RGB}{184, 134, 11}
\definecolor{heatlow}{RGB}{142, 85, 160}
\definecolor{heathigh}{RGB}{61, 145, 93}
\newlength{\ctxcolwidth}
\newcolumntype{P}{>{\raggedleft\arraybackslash}m{\ctxcolwidth}}
\newcolumntype{R}{>{\raggedleft\arraybackslash}m{\ctxcolwidth}}
\newcommand{\placeholderhead}[1]{\textcolor{placeholderctx}{\emph{#1}}}
\newcommand{\realhead}[1]{\textcolor{realctx}{\emph{#1}}}
\newcommand{\placeholderctxtext}[1]{\textcolor{placeholderctx}{#1}}
\newcommand{\realctxtext}[1]{\textcolor{realctx}{#1}}
\newcommand{\rankheatcell}[2]{%
  \ifnum#1<45
    \cellcolor{heatlow!\the\numexpr(45-#1)*2\relax}{#2}%
  \else\ifnum#1>45
    \cellcolor{heathigh!\the\numexpr(#1-45)*2\relax}{#2}%
  \else
    \cellcolor{white}{#2}%
  \fi\fi}
\newcommand{\rmsecell}[2]{\rankheatcell{#1}{#2}}
\newcommand{\corrcell}[2]{\rankheatcell{#1}{#2}}
\newcommand{\shiftcell}[2]{\rankheatcell{#1}{#2}}
\newcommand{\ctxmark}[2]{\tikzmarknode[minimum width=\ctxcolwidth,text width=\ctxcolwidth,align=right,inner xsep=0pt,inner ysep=1pt,outer sep=0pt]{#1}{#2}}
\newcommand{\ctxoutline}[3]{%
  \tikz[remember picture,overlay]
  \draw[#1,line width=0.8pt,rounded corners=1pt]
    ([xshift=-0.5pt,yshift=1.5pt]#2.north west)
    rectangle
    ([xshift=0.5pt,yshift=-1.5pt]#3.south east);%
}

\title{Real Images, Worse Judgments: Evaluating Vision-Language Models on Concreteness and Imagery}

\author{
  \textbf{Yifan Jiang} \quad
  \textbf{Ruoxi Ning} \quad
  \textbf{Sheng Yao} \quad
  \textbf{Freda Shi} \\
  University of Waterloo \quad Vector Institute \\
  \texttt{\{yifan.jiang,ruoxi.ning,s57yao,fhs\}@uwaterloo.ca}
}

\begin{document}
\maketitle
\begin{abstract}
	Visual inputs are often assumed to improve language understanding in multimodal models.
	We examine this assumption by asking whether vision--language models (VLMs) can distinguish useful visual evidence from incidental image context in lexical judgments.
	We use human concreteness and imagery ratings because they span words with varying expected visual relevance, from abstract and low-imagery words to concrete and high-imagery words.
	We find that real-image contexts do not yield consistent gains and often hurt alignment with human ratings, most sharply when visual evidence is least relevant.
	Through probing and canonical correlation analysis, complemented by an attribution case study, we find that real-image contexts are associated with representational shifts and greater sensitivity to spurious visual cues, coinciding with weaker recoverability of the targeted lexical properties.
	We further show that instructing models to focus solely on textual content at inference time can reduce this degradation, with the clearest gains on these vulnerable subsets.
	Our findings suggest that current instruction-tuned VLMs need better calibration of when visual context should inform lexical judgments.
\end{abstract}

\begin{figure*}[t]
	\centering
	\includegraphics[width=\textwidth]{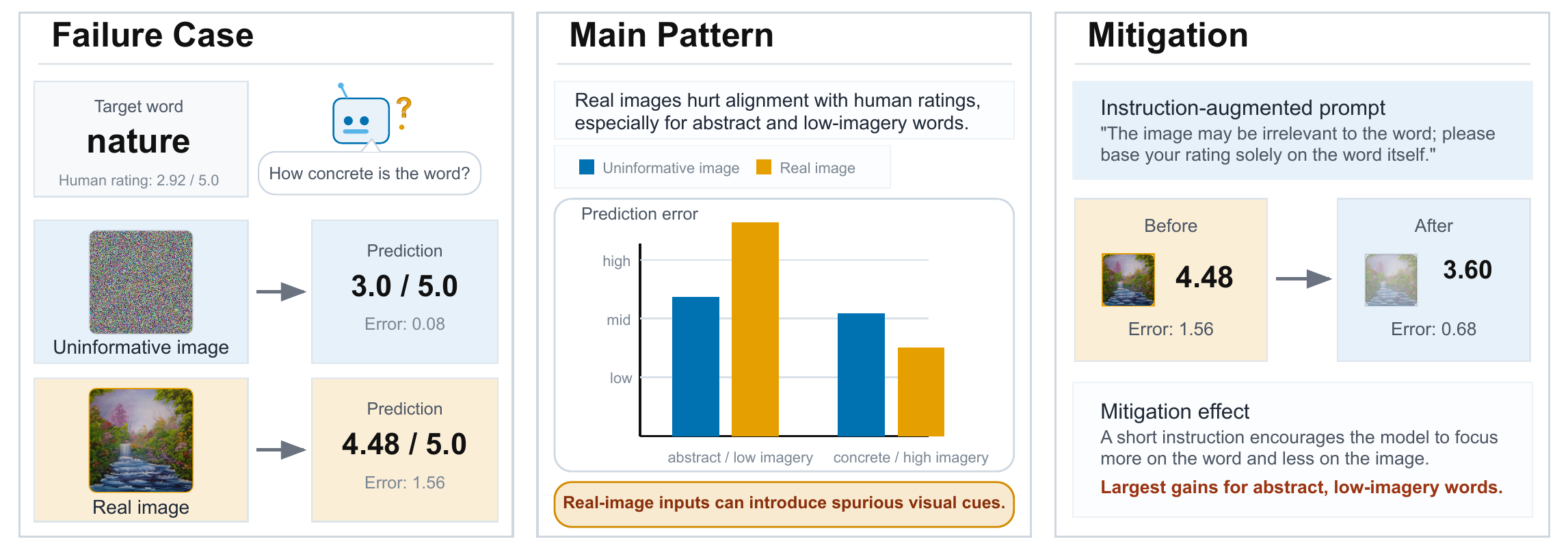}
	\caption{Real-image contexts can hurt alignment with human concreteness and imagery judgments, especially for abstract and low-imagery words. This pattern is associated with greater sensitivity to spurious visual cues, and instruction-augmented prompting reduces some errors by encouraging the model to base its rating on the word itself.}
	\label{fig:teaser}
\end{figure*}

\section{Introduction}

Visual grounding, the task of linking linguistic elements to visual representations, has gained renewed attention with the rise of multimodal models \citep{Radford2021LearningTV,Alayrac2022FlamingoAV,Li2023BLIP2BL,Liu2023VisualIT}.
By incorporating visual input, these models are often expected to enrich linguistic understanding with concrete sensory context, and they have shown strong performance on tasks such as image captioning \citep{Li2023BLIP2BL}, visual question answering \citep{Alayrac2022FlamingoAV}, and multimodal reasoning \citep{Liu2023VisualIT}.

However, current multimodal models still face challenges in using visual context in ways that align with human interpretation.
Prior work shows that models can over-rely on superficial visual--textual correlations \citep{Goyal2016MakingTV,Agrawal2017DontJA,Yuksekgonul2022WhenAW}, leading, for example, to object hallucinations associated with frequently mentioned or co-occurring objects \citep{Li2023EvaluatingOH}.
These failure modes suggest that strong downstream performance does not necessarily imply robust grounding, and they raise broader concerns about when models use visual context as meaningful evidence rather than as a shortcut, especially when the image is incidental.
For example, a multimodal vocabulary-learning system may retain an illustration from an earlier turn when it is later asked to identify words that are abstract or difficult to visualize and may need additional explanation.
If the incidental image makes those words appear easier to ground visually, the system could overlook them.

These concerns motivate a more targeted question: when asked to judge a lexical property of a word, can a vision--language model (VLM) distinguish useful visual evidence from incidental image context?
Concreteness and imagery ratings are well suited to this question because they reflect human judgments of general lexical properties: how much a word denotes something concrete, and how easily it evokes a mental image.
Although these properties are perceptual in nature, the requested rating concerns the word itself rather than any particular picture.
They therefore provide a controlled setting for testing whether VLMs preserve a lexical judgment when visual context is present.

We use these ratings to test whether state-of-the-art VLMs preserve the requested word-level judgment when an image is present, including when the image is incidental to that judgment, or instead appear to treat retrieved images as direct evidence for a property of the word itself.
We construct word--image pairs by varying the visual input while holding the target word and rating task fixed, comparing no-image input, uninformative placeholders, and \emph{real images}, defined here as retrieved natural-image inputs.

In summary, our contributions are as follows:
\begin{itemize}[leftmargin=*,itemsep=0pt]
	\item We conduct a comprehensive evaluation of instruction-tuned text-only and vision–language models on human judgments of concreteness and imagery across multiple visual contexts.
	\item We show that real-image contexts can hurt alignment with human ratings, especially for abstract and low-imagery words, and validate this contrast with a matched human study showing no comparable human degradation under the same retrieved-image condition.
	\item We demonstrate a simple inference-time prompting intervention that reduces some of the degradation associated with real-image contexts.
\end{itemize}
Code and data are publicly available.\footnote{\url{https://github.com/compling-wat/vlm-concreteness-imagery}}

\section{Related Work}

\paragraph{Concreteness and imagery.}
Concreteness and imagery are well-established psycholinguistic properties that quantify the degree to which words are tied to perceptual experience.
Concreteness measures how directly a word denotes something perceptible, whereas imagery, often discussed as imageability, captures the ease with which a word evokes a mental image \citep{Paivio1968ConcretenessIA,Friendly1982TheTW,Turney2011LiteralAM,Brysbaert2014ConcretenessRF}.
The two properties are strongly correlated and often analyzed together, though they are not identical. The fine-grained distinction is not central to our diagnostic use of them.
In empirical research, they are typically operationalized through large-scale human rating norms.

Several resources have been developed to measure human judgments of concreteness and imagery.
Standard resources include MT40k \citep{Brysbaert2014ConcretenessRF}, PYM \citep{Paivio1968ConcretenessIA} and its extension \citep{Clark2004ExtensionsOT}, and the Toronto Word Pool \citep{Friendly1982TheTW}, which have become widely used benchmarks for studying the relationship between language and perception.

Concreteness and imagery have also been employed as diagnostic signals in AI grounding research.
Concreteness has served as a learned grounding signal in visually grounded syntax induction \citep{Shi2019VisuallyGN}.
Concreteness prediction has also been proposed as a test of visual language understanding, reflecting the extent to which models encode perceptually grounded aspects of meaning \citep{Alper2023IsBB}.
Human ratings of these properties therefore offer principled targets for studying perceptually grounded aspects of meaning.
Motivated by prior work, we use these ratings to probe how modern multimodal models handle lexical judgments under visual context.

\paragraph{Visual grounding in multimodal models.}

VLMs learn joint visual–textual representations through large-scale pretraining, ranging from early contrastive models such as CLIP \citep{Radford2021LearningTV} to more recent instruction-tuned models like LLaVA \citep{Liu2023VisualIT}.
A broad line of work suggests that perceptual grounding can improve semantic representations of both concrete and abstract concepts \citep{Hill2014LearningAC,Zhuang2023VisualGH,Zhuang2024LexiconLevelCV}.
However, recent work complicates this picture: text-only representations can outperform multimodal ones in capturing experiential semantic information and aligning with human fMRI responses \citep{BavarescoFernandez2025Experiential}, and ungrounded LLMs align better with human representations for non-sensorimotor than sensorimotor features of concepts \citep{Xu2025LargeLanguageModels}.
Complementary work finds that VLMs are distracted by irrelevant images in short and long contexts \citep{Sharma2024LosingVisualNeedles} and that their predictions are sensitive to conflicts between visual evidence and textual or world-knowledge evidence \citep{Golovanevsky2025PixelsPriors,Zhang2026ImagesWords}.

Building on this line of work, we examine the relationship between visual grounding and human judgments of concreteness and imagery, with a focus on modern instruction-tuned VLMs.
We ask how different visual contexts affect VLM judgments of concreteness and imagery, and when they help or harm alignment with human ratings.

\begin{table*}[!t]
\centering
\footnotesize
\begin{minipage}{0.48\textwidth}
\centering
\setlength{\tabcolsep}{3pt}
\resizebox{\linewidth}{!}{
\begin{tabular}{l r P P R R R}
\hline
& \emph{None} & \placeholderhead{White} & \placeholderhead{Noise} & \placeholderhead{Shuffled} & \realhead{ImageNet} & \realhead{Wikimedia} \\
\hline
\multicolumn{7}{c}{\textbf{All}} \\
\hline
LLaMA 2 & \rmsecell{85}{\underline{\textbf{1.152}}} & -- & -- & -- & -- & -- \\
Mistral & \rmsecell{85}{\underline{\textbf{0.803}}} & -- & -- & -- & -- & -- \\
Qwen2.5 & \rmsecell{85}{\underline{\textbf{0.745}}} & -- & -- & -- & -- & -- \\
Gemma 3 & \rmsecell{85}{\textbf{0.745}} & \ctxmark{tab-model-predictions-mt-all-p-t}{\rmsecell{37}{\textsuperscript{***}0.933}} & \rmsecell{21}{\textsuperscript{***}1.187} & \rmsecell{5}{\textsuperscript{***}\underline{1.511}} & \ctxmark{tab-model-predictions-mt-all-r-t}{\rmsecell{69}{\textsuperscript{***}0.787}} & \rmsecell{53}{\textsuperscript{***}0.790} \\
InternVL3 & \rmsecell{53}{0.883} & \rmsecell{37}{\textsuperscript{***}0.996} & \rmsecell{5}{\textsuperscript{***}\underline{1.700}} & \rmsecell{21}{\textsuperscript{***}1.504} & \rmsecell{85}{\textsuperscript{***}\textbf{0.799}} & \rmsecell{69}{\textsuperscript{***}0.813} \\
LLaVA 1.5 & \rmsecell{5}{\underline{1.377}} & \rmsecell{85}{\textsuperscript{***}\textbf{1.356}} & \rmsecell{21}{\textsuperscript{***}1.358} & \rmsecell{69}{\textsuperscript{***}1.357} & \rmsecell{37}{\textsuperscript{***}1.357} & \rmsecell{53}{\textsuperscript{***}1.357} \\
Pixtral & \rmsecell{21}{1.138} & \rmsecell{85}{\textsuperscript{***}\textbf{0.925}} & \rmsecell{37}{\textsuperscript{**}1.105} & \rmsecell{5}{\textsuperscript{***}\underline{1.444}} & \rmsecell{53}{\textsuperscript{***}0.997} & \rmsecell{69}{\textsuperscript{***}0.961} \\
Qwen2.5-VL & \rmsecell{53}{0.984} & \rmsecell{21}{\textsuperscript{***}1.085} & \rmsecell{37}{\textsuperscript{***}1.022} & \ctxmark{tab-model-predictions-mt-all-p-b}{\rmsecell{5}{\textsuperscript{***}\underline{1.178}}} & \rmsecell{69}{\textsuperscript{***}0.943} & \ctxmark{tab-model-predictions-mt-all-r-b}{\rmsecell{85}{\textsuperscript{***}\textbf{0.931}}} \\
\noalign{\ctxoutline{placeholderctx}{tab-model-predictions-mt-all-p-t}{tab-model-predictions-mt-all-p-b}\ctxoutline{realctx}{tab-model-predictions-mt-all-r-t}{tab-model-predictions-mt-all-r-b}}
\hline
\multicolumn{7}{c}{\textbf{Abstract}} \\
\hline
LLaMA 2 & \rmsecell{85}{\underline{\textbf{1.506}}} & -- & -- & -- & -- & -- \\
Mistral & \rmsecell{85}{\underline{\textbf{0.939}}} & -- & -- & -- & -- & -- \\
Qwen2.5 & \rmsecell{85}{\underline{\textbf{0.715}}} & -- & -- & -- & -- & -- \\
Gemma 3 & \rmsecell{37}{0.730} & \ctxmark{tab-model-predictions-mt-lo-p-t}{\rmsecell{85}{\textsuperscript{***}\textbf{0.510}}} & \rmsecell{69}{\textsuperscript{***}0.607} & \rmsecell{53}{\textsuperscript{***}0.617} & \ctxmark{tab-model-predictions-mt-lo-r-t}{\rmsecell{5}{\textsuperscript{***}\underline{0.951}}} & \rmsecell{21}{\textsuperscript{***}0.932} \\
InternVL3 & \rmsecell{37}{0.600} & \rmsecell{69}{\textsuperscript{***}0.515} & \rmsecell{85}{\textsuperscript{***}\textbf{0.507}} & \rmsecell{53}{0.596} & \rmsecell{5}{\textsuperscript{***}\underline{0.837}} & \rmsecell{21}{\textsuperscript{***}0.813} \\
LLaVA 1.5 & \rmsecell{5}{\underline{1.908}} & \rmsecell{85}{\textsuperscript{***}\textbf{1.865}} & \rmsecell{21}{\textsuperscript{***}1.867} & \rmsecell{69}{\textsuperscript{***}1.866} & \rmsecell{37}{\textsuperscript{***}1.867} & \rmsecell{53}{\textsuperscript{***}1.867} \\
Pixtral & \rmsecell{5}{\underline{1.560}} & \rmsecell{85}{\textsuperscript{***}\textbf{0.696}} & \rmsecell{21}{\textsuperscript{***}1.361} & \rmsecell{69}{\textsuperscript{***}0.826} & \rmsecell{37}{\textsuperscript{***}1.303} & \rmsecell{53}{\textsuperscript{***}1.253} \\
Qwen2.5-VL & \rmsecell{85}{\textbf{0.765}} & \rmsecell{53}{\textsuperscript{***}0.803} & \rmsecell{69}{0.782} & \ctxmark{tab-model-predictions-mt-lo-p-b}{\rmsecell{37}{\textsuperscript{***}0.924}} & \rmsecell{5}{\textsuperscript{***}\underline{1.069}} & \ctxmark{tab-model-predictions-mt-lo-r-b}{\rmsecell{21}{\textsuperscript{***}1.014}} \\
\noalign{\ctxoutline{placeholderctx}{tab-model-predictions-mt-lo-p-t}{tab-model-predictions-mt-lo-p-b}\ctxoutline{realctx}{tab-model-predictions-mt-lo-r-t}{tab-model-predictions-mt-lo-r-b}}
\hline
\multicolumn{7}{c}{\textbf{Concrete}} \\
\hline
LLaMA 2 & \rmsecell{85}{\underline{\textbf{0.728}}} & -- & -- & -- & -- & -- \\
Mistral & \rmsecell{85}{\underline{\textbf{0.667}}} & -- & -- & -- & -- & -- \\
Qwen2.5 & \rmsecell{85}{\underline{\textbf{0.769}}} & -- & -- & -- & -- & -- \\
Gemma 3 & \rmsecell{53}{0.758} & \ctxmark{tab-model-predictions-mt-hi-p-t}{\rmsecell{37}{\textsuperscript{***}1.178}} & \rmsecell{21}{\textsuperscript{***}1.513} & \rmsecell{5}{\textsuperscript{***}\underline{1.973}} & \ctxmark{tab-model-predictions-mt-hi-r-t}{\rmsecell{85}{\textsuperscript{***}\textbf{0.616}}} & \rmsecell{69}{\textsuperscript{***}0.647} \\
InternVL3 & \rmsecell{53}{1.066} & \rmsecell{37}{\textsuperscript{***}1.267} & \rmsecell{5}{\textsuperscript{***}\underline{2.262}} & \rmsecell{21}{\textsuperscript{***}1.968} & \rmsecell{85}{\textsuperscript{***}\textbf{0.766}} & \rmsecell{69}{\textsuperscript{***}0.813} \\
LLaVA 1.5 & \rmsecell{85}{\textbf{0.649}} & \rmsecell{37}{\textsuperscript{***}0.672} & \rmsecell{5}{\textsuperscript{***}\underline{0.674}} & \rmsecell{21}{\textsuperscript{***}0.674} & \rmsecell{53}{\textsuperscript{***}0.671} & \rmsecell{69}{\textsuperscript{***}0.671} \\
Pixtral & \rmsecell{85}{\textbf{0.578}} & \rmsecell{21}{\textsuperscript{***}1.082} & \rmsecell{37}{\textsuperscript{***}0.830} & \rmsecell{5}{\textsuperscript{***}\underline{1.809}} & \rmsecell{53}{\textsuperscript{***}0.630} & \rmsecell{69}{\textsuperscript{*}0.612} \\
Qwen2.5-VL & \rmsecell{53}{1.136} & \rmsecell{21}{\textsuperscript{***}1.277} & \rmsecell{37}{\textsuperscript{***}1.188} & \ctxmark{tab-model-predictions-mt-hi-p-b}{\rmsecell{5}{\textsuperscript{***}\underline{1.375}}} & \rmsecell{85}{\textsuperscript{***}\textbf{0.821}} & \ctxmark{tab-model-predictions-mt-hi-r-b}{\rmsecell{69}{\textsuperscript{***}0.855}} \\
\noalign{\ctxoutline{placeholderctx}{tab-model-predictions-mt-hi-p-t}{tab-model-predictions-mt-hi-p-b}\ctxoutline{realctx}{tab-model-predictions-mt-hi-r-t}{tab-model-predictions-mt-hi-r-b}}
\hline
\end{tabular}
}
\caption*{MT40k (Concreteness)}
\end{minipage}
\hfill
\begin{minipage}{0.48\textwidth}
\centering
\setlength{\tabcolsep}{3pt}
\resizebox{\linewidth}{!}{
\begin{tabular}{l r P P R R R}
\hline
& \emph{None} & \placeholderhead{White} & \placeholderhead{Noise} & \placeholderhead{Shuffled} & \realhead{ImageNet} & \realhead{Wikimedia} \\
\hline
\multicolumn{7}{c}{\textbf{All}} \\
\hline
LLaMA 2 & \rmsecell{85}{\underline{\textbf{1.568}}} & -- & -- & -- & -- & -- \\
Mistral & \rmsecell{85}{\underline{\textbf{1.392}}} & -- & -- & -- & -- & -- \\
Qwen2.5 & \rmsecell{85}{\underline{\textbf{1.275}}} & -- & -- & -- & -- & -- \\
Gemma 3 & \rmsecell{85}{\textbf{1.396}} & \ctxmark{tab-model-predictions-cp-all-p-t}{\rmsecell{69}{\textsuperscript{*}1.420}} & \rmsecell{37}{\textsuperscript{***}1.925} & \rmsecell{53}{\textsuperscript{***}1.805} & \ctxmark{tab-model-predictions-cp-all-r-t}{\rmsecell{5}{\textsuperscript{***}\underline{2.046}}} & \rmsecell{21}{\textsuperscript{***}1.973} \\
InternVL3 & \rmsecell{69}{1.289} & \rmsecell{85}{\textsuperscript{***}\textbf{1.234}} & \rmsecell{5}{\textsuperscript{***}\underline{2.100}} & \rmsecell{21}{\textsuperscript{***}1.696} & \rmsecell{37}{\textsuperscript{***}1.464} & \rmsecell{53}{\textsuperscript{***}1.417} \\
LLaVA 1.5 & \rmsecell{85}{\textbf{1.643}} & \rmsecell{69}{\textsuperscript{***}1.941} & \rmsecell{37}{\textsuperscript{***}2.247} & \rmsecell{53}{\textsuperscript{***}2.192} & \rmsecell{5}{\textsuperscript{***}\underline{2.267}} & \rmsecell{21}{\textsuperscript{***}2.264} \\
Pixtral & \rmsecell{37}{2.009} & \rmsecell{85}{\textsuperscript{***}\textbf{1.403}} & \rmsecell{69}{\textsuperscript{***}1.577} & \rmsecell{53}{\textsuperscript{***}1.674} & \rmsecell{21}{\textsuperscript{*}2.039} & \rmsecell{5}{\textsuperscript{***}\underline{2.069}} \\
Qwen2.5-VL & \rmsecell{85}{\textbf{1.394}} & \rmsecell{69}{\textsuperscript{***}1.440} & \rmsecell{53}{\textsuperscript{***}1.584} & \ctxmark{tab-model-predictions-cp-all-p-b}{\rmsecell{37}{\textsuperscript{***}1.664}} & \rmsecell{5}{\textsuperscript{***}\underline{1.782}} & \ctxmark{tab-model-predictions-cp-all-r-b}{\rmsecell{21}{\textsuperscript{***}1.706}} \\
\noalign{\ctxoutline{placeholderctx}{tab-model-predictions-cp-all-p-t}{tab-model-predictions-cp-all-p-b}\ctxoutline{realctx}{tab-model-predictions-cp-all-r-t}{tab-model-predictions-cp-all-r-b}}
\hline
\multicolumn{7}{c}{\textbf{Low imagery}} \\
\hline
LLaMA 2 & \rmsecell{85}{\underline{\textbf{1.278}}} & -- & -- & -- & -- & -- \\
Mistral & \rmsecell{85}{\underline{\textbf{1.819}}} & -- & -- & -- & -- & -- \\
Qwen2.5 & \rmsecell{85}{\underline{\textbf{1.389}}} & -- & -- & -- & -- & -- \\
Gemma 3 & \rmsecell{53}{1.662} & \ctxmark{tab-model-predictions-cp-lo-p-t}{\rmsecell{69}{\textsuperscript{***}1.611}} & \rmsecell{85}{\textsuperscript{***}\textbf{0.804}} & \rmsecell{37}{\textsuperscript{***}1.911} & \ctxmark{tab-model-predictions-cp-lo-r-t}{\rmsecell{5}{\textsuperscript{***}\underline{2.673}}} & \rmsecell{21}{\textsuperscript{***}2.567} \\
InternVL3 & \rmsecell{37}{1.590} & \rmsecell{53}{\textsuperscript{***}1.372} & \rmsecell{85}{\textsuperscript{***}\textbf{0.739}} & \rmsecell{69}{\textsuperscript{***}1.355} & \rmsecell{5}{\textsuperscript{***}\underline{1.926}} & \rmsecell{21}{\textsuperscript{***}1.842} \\
LLaVA 1.5 & \rmsecell{85}{\textbf{2.031}} & \rmsecell{69}{\textsuperscript{***}2.627} & \rmsecell{37}{\textsuperscript{***}3.040} & \rmsecell{53}{\textsuperscript{***}2.961} & \rmsecell{5}{\textsuperscript{***}\underline{3.066}} & \rmsecell{21}{\textsuperscript{***}3.063} \\
Pixtral & \rmsecell{21}{2.680} & \rmsecell{85}{\textsuperscript{***}\textbf{1.248}} & \rmsecell{69}{\textsuperscript{***}1.867} & \rmsecell{53}{\textsuperscript{***}1.868} & \rmsecell{37}{2.666} & \rmsecell{5}{\underline{2.709}} \\
Qwen2.5-VL & \rmsecell{69}{1.641} & \rmsecell{53}{1.649} & \rmsecell{37}{\textsuperscript{***}1.796} & \ctxmark{tab-model-predictions-cp-lo-p-b}{\rmsecell{85}{\textsuperscript{***}\textbf{1.533}}} & \rmsecell{5}{\textsuperscript{***}\underline{2.339}} & \ctxmark{tab-model-predictions-cp-lo-r-b}{\rmsecell{21}{\textsuperscript{***}2.239}} \\
\noalign{\ctxoutline{placeholderctx}{tab-model-predictions-cp-lo-p-t}{tab-model-predictions-cp-lo-p-b}\ctxoutline{realctx}{tab-model-predictions-cp-lo-r-t}{tab-model-predictions-cp-lo-r-b}}
\hline
\multicolumn{7}{c}{\textbf{High imagery}} \\
\hline
LLaMA 2 & \rmsecell{85}{\underline{\textbf{1.800}}} & -- & -- & -- & -- & -- \\
Mistral & \rmsecell{85}{\underline{\textbf{0.809}}} & -- & -- & -- & -- & -- \\
Qwen2.5 & \rmsecell{85}{\underline{\textbf{1.158}}} & -- & -- & -- & -- & -- \\
Gemma 3 & \rmsecell{85}{\textbf{1.087}} & \ctxmark{tab-model-predictions-cp-hi-p-t}{\rmsecell{37}{\textsuperscript{***}1.213}} & \rmsecell{5}{\textsuperscript{***}\underline{2.567}} & \rmsecell{21}{\textsuperscript{***}1.698} & \ctxmark{tab-model-predictions-cp-hi-r-t}{\rmsecell{53}{\textsuperscript{**}1.181}} & \rmsecell{69}{\textsuperscript{**}1.160} \\
InternVL3 & \rmsecell{53}{0.918} & \rmsecell{37}{\textsuperscript{***}1.088} & \rmsecell{5}{\textsuperscript{***}\underline{2.838}} & \rmsecell{21}{\textsuperscript{***}1.964} & \rmsecell{85}{\textsuperscript{***}\textbf{0.815}} & \rmsecell{69}{\textsuperscript{***}0.836} \\
LLaVA 1.5 & \rmsecell{5}{\underline{1.163}} & \rmsecell{85}{\textsuperscript{***}\textbf{0.903}} & \rmsecell{53}{\textsuperscript{***}1.045} & \rmsecell{69}{\textsuperscript{***}1.035} & \rmsecell{21}{\textsuperscript{***}1.061} & \rmsecell{37}{\textsuperscript{***}1.057} \\
Pixtral & \rmsecell{85}{\textbf{1.034}} & \rmsecell{5}{\textsuperscript{***}\underline{1.534}} & \rmsecell{37}{\textsuperscript{***}1.243} & \rmsecell{21}{\textsuperscript{***}1.467} & \rmsecell{69}{\textsuperscript{***}1.173} & \rmsecell{53}{\textsuperscript{***}1.182} \\
Qwen2.5-VL & \rmsecell{53}{1.113} & \rmsecell{37}{\textsuperscript{***}1.209} & \rmsecell{21}{\textsuperscript{***}1.354} & \ctxmark{tab-model-predictions-cp-hi-p-b}{\rmsecell{5}{\textsuperscript{***}\underline{1.780}}} & \rmsecell{69}{\textsuperscript{***}1.006} & \ctxmark{tab-model-predictions-cp-hi-r-b}{\rmsecell{85}{\textsuperscript{***}\textbf{0.963}}} \\
\noalign{\ctxoutline{placeholderctx}{tab-model-predictions-cp-hi-p-t}{tab-model-predictions-cp-hi-p-b}\ctxoutline{realctx}{tab-model-predictions-cp-hi-r-t}{tab-model-predictions-cp-hi-r-b}}
\hline
\end{tabular}
}
\caption*{CP2004B (Imagery)}
\end{minipage}

\caption{RMSE of model predictions against human concreteness (left) and imagery (right) ratings.
\emph{None} denotes no image. \placeholderctxtext{\emph{White}/\emph{Noise}} are uninformative contexts, and \placeholderctxtext{\emph{Shuffled}} is a mismatched-image control. \realctxtext{\emph{ImageNet}/\emph{Wikimedia}} are matched real-image contexts.
Green/purple cells indicate better/worse within-row ranks. \textbf{Boldface}/\underline{underlining} mark best/worst contexts.
Superscripts mark significance versus \emph{None} for the same model (\textsuperscript{*} $q<0.05$, \textsuperscript{**} $q<0.01$, \textsuperscript{***} $q<0.001$, paired sign-flip tests with Benjamini--Hochberg correction).}
\label{tab:model_predictions}
\end{table*}

\section{Experiment Setup}
\subsection{Models}
For text-only models, we evaluate LLaMA 2 \citep{Touvron2023Llama2O}, Mistral \citep{Jiang2023Mistral7}, and Qwen2.5 \citep{Yang2024Qwen25TR}.
For VLMs, we evaluate LLaVA 1.5 \citep{Liu2023ImprovedBW}, InternVL3 \citep{Zhu2025InternVL3EA}, Pixtral \citep{Agrawal2024Pixtral1}, Qwen2.5-VL \citep{Bai2025Qwen25VLTR}, and Gemma 3 \citep{Kamath2025Gemma3T}.
All models are instruction-tuned.
Qwen2.5-VL is the vision--language counterpart of Qwen2.5, making this family-level comparison more direct when contextualizing the effect of vision fine-tuning.
Checkpoint and hardware details are provided in Appendices~\ref{sec:appendix_model_checkpoints} and \ref{sec:appendix_hardware}.

\subsection{Dataset}
We evaluate the models on two standard datasets of human ratings: MT40k \citep{Brysbaert2014ConcretenessRF} for concreteness and CP2004B \citep{Clark2004ExtensionsOT} for imagery.
For concreteness, we use a CP2004B-aligned subset of MT40k. For imagery, we use CP2004B entries with valid imagery ratings.
After restricting both resources to single-word entries, the evaluation contains 2,082 MT40k words and 2,108 CP2004B words.
In MT40k, concreteness is rated on a 5-point scale from 1 (highly abstract) to 5 (highly concrete).
In CP2004B, imagery is rated on a 7-point scale from 1 (low imagery) to 7 (high imagery).

For each word, we retrieve real images from ImageNet \citep{Deng2009ImageNetAL} and Wikimedia Commons using Navigu \citep{Barthel2023navigunetNI}.
Because abstract and low-imagery words often lack clear visual referents, these retrieved images should be understood as best-effort matches rather than guaranteed semantic ground truth.
We assess retrieval quality through image--word relevance annotations on a CP2004B subset in Appendix~\ref{sec:appendix_relevance_annotation}.
In addition to these two matched real-image sources, we evaluate four controls: \emph{None} (no image), \emph{White} (a white image), \emph{Noise} (a random-noise image), and \emph{Shuffled} (a mismatched ImageNet image). Additional dataset preprocessing and visual-input details are provided in Appendices~\ref{sec:appendix_source_data} and \ref{sec:appendix_visual_input_contexts}.

\subsection{Prompt Design}
For both tasks, we prompt the models to assign a numerical rating to a target word using the original dataset scales.
The intended judgment is lexical and metalinguistic: the model is asked to rate a general property of the word, not the concreteness or imagery of the accompanying image.
Thus, the matched real-image contexts should be read as an incidental-context robustness test rather than as evidence that should determine the rating.
To reduce prompt-template variation, we use five prompt variants per task and average the parsed predictions across them.
This aggregation reduces prompt-template variance, but may hide prompt-specific behavior. We therefore report prompt-level variation in the representation analyses.

In the primary prediction experiment, VLMs receive the same textual prompts together with one of the six visual contexts (\emph{None}, \emph{White}, \emph{Noise}, \emph{Shuffled}, \emph{ImageNet}, \emph{Wikimedia}).
The representation analyses use the same contexts except \emph{Shuffled}, which serves as a prediction-level control.
We recognize that real-image contexts can nevertheless change the pragmatic frame of the prompt by making a visual instance salient.
This ambiguity across word--image pairs is part of the phenomenon under study: a VLM may treat visual context as task-relevant even when the requested judgment is lexical.
The \emph{None}, \emph{White}, and \emph{Noise} controls help separate image presence and low-level visual input from real-image content. The matched human validation study in Appendix~\ref{sec:appendix_human_validation} provides the corresponding human comparison.
The full prompt templates are listed in Appendix~\ref{sec:appendix_prompt_variants}.

\subsection{Evaluation Methodology}

\paragraph{Model predictions.}
For each model and each word--image pair, we extract a numerical prediction from the model output and evaluate it against the human rating using root mean square error (RMSE).
We report RMSE on the full evaluation set and on abstract, concrete, low-imagery, and high-imagery subsets.
We define abstract words as MT40k ratings below 3 (954 items) and concrete words as ratings of at least 3 (1,128 items). Low-imagery words have CP2004B ratings below 4 (1,023 items) and high-imagery words have ratings of at least 4 (1,085 items).
Because RMSE penalizes deviations on the original rating scale, it measures whether predictions remain calibrated to human judgments under different visual contexts.
Across all evaluated model--context combinations, 6.19\% of MT40k responses and 2.29\% of CP2004B responses do not contain a valid in-range number. Averaging across the five prompt variants reduces this to 140 of 68,706 missing MT40k predictions (0.204\%) and 18 of 69,564 missing CP2004B predictions (0.026\%).
Parsing and aggregation details are provided in Appendix~\ref{sec:appendix_output_parsing_aggregation}.

\paragraph{Probing.}
We probe layer-wise hidden representations by training regressors to predict concreteness and imagery ratings from hidden states.
This lets us test how strongly these properties are recoverable from hidden representations across network depth under different visual contexts.
For each layer, text-token representations are obtained by pooling over attended non-image tokens.
For VLMs, we also probe image-token representations, pooled over model-specific image placeholder tokens, and combined representations, computed by averaging the pooled text-token and image-token representations.
As a sanity check, we also report permuted-label probing baselines in Appendix~\ref{sec:appendix_additional_results}, obtained by training the same probes after permuting the target ratings.
Complete probing configurations are provided in Appendix~\ref{sec:appendix_probing_details}.

\paragraph{Canonical correlation analysis.}
We use canonical correlation analysis (CCA) to compare final-layer hidden representations produced under different visual contexts, such as \emph{None} versus real-image contexts.
Given paired representation matrices \(\mathbf{X}\) and \(\mathbf{Z}\), where rows correspond to words and columns to hidden dimensions under two visual contexts, CCA finds projection vectors \(\mathbf{a}\) and \(\mathbf{b}\) that maximize
\begin{equation}
	\rho = \max_{\mathbf{a},\mathbf{b}} \mathrm{corr}(\mathbf{X}\mathbf{a}, \mathbf{Z}\mathbf{b}).
\end{equation}
Here \(\rho\) denotes the resulting canonical correlation. CCA gives a similarity spectrum between two representation spaces: higher canonical correlations indicate that the representations remain relatively stable across contexts, whereas lower correlations indicate larger context-associated shifts.
Implementation details are provided in Appendix~\ref{sec:appendix_cca_details}.

\paragraph{Interpretability.}
To examine how the models integrate textual and visual information when producing concreteness and imagery judgments, we conduct attribution analysis using Integrated Gradients (IG), implemented with Layer Integrated Gradients \citep[LayerIG;][]{Sundararajan2017AxiomaticAF} in Captum \citep{Kokhlikyan2020CaptumAU}.
We use a structure-preserving baseline and compute token-level attributions for both text and image tokens.
This allows us to visualize which tokens drive a prediction and to summarize modality reliance by comparing the total attribution mass assigned to text tokens versus image tokens across visual contexts.
The full attribution formulation and implementation details are provided in Appendix~\ref{sec:appendix_interpretability_details}.

\begin{figure*}[t]
	\centering

	\begin{minipage}[t]{\textwidth}
		\centering
		\includegraphics[width=\linewidth]{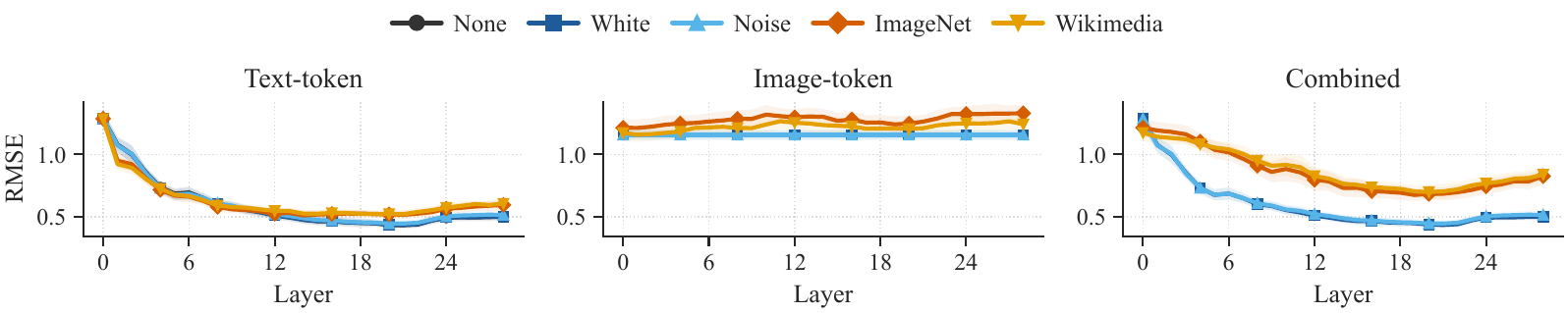}
		\caption*{(a) MT40k (Concreteness)}
	\end{minipage}

	\vspace{0.35em}

	\begin{minipage}[t]{\textwidth}
		\centering
		\includegraphics[width=\linewidth]{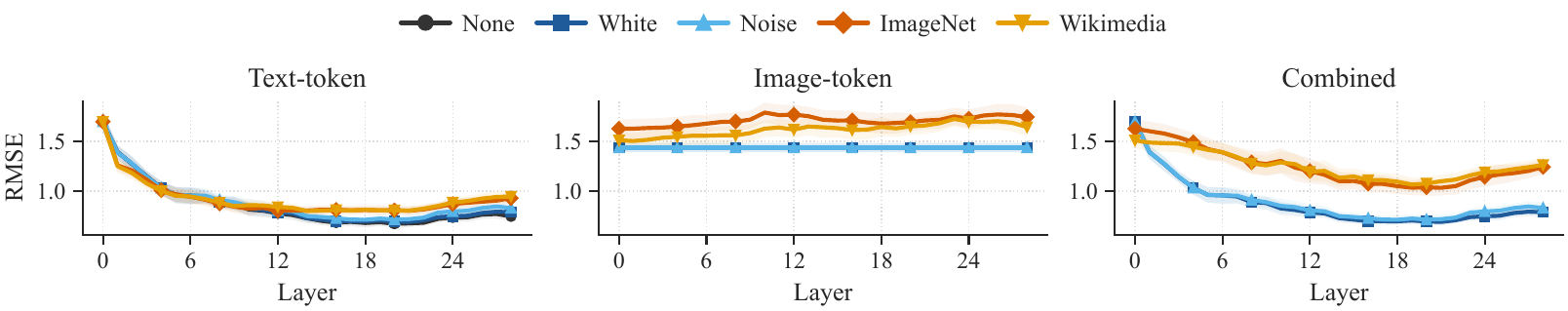}
		\caption*{(b) CP2004B (Imagery)}
	\end{minipage}

	\caption{Layer-wise ridge probing for Qwen2.5-VL on (a) concreteness and (b) imagery.
		Panels group representation types. Curves show visual contexts.
		Error bands show standard deviations across cross-validation folds and prompt templates.}
	\label{fig:probing}
\end{figure*}

\begin{figure*}[t]
	\centering
	\begin{minipage}{0.49\textwidth}
		\centering
		\includegraphics[width=\linewidth]{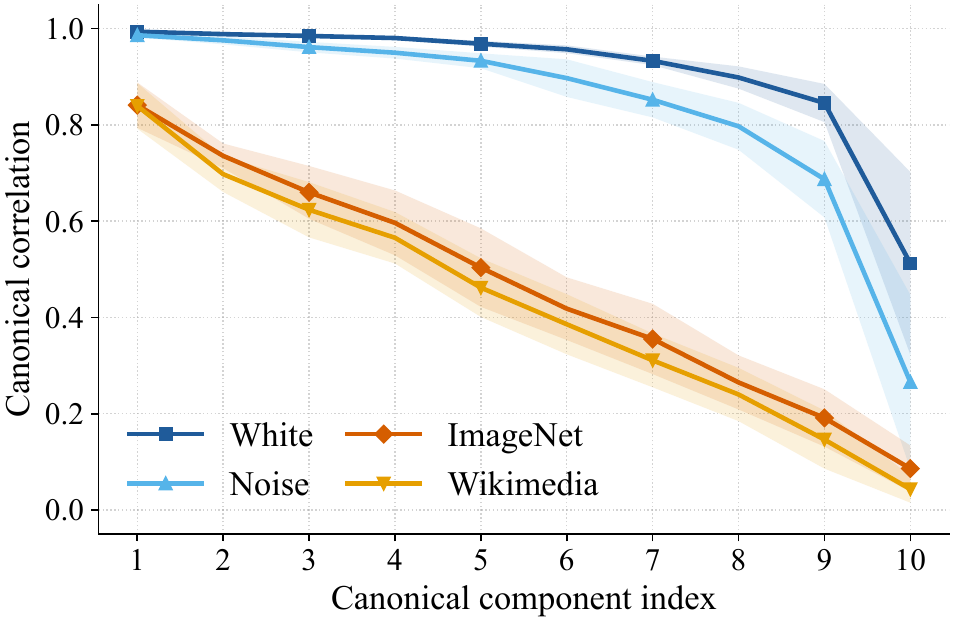}
		\caption*{(a) MT40k (Concreteness)}
	\end{minipage}
	\hfill
	\begin{minipage}{0.49\textwidth}
		\centering
		\includegraphics[width=\linewidth]{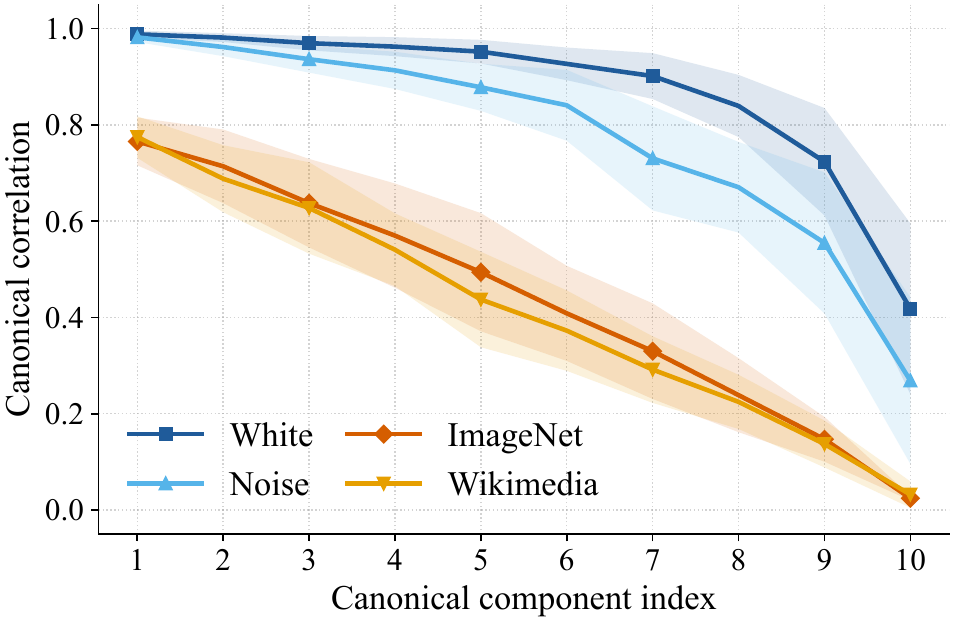}
		\caption*{(b) CP2004B (Imagery)}
	\end{minipage}

	\caption{Canonical correlations between Qwen2.5-VL final-layer representations under \emph{None} and each of the other four representation-analysis contexts for (a) concreteness and (b) imagery.
		Correlations use the top 10 PCA components. Error bands show standard deviations across prompt templates.}
	\label{fig:cca}
\end{figure*}

\begin{figure*}[t]
	\centering
	\includegraphics[width=\textwidth]{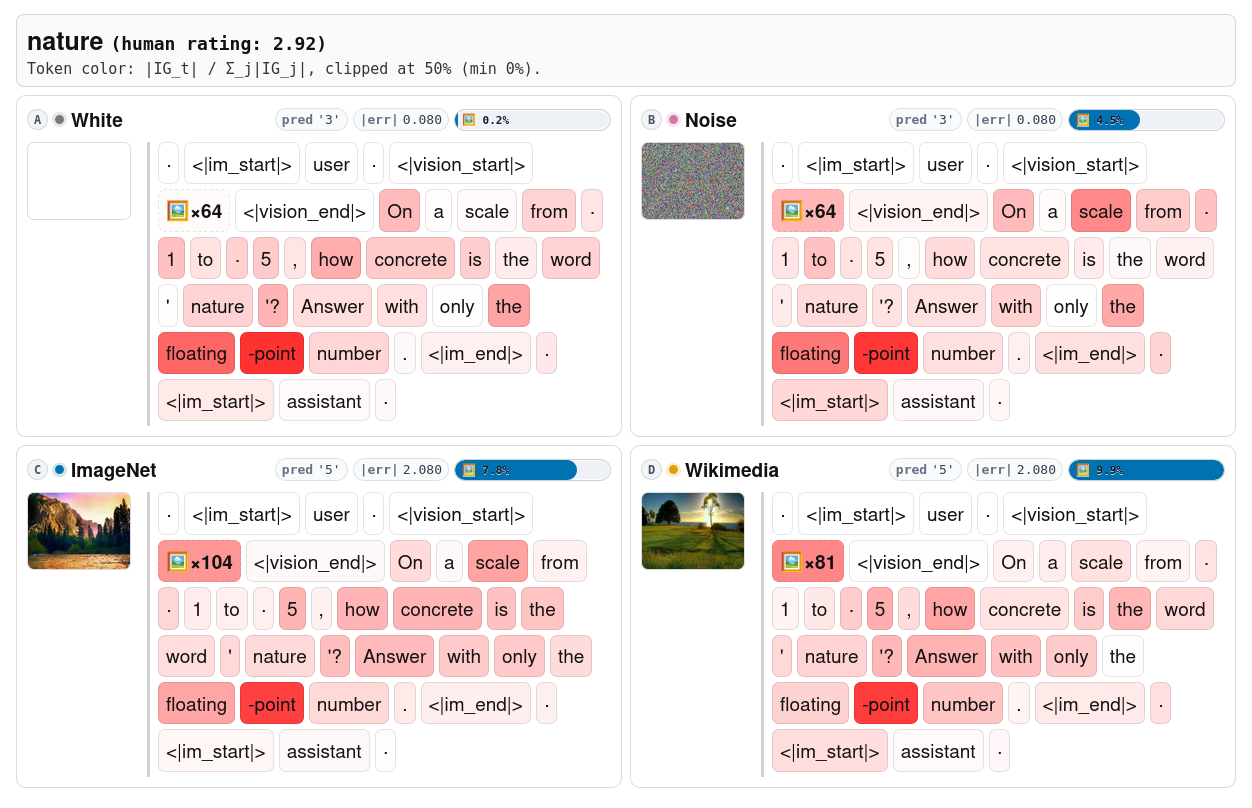}
	\caption{
		Integrated Gradients for Qwen2.5-VL on \textit{``nature''} (human rating: 2.92) under four visual contexts.
		Panels report the predicted next token, absolute error, and image-token attribution share. Token color shows normalized attribution magnitude.
	}
	\label{fig:interpretability}
\end{figure*}

\begin{figure*}[t]
	\centering
	\begin{minipage}{0.49\textwidth}
		\centering
		\includegraphics[width=\linewidth]{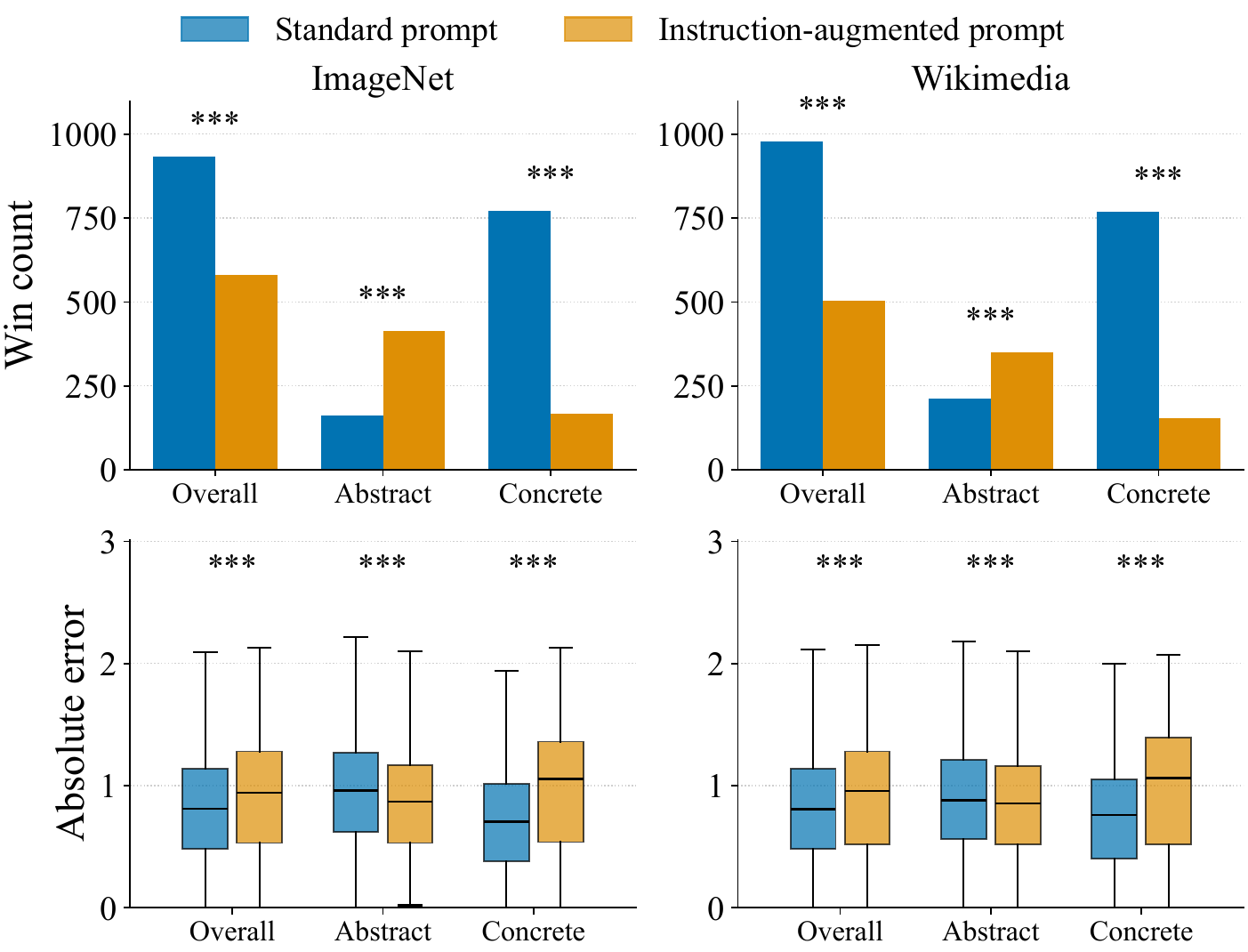}
		\caption*{(a) MT40k (Concreteness)}
	\end{minipage}
	\hfill
	\begin{minipage}{0.49\textwidth}
		\centering
		\includegraphics[width=\linewidth]{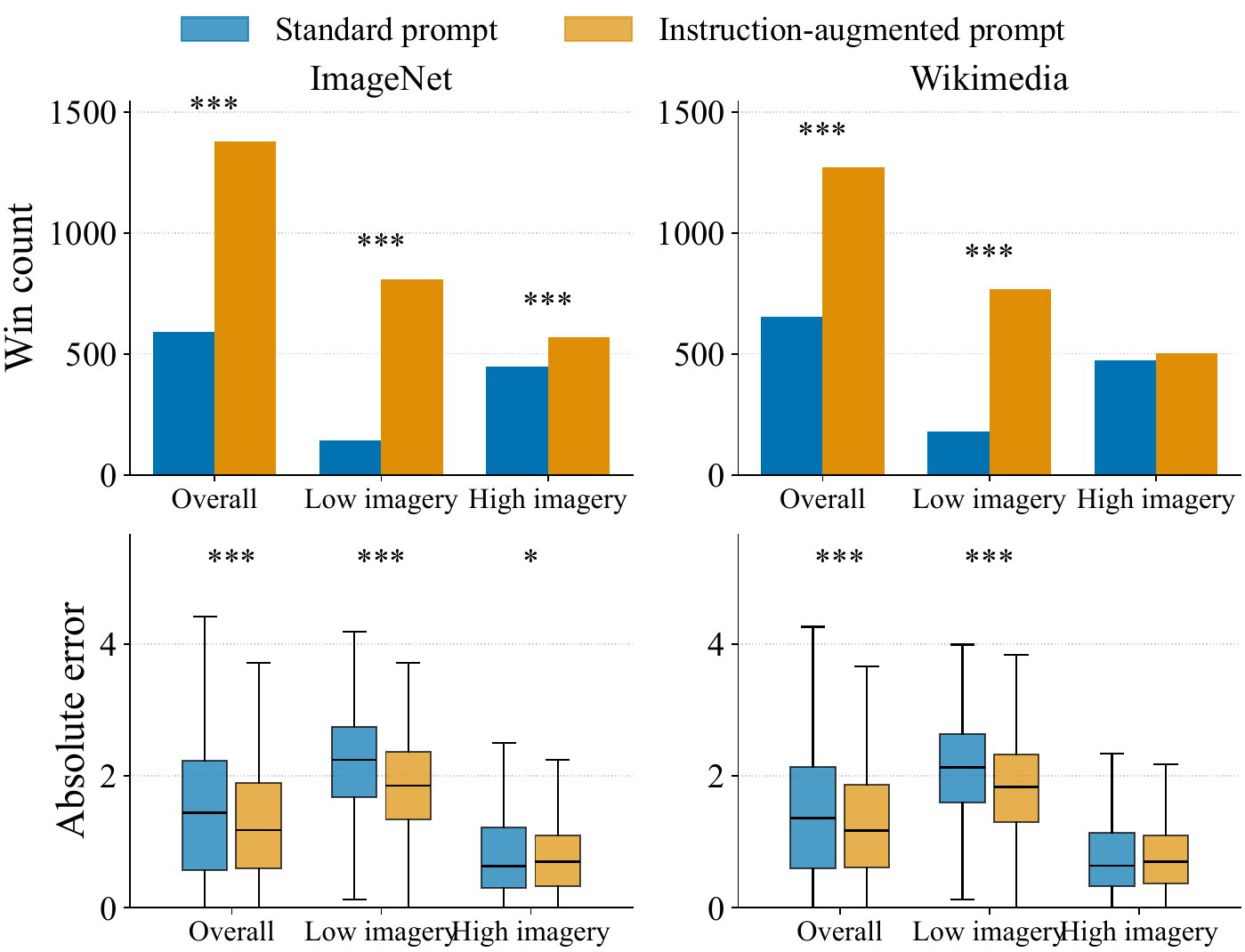}
		\caption*{(b) CP2004B (Imagery)}
	\end{minipage}
	\caption{Instruction-augmented prompting for Qwen2.5-VL in real-image contexts on (a) concreteness and (b) imagery.
		Top rows show non-tied win counts, excluding tied absolute errors. Bottom rows show absolute-error distributions over all valid paired predictions, overall and by rating subset.
		Superscripts mark significance versus standard prompting after Benjamini--Hochberg correction: exact sign tests for non-tied win counts and paired sign-flip tests for absolute-error differences (\textsuperscript{*} $q<0.05$, \textsuperscript{**} $q<0.01$, \textsuperscript{***} $q<0.001$).}
	\label{fig:mitigation}
\end{figure*}

\section{Results and Discussion}
Our primary finding is that VLMs are vulnerable to visual cues that can be spurious for lexical judgments of concreteness and imagery.
With real-image contexts, these models often show calibration loss and stronger sensitivity to visual content.
This is most evident on abstract and low-imagery words, where text-only models or VLMs given uninformative contexts often remain better calibrated.
The following sections present detailed analyses supporting this conclusion.

\subsection{Model Predictions}

Table~\ref{tab:model_predictions} presents the RMSE of model predictions compared to human ratings for concreteness and imagery.
Text-only models (LLaMA, Mistral, Qwen) are competitive with and often stronger than their vision–language counterparts, indicating that vision fine-tuning does not inherently enhance performance on these tasks in our setup.
Among VLMs, performance varies substantially based on the type of visual input provided.
When real-image contexts are provided (from either \emph{ImageNet} or \emph{Wikimedia}), models often improve for concrete and high-imagery words but frequently worsen for abstract and low-imagery words.
For abstract and low-imagery subsets, at least one of the \emph{None} and \emph{White} contexts yields lower RMSE than both matched real-image contexts for every VLM.

Overall effects are therefore heterogeneous across models and subsets rather than uniformly positive or uniformly negative.
Mean signed error analyses further clarify the direction of these effects.
Matched real-image contexts often have the most positive signed errors within a model/subset, indicating an upward shift toward higher concreteness or imagery ratings relative to other visual contexts, rather than simply an increase in RMSE (Appendix~\ref{sec:appendix_signed_shifts}).
Spearman rank correlation results broadly support the RMSE pattern, while also showing that calibration and relative word ordering can diverge (Appendix~\ref{sec:appendix_correlation_tables}).

The improved performance on concrete and high-imagery words suggests that real-image contexts can be useful when the lexical target has clearer visual content, consistent with prior work showing that visual supervision can aid word learning in some settings \citep{Zhuang2023VisualGH,Zhuang2024LexiconLevelCV}.
However, the substantial performance drop on abstract and low-imagery words supports the hypothesis that VLMs can be distracted by incidental visual cues in real-image contexts when image--word alignment is weak.

\subsection{Shuffled-Image Control}
To isolate the role of image--word alignment, we construct a mismatched real-image control by shuffling the \emph{ImageNet} images across target words while holding the target words, prompts, models, and image multiset fixed.
Compared with \emph{None} and \emph{White}, \emph{Shuffled} generally yields higher RMSE (Table~\ref{tab:model_predictions}) and almost uniformly lower rank correlation (Appendix~\ref{sec:appendix_correlation_tables}).
Relative to \emph{Noise}, its RMSE pattern is mixed, although rank correlation is usually lower.
Unlike the matched real-image contexts, \emph{Shuffled} does not produce a consistent upward rating shift.
Its signed-error differences from the controls vary across models and subsets (Appendix~\ref{sec:appendix_signed_shifts}).
Compared with \emph{Shuffled}, matched \emph{ImageNet} and \emph{Wikimedia} contexts produce higher ratings, increasing error for abstract and low-imagery words while often reducing error for concrete and high-imagery words.
This contrast distinguishes the general disruption associated with mismatched real images from the directional upward shift associated with plausible but task-incidental image--word matches.

\subsection{Human Validation}
To test whether the broader retrieved-image pattern reflects VLM-specific sensitivity rather than a general pragmatic effect of making an image salient, Appendix~\ref{sec:appendix_human_validation} reports a matched human validation study on a 500-word CP2004B subset sampled evenly across ten imagery deciles.
Fifty participants rated the same \emph{White}/\emph{ImageNet} manipulation, and the VLM pattern on this subset matches the full CP2004B result.
Human raters also shift upward under retrieved images, indicating that images can influence lexical ratings in both humans and VLMs.
However, humans do not show the same broad loss of calibration as VLMs, indicating that the VLM degradation cannot be reduced to the pragmatic effect of making an image salient during the judgment.

A separate question is whether retrieval quality accounts for this model--human difference.
Appendix~\ref{sec:appendix_relevance_annotation} reports image--word relevance annotations for the same subset. Because annotation agreement was moderate (Fleiss' $\kappa=0.40$), we treat these labels as a noisy diagnostic.
Within each relevance category, many VLM rating shifts remain significantly larger than the corresponding human shifts, including for images judged unrelated or misleading.
Retrieval quality therefore explains some variation but does not eliminate the VLM-specific sensitivity to visual context.

\subsection{Probing}
Figure~\ref{fig:probing} shows layer-wise probing performance for concreteness and imagery for Qwen2.5-VL under different visual contexts.
For brevity, we present results using ridge regressors. Results for other models, regressors, and metrics exhibit similar trends (see Appendix~\ref{sec:appendix_additional_results}).
For image inputs, we separately probe text-token and image-token hidden representations as well as the averaged hidden representations combining both modalities.
Under \emph{None}, probing performance increases steadily across layers and stabilizes at higher layers, indicating that concreteness and imagery are robustly encoded in later hidden representations.
A similar pattern appears with uninformative input images (\emph{White} or \emph{Noise}), whose probing curves closely match \emph{None}.

However, using real-image contexts in our Qwen2.5-VL probing setup is associated with degraded probing performance across layers.
The image-token hidden representations yield consistently poor performance, suggesting that visual features alone do not effectively capture concreteness or imagery information.
The text-token hidden representations in real-image contexts also show reduced performance compared with \emph{None} and uninformative image contexts, indicating that these linguistic properties are less recoverable from text-token representations in real-image settings.
When combining text and image hidden representations, performance remains between that of text and image tokens alone, which is consistent with real-image contexts contributing visual signals that are not always helpful for this task.

This pattern is consistent with prior findings of asymmetric cross-modal influence in multimodal transformers, where textual representations can be more strongly affected by visual input than the reverse direction \citep{Frank2021VisionandLanguageOV}.
In particular, although early and mid layers sometimes exhibit transient increases, performance drops markedly in later layers, which prior work on {BERT} has associated with higher-level semantic tasks \citep{Tenney2019BERTRT}.
Across representation types, these results indicate that, in this setting, real-image contexts are associated with weaker recoverability of concreteness and imagery, especially when visual evidence is weakly aligned with the lexical target.

\subsection{Canonical Correlation Analysis}
Figure~\ref{fig:cca} presents the canonical correlations between the final-layer hidden representations obtained with different visual contexts for Qwen2.5-VL.
The \emph{None} visual context is compared against \emph{White}, \emph{Noise}, and real-image contexts.
Hidden representations from uninformative contexts (\emph{White} and \emph{Noise}) exhibit high canonical correlations with \emph{None} across the top principal components, indicating that the model's hidden representations remain largely consistent regardless of the presence of uninformative visual input.
In contrast, hidden representations from real-image contexts show markedly lower canonical correlations with \emph{None}, with the last few components exhibiting correlations close to zero.
This divergence indicates that real-image contexts are associated with representational shifts relative to \emph{None}, consistent with stronger sensitivity to visual content.
These findings align with the probing results and the prediction shifts observed under matched real-image contexts.

\subsection{Interpretability: A Case Study}

Figure~\ref{fig:interpretability} presents a case study of Qwen2.5-VL on the target word \textit{nature} (human concreteness rating: 2.92) using LayerIG.
The example shows how attribution shifts from text tokens toward image tokens as the input changes from uninformative to real-image contexts, and how this shift tracks changes in prediction error.

With \emph{White} and \emph{Noise} contexts, attributions are concentrated on text tokens, and the model predicts 3, close to the human rating (absolute error 0.08).
This indicates that when visual input is uninformative, the model relies primarily on linguistic cues and remains relatively well calibrated.

However, in real-image contexts (\emph{ImageNet} and \emph{Wikimedia}), the attribution landscape shifts.
Image-token attribution rises from 0.2\% for \emph{White} and 4.5\% for \emph{Noise} to 7.8\% for \emph{ImageNet} and 9.9\% for \emph{Wikimedia}, while the predicted token shifts upward to 5 in both real-image contexts (absolute error 2.08).
The concurrent rise in image-token attribution and error suggests greater image reliance and weaker calibration.
The resulting overestimation is consistent with our broader finding that real-image contexts are associated with upward-biased concreteness judgments when visual evidence is incidental to the lexical target.

The probing and CCA results show that prediction shifts under matched real-image contexts coincide with weaker recoverability of lexical properties and larger representational changes.
The attribution case study complements these aggregate analyses by illustrating how increased image-token influence can accompany greater error in an individual prediction.
Across methods, the evidence indicates that VLMs can overuse incidental visual context in lexical judgments.

\section{Mitigation Strategies}
Our preceding analyses showed that matched real-image contexts can worsen alignment with human ratings and are associated with shifted hidden representations and higher image-token attribution in a case study, especially when the visual content is incidental to the target word.
One plausible interpretation is that the model exhibits a form of shortcut learning at inference time: once a real-image context is present, it may rely on salient visual content as an easy cue, even when that cue is not reliably informative for the lexical property being rated \citep{Geirhos2020ShortcutLI}.
If this behavior partly depends on how the model interprets the prompt, then making the possible irrelevance of the image explicit may help it discount misleading visual evidence.

We therefore test a simple prompting intervention with Qwen2.5-VL as the representative model.
For each of the five original rating prompt variants, we prepend the same sentence:
\begin{quote}
	\textit{The image may be irrelevant to the word; please base your rating solely on the word itself.}
\end{quote}
Following the instructed-prompting approach of \citet{Shi2023LargeLM}, this intervention does not alter the model's parameters or representations directly. Instead, it changes the decision context by making the possible irrelevance of the image explicit.

Figure~\ref{fig:mitigation} compares the standard and instruction-augmented prompts under both matched real-image contexts using non-tied win counts and absolute-error distributions.
The effect is selective rather than uniform.
On MT40k, the instruction-augmented prompt wins frequently on abstract words, but gains are small in magnitude and the concrete subset still favors the standard prompt.
On CP2004B, the instruction-augmented prompt is preferred overall and especially for low-imagery words, with visibly lower error distributions in those subsets. By contrast, the high-imagery subset is close to parity and shows little magnitude difference between prompts.
Overall, even this single instruction can reduce some errors in real-image contexts, primarily for subsets most vulnerable to real-image-associated overestimation.
Appendix~\ref{sec:appendix_additional_mitigation} extends this analysis to other VLMs and tests five Qwen2.5-VL mitigation instruction variants.
These extensions show that the mitigation pattern is not unique to Qwen2.5-VL: several other VLMs also benefit, with the most consistent gains on CP2004B low-imagery words, while Qwen2.5-VL wording variants preserve the same selective tradeoff between improved abstract or low-imagery subsets and weaker concrete-word results.

Because this intervention operates entirely at inference time, it offers a lightweight guardrail for noisy, weakly aligned, or incidental visual contexts, though stronger mitigation will likely benefit from methods that adaptively calibrate visual evidence.

\section{Conclusion}
We show that visual grounding is not uniformly beneficial for lexical judgments.
Matched real-image contexts can improve predictions for concrete and high-imagery words, yet undermine alignment with human ratings for abstract and low-imagery words.
Crucially, a matched human study shows no comparable degradation, indicating that this pattern is not solely a pragmatic effect of making an image salient.
Probing and CCA associate these errors with weaker recoverability of lexical properties and context-associated representational shifts, while the attribution case study illustrates increased image-token influence alongside greater error.
A simple instruction reduces some errors in vulnerable subsets, although the gains remain selective.
Together, these findings distinguish the availability of visual evidence from its relevance to the requested judgment.
Current instruction-tuned VLMs need better calibration of when visual context should affect their predictions.

\section*{Limitations}
First, we evaluate only English single-word norms (MT40k and CP2004B), so the findings may not transfer to other languages, multi-word expressions, or sentence-level grounding.
Human validation covers 500 CP2004B words and only the \emph{White}/\emph{ImageNet} comparison, leaving MT40k and other image sources untested.
Broader human studies could test whether the model--human contrast generalizes across tasks and populations.

Second, retrieved \emph{ImageNet} and \emph{Wikimedia} images may poorly match their target words, particularly abstract and low-imagery words.
Our relevance labels cover one image per word and have moderate agreement.
Shuffled images may also be incidentally relevant.
Controlled stimuli and multiple images per word would better separate context effects from retrieval artifacts.

Third, our mitigation remains lightweight and prompt-level.
Five Qwen2.5-VL instruction wordings (Appendix~\ref{sec:appendix_mitigation_prompt_variants}) do not establish the limits of stronger calibration, and the fixed intervention does not estimate item-level image usefulness.
Adaptive or training-based calibration may yield broader improvements.

\section*{Acknowledgment}
This work is supported in part by the NSERC Discovery Grant RGPIN-2024-04395 and a Canada CIFAR AI Chair Award to FS.
AI assistants were used for coding and proofreading. All AI-assisted outputs were reviewed and verified by the authors.


\bibliography{custom}

\clearpage
\appendix
\section{Implementation Details}
\label{sec:appendix}

\subsection{Model Checkpoints}
\label{sec:appendix_model_checkpoints}
Table~\ref{tab:appendix_models} lists the model identifiers used in our pipeline.
All models are loaded via their Hugging Face implementations.

\begin{table}[h]
	\centering
	\footnotesize
	\resizebox{\linewidth}{!}{%
	\begin{tabular}{l l}
		\hline
		Model family & Hugging Face identifier                     \\
		\hline
		LLaMA 2      & \texttt{meta-llama/Llama-2-7b-chat-hf}      \\
		Mistral      & \texttt{mistralai/Mistral-7B-Instruct-v0.3} \\
		Qwen2.5      & \texttt{Qwen/Qwen2.5-7B-Instruct}           \\
		LLaVA 1.5    & \texttt{llava-hf/llava-1.5-7b-hf}           \\
		InternVL3    & \texttt{OpenGVLab/InternVL3-8B-hf}          \\
		Pixtral      & \texttt{mistral-community/pixtral-12b}      \\
		Qwen2.5-VL   & \texttt{Qwen/Qwen2.5-VL-7B-Instruct}        \\
		Gemma 3      & \texttt{google/gemma-3-4b-it}               \\
		\hline
	\end{tabular}%
	}
	\caption{Model checkpoints used in the experiments.}
	\label{tab:appendix_models}
\end{table}

\subsection{Source Data, Filtering, and Licenses}
\label{sec:appendix_source_data}
We construct task-specific single-word evaluation sets from MT40k and CP2004B.
For MT40k, we start from a CP2004B-aligned file and drop missing ratings, bigrams, hyphenated forms, and duplicate lowercased surface forms.
For CP2004B, we retain entries with valid positive imagery ratings and apply the same normalization, hyphen filtering, and deduplication.
Table~\ref{tab:appendix_dataset_stats} summarizes the resulting rating distributions.

\begin{table}[h]
	\centering
	\footnotesize
	\setlength{\tabcolsep}{3pt}
	\resizebox{\linewidth}{!}{
		\begin{tabular}{lrrrrrr}
			\hline
			Dataset & Words & Mean & SD   & Range      & Unique & Med. gap \\
			\hline
			MT40k   & 2,082 & 3.29 & 1.16 & 1.07--5.00 & 326    & 0.01     \\
			CP2004B & 2,108 & 4.21 & 1.44 & 1.16--6.90 & 295    & 0.02     \\
			\hline
		\end{tabular}
	}
	\caption{Rating statistics for the processed single-word evaluation files. \emph{Unique}: number of distinct rating values. \emph{Med. gap}: median positive gap between adjacent values.}
	\label{tab:appendix_dataset_stats}
\end{table}

MT40k is distributed under CC BY-NC-ND 3.0,\footnote{\url{https://rdrr.io/github/JackEdTaylor/LexOPS/f/inst/extdata/data-licenses/BRYSBAERT-ET-AL-2014.md}} while CP2004B remains under Psychonomic archive terms that retain author rights and permit non-commercial research or educational use with citation.\footnote{\url{https://rdrr.io/github/JackEdTaylor/LexOPS/f/inst/extdata/data-licenses/CLARK-AND-PAIVIO-LICENSE.md}}
We do not redistribute original or processed item-level rating tables. Instead, we report summary statistics and aggregate results and provide preprocessing code for locally obtained sources.
Retrieved images remain subject to ImageNet and Wikimedia Commons source terms.\footnote{\url{https://www.image-net.org/download.php}}\footnote{\url{https://commons.wikimedia.org/wiki/Commons:Licensing}}
We release anonymized human-study ratings and newly produced relevance annotations under CC BY 4.0, excluding Prolific/session identifiers and raw third-party images.\footnote{\url{https://creativecommons.org/licenses/by/4.0/}}

\subsection{Prompt Variants}
\label{sec:appendix_prompt_variants}
For each word, we query each model with five prompt templates and average parsed numeric predictions across the five responses.
Below are the templates used for concreteness rating:
\begin{enumerate}
	\item \textit{On a scale from 1 to 5, how concrete is the word '\{word\}'? Answer with only the floating-point number.}
	\item \textit{Rate the concreteness of the word '\{word\}' from 1 (very abstract) to 5 (very concrete). Answer with only the floating-point number.}
	\item \textit{How tangible or concrete is the concept of '\{word\}'? Give a score from 1 to 5. Answer with only the floating-point number.}
	\item \textit{Evaluate how physically grounded the word '\{word\}' is, using a scale from 1 to 5. Answer with only the floating-point number.}
	\item \textit{Is the word '\{word\}' something you can perceive with your senses? Rate its concreteness from 1 to 5. Answer with only the floating-point number.}
\end{enumerate}
The templates used for imagery rating are:
\begin{enumerate}
	\item \textit{On a scale from 1 to 7, how easy is it to form a mental image of the word '\{word\}'? Answer with only the floating-point number.}
	\item \textit{How easy is it to picture the word '\{word\}' in your mind, from 1 (hard) to 7 (easy)? Answer with only the floating-point number.}
	\item \textit{Rate the imagery of the word '\{word\}' on a 1--7 scale. Answer with only the floating-point number.}
	\item \textit{Imagine the word '\{word\}'. How vivid is the mental image? Rate from 1 to 7. Answer with only the floating-point number.}
	\item \textit{How imageable is the word '\{word\}'? Respond with a number from 1 (least) to 7 (most). Answer with only the floating-point number.}
\end{enumerate}

\subsection{Visual Input Contexts}
\label{sec:appendix_visual_input_contexts}
We evaluate the same six visual contexts used in the main paper: \emph{None}, \emph{White}, \emph{Noise}, \emph{Shuffled}, \emph{ImageNet}, and \emph{Wikimedia}.
In the pipeline, the \emph{Noise} context is implemented as a random-noise placeholder image.
For matched real-image contexts, each word is associated with a directory of up to five retrieved images, which are randomly assigned to prompt templates within that word.
For the shuffled-image control, we apply a deterministic derangement with random seed 42 to the \emph{ImageNet} image paths.
The target words, ratings, prompts, and image multiset are unchanged, but every target receives an image retrieved for a different word.

\subsection{Output Parsing and Aggregation}
\label{sec:appendix_output_parsing_aggregation}
Model outputs are obtained by greedy decoding with a maximum generation length of 128 tokens.
To extract the predicted numeric rating from the model output, we keep the first numeric span when it falls within the valid task range.
Predictions outside the valid task range are discarded (1--5 for concreteness, 1--7 for imagery).
For each word, the final prediction is the mean of valid parsed numbers across the five prompt variants.
If no valid number is found, the item is treated as missing.

\subsection{Probing Details}
\label{sec:appendix_probing_details}
Probing uses 10-fold cross-validation on hidden representations.
For each layer and prompt template, text-token representations are computed by mean pooling hidden states over all attended non-image tokens after chat templating, and image-token representations are computed by mean pooling hidden states over model-specific image placeholder tokens.
The combined representation is the average of the pooled text-token and image-token vectors. When no image tokens are present, only text-token representations are used.
Curves report the mean across cross-validation folds and prompt templates, with standard deviations shown as error bands.

We report probing results with three regressor architectures: ridge, SVM, and MLP.
For ridge, we use scikit-learn's \texttt{Ridge} model.
For SVM, we use scikit-learn's \texttt{SVR} model with an RBF kernel.
For MLP, we use a one-hidden-layer regressor (hidden size 2048, ReLU, dropout 0.2) trained with Adam for 100 epochs (batch size 512, learning rate 0.001, weight decay \(1\times10^{-5}\)).
In all cases, features are scaled with min-max normalization.
All runs use random seed 42.

\subsection{CCA Details}
\label{sec:appendix_cca_details}
For each word, we extract final-layer hidden representations under each representation-analysis context and align words across the two contexts being compared (e.g., \emph{None} vs.\ \emph{White}, \emph{Noise}, or a matched real-image context).
This yields paired representation matrices $\mathbf{X}$ and $\mathbf{Z}$, where rows correspond to words and columns to hidden dimensions.
More concretely, if $\mathbf{x}_i, \mathbf{z}_i \in \mathbb{R}^d$ denote the paired final-layer hidden representations for word $i$, then $\mathbf{X} = [\mathbf{x}_1,\dots,\mathbf{x}_N]^\top$ and $\mathbf{Z} = [\mathbf{z}_1,\dots,\mathbf{z}_N]^\top$.

CCA then finds linear projection vectors $\mathbf{a}$ and $\mathbf{b}$ that maximize the correlation between the projected representations:
\begin{equation}
	\rho
	=
	\max_{\mathbf{a},\mathbf{b}}
	\frac{\mathbf{a}^\top \Sigma_{XZ} \mathbf{b}}
	{\sqrt{\mathbf{a}^\top \Sigma_{XX} \mathbf{a}\,
			\mathbf{b}^\top \Sigma_{ZZ} \mathbf{b}}},
\end{equation}
Here $\rho$ denotes the resulting canonical correlation, and $\Sigma_{XX}$, $\Sigma_{ZZ}$, and $\Sigma_{XZ}$ are the empirical covariance matrices and cross-covariance matrix.
Repeating this procedure yields a spectrum of canonical correlations $\{\rho_k\}_{k=1}^K$, which summarizes the similarity between the two representation spaces.

Before running CCA, we standardize each matrix feature-wise across words.
To improve numerical stability and reduce computational cost, we then apply Principal Component Analysis (PCA) separately to each matrix and retain the top \(K=10\) components.
CCA is performed on these reduced matrices to compute component-wise canonical correlations.
In Figure~\ref{fig:cca}, each curve reports the mean canonical correlation across prompt variants, and the error band denotes one standard deviation across prompt variants.

\subsection{Interpretability Details}
\label{sec:appendix_interpretability_details}
For attribution analysis, we use Captum's Layer Integrated Gradients (LayerIG) on the first language transformer block.
The attribution target is the logit of the model's predicted next token (argmax at the final position).
In our runs, we use \(\texttt{n\_steps}=8\) and prompt variant index 0.

The baseline is structure-preserving: template and special tokens remain unchanged, the attention mask is fixed, the user text span is replaced with a padding token (or EOS if padding is unavailable), and visual input is replaced with a black-image baseline in normalized pixel space.

Let \(F(\mathbf{x})\) denote the model output for input \(\mathbf{x} = (\mathbf{x}^{\text{text}}, \mathbf{x}^{\text{img}})\), and let \(\mathbf{x}'\) denote the corresponding structure-preserving baseline input.
For a token \(t\), let \(\mathbf{e}_t \in \mathbb{R}^d\) be its embedding vector and \(\mathbf{e}'_t \in \mathbb{R}^d\) the corresponding baseline embedding, with \(e_{t,k}\) and \(e'_{t,k}\) denoting their \(k\)-th dimensions. The dimension-wise Integrated Gradients are
\begin{equation}
	\begin{aligned}
	\mathrm{IG}_{t,k}(\mathbf{x})
	&=
	(e_{t,k} - e'_{t,k}) \\
	&\quad \times \int_{0}^{1}
	\frac{\partial F\big(\mathbf{x}' + \alpha(\mathbf{x}-\mathbf{x}')\big)}
	{\partial e_{t,k}}
	\, d\alpha.
	\end{aligned}
\end{equation}
Here \(\alpha\) interpolates along the path from \(\mathbf{x}'\) to \(\mathbf{x}\).
Token-level attributions are then obtained by summing signed attributions across embedding dimensions:
\begin{equation}
	\mathrm{IG}_t
	=
	\sum_{k=1}^{d}
	\mathrm{IG}_{t,k}.
\end{equation}
In practice, the path integral is approximated using a finite number of interpolation steps.
This formulation satisfies the completeness property, meaning that the sum of token attributions equals the difference between the model outputs at \(\mathbf{x}\) and \(\mathbf{x}'\).

For our case study, we convert these signed token attributions into absolute attribution mass.
Specifically, we report normalized token intensity \(|\mathrm{IG}_t|/\sum_j |\mathrm{IG}_j|\) and compute modality-level attribution splits as
\begin{equation}
	A_{\text{modality}} = \sum_{t \in \mathcal{T}_{\text{modality}}} |\mathrm{IG}_t|,
\end{equation}
for text tokens and image tokens in Figure~\ref{fig:interpretability}.

\subsection{Hardware}
\label{sec:appendix_hardware}
Experiments were run on NVIDIA L40S and H100 GPUs (48 GB and 80 GB memory, respectively).
Model loading uses Hugging Face \texttt{device\_map="auto"}, so computation is placed automatically on available GPUs.

\section{Human Validation Experiments}
\label{sec:appendix_human_validation}

\subsection{Experiment Setup}
\label{sec:appendix_human_validation_experiment_setup}

To check whether the real-image rating shift observed in VLMs also appears in human ratings, we ran a human validation study on a 500-word subset of CP2004B.
The subset was sampled evenly across ten CP2004B imagery deciles, so that both low- and high-imagery words were covered.
Participants were recruited through Prolific using screening criteria for English as a first language and residence in the United States, United Kingdom, Canada, Australia, New Zealand, or Ireland.
They rated target words on the CP2004B 1--7 imagery scale.

The study compared two visual contexts from the model analysis: the uninformative \emph{White} context and the retrieved \emph{ImageNet} context.
Fifty participants each completed 100 trials, with an equal number of \emph{White} and \emph{ImageNet} trials and no repeated word within a participant.
Across the full study, each word--image pair received five ratings.
We averaged human ratings at the word--image level and compared them with both the published CP2004B ratings and the corresponding VLM predictions on the same 500-word subset.

The study was configured for an estimated duration of 8--10 minutes and a maximum allowed time of 15 minutes. The observed mean completion time was 6.68 minutes (median 5.98 minutes) for the 50 completed sessions.
Prolific reported an average reward rate of \pounds{}12.10 per hour, above common minimum-wage thresholds.
The task collected ratings and Prolific URL identifiers for submission tracking. No free-text responses or other personal data were requested.
The data collection protocol was approved by an institutional research ethics review board.

\subsection{Participant Instructions}
\label{sec:appendix_human_validation_instructions}

Before the instructions, participants viewed a consent page stating the expected study duration and that Prolific submission information would be recorded. They proceeded only after indicating consent.
They were then warned that the task could include potentially disturbing images and unpleasant words, and were informed that they would rate about 100 English words.

The instruction page then presented the question ``How easily and quickly does this word arouse a mental image?''
It defined imagery as follows: ``Words differ in how easily they arouse mental images of things or events. A mental image can be a mental picture, sound, or other sensory experience. Give a high rating to words that arouse a mental image quickly and easily. Give a low rating to words that arouse a mental image only with difficulty, after a delay, or not at all. Use the scale to indicate the ease or difficulty with which each word arouses a mental image for you.''
The same page explained that the visual display could be a blank image or an image from an image dataset.
The response scale was shown as 1 = very difficult to image, 2 = difficult, 3 = somewhat difficult, 4 = moderate, 5 = somewhat easy, 6 = easy, and 7 = very easy to image.

\subsection{Results and Discussion}
\label{sec:appendix_human_validation_results}

Table~\ref{tab:human_vlm_metrics_subset} compares the averaged human ratings with VLM predictions on the same 500-word subset.
The central comparison is the change from the uninformative \emph{White} context to the retrieved \emph{ImageNet} context.
For RMSE, humans show only a small overall increase, whereas every VLM shows a larger and significant increase.
The contrast is clearest for low-imagery words: human ratings move modestly away from the published CP2004B ratings, but several VLMs show much larger losses in calibration.
The \(\Delta-\Delta_H\) columns confirm this contrast statistically: for the full subset and the low-imagery subset, every VLM's RMSE increase is significantly larger than the corresponding human increase.
For high-imagery words, humans and several VLMs instead improve under \emph{ImageNet}, consistent with real-image contexts being more useful when the word has clearer visual content.

Mean signed error shows the direction of the shift.
Both humans and VLMs assign higher imagery ratings under \emph{ImageNet}, consistent with prior evidence that semantic context can influence both picture and word processing \citep{Sperber1979SemanticPE}.
However, the upward shift is generally larger for VLMs, especially for low-imagery words.
The model--human difference in signed-error shift is significant for most VLMs across these splits, indicating that the retrieved images produce a stronger upward bias in model ratings than in human ratings.
Spearman rank correlation adds a complementary view: rank agreement can improve even when RMSE and signed error worsen, showing that visual context may preserve or sharpen item ordering while still shifting predictions away from the human rating scale.
The human comparison shows that the VLM degradation cannot be explained by the pragmatic effect of making an image salient during the judgment alone. Instead, VLMs are more sensitive than human raters to incidental visual content when making the lexical judgment.

\section{Image-Word Relevance Annotation}
\label{sec:appendix_relevance_annotation}

To assess retrieval quality as a potential confound in the CP2004B results, we manually annotated image--word relevance for the same 500-word subset used in the human validation study.
For each word, we reproducibly sampled one of the five \emph{ImageNet} images used in the human study.
Three annotators independently rated each word--image pair on a three-point scale: 0 for unrelated or misleading, 1 for weak, indirect, or ambiguous, and 2 for a clear visual match.
Each of the 500 pairs received three ratings.
Inter-annotator agreement was moderate (Fleiss' $\kappa=0.40$). Exact three-way agreement was 46.2\%, and mean pairwise agreement was 62.5\%.
All three annotators were authors of this paper.

Table~\ref{tab:relevance_annotation_distribution} summarizes the majority relevance labels.
Retrieved images were clear matches for most high-imagery words (72.3\%), but for only 20.9\% of low-imagery words.
Conversely, 26.6\% of low-imagery words were judged unrelated or misleading, compared with 3.9\% of high-imagery words.
This confirms that image--word match quality is strongly tied to the target lexical property: low-imagery words are exactly the cases where retrieved visual evidence is least reliable.

\begin{table}[h!]
	\centering
	\footnotesize
	\setlength{\tabcolsep}{4pt}
	\begin{tabular}{lrrr}
		\toprule
		Majority label & All      & Low imagery & High imagery \\
		\midrule
		Unrelated      & 75       & 65          & 10           \\
		               & (15.0\%) & (26.6\%)    & (3.9\%)      \\
		\addlinespace[1pt]
		Weak/ambig.    & 189      & 128         & 61           \\
		               & (37.8\%) & (52.5\%)    & (23.8\%)     \\
		\addlinespace[1pt]
		Clear match    & 236      & 51          & 185          \\
		               & (47.2\%) & (20.9\%)    & (72.3\%)     \\
		\bottomrule
	\end{tabular}
	\caption{Image--word relevance annotations for one sampled \emph{ImageNet} image per word in the 500-word CP2004B subset. Labels are majority votes over three annotators.}
	\label{tab:relevance_annotation_distribution}
\end{table}

To control for this retrieval-quality confound, we also conditioned the \emph{ImageNet}-minus-\emph{White} shift on the majority relevance labels (Table~\ref{tab:relevance_conditioned_shift}).
Human ratings show only a small shift for unrelated images (+0.05), whereas VLMs generally shift upward even when the image is judged unrelated or misleading.
Many VLM shifts are significantly larger than the corresponding human shifts, including for unrelated images.
Because these comparisons are made within each relevance category, the larger VLM shifts cannot be attributed only to poorer image matches for low-imagery words.
Even clear matches are often associated with larger model than human rating shifts, showing that relevance helps explain part of the variation but does not eliminate the VLM-specific sensitivity to visual context.
Thus, retrieval quality is a meaningful source of variation, but the relevance-conditioned analysis supports the main conclusion: VLMs are more sensitive than human raters to incidental visual content in this lexical judgment task.

\begin{table}[h!]
	\centering
	\scriptsize
	\setlength{\tabcolsep}{8pt}
	\resizebox{\linewidth}{!}{
	\begin{tabular}{@{}lrrr@{}}
		\toprule
		Source     & Unrel.            & Weak              & Clear             \\
		\midrule
		Human      & \textbf{+0.05}    & \textbf{+0.38}    & \textbf{+0.51}    \\
		\midrule
		Gemma 3    & \textsuperscript{***}+0.83 & \textsuperscript{***}+1.22 & \textsuperscript{***}+1.19 \\
		InternVL3  & +0.30             & \textsuperscript{*}+0.61   & \textsuperscript{***}+0.90 \\
		LLaVA 1.5  & \textsuperscript{*}+0.43   & +0.53             & +0.52             \\
		Pixtral    & \textsuperscript{***}\underline{+1.65} & \textsuperscript{***}\underline{+1.65} & \textsuperscript{***}\underline{+1.91} \\
		Qwen2.5-VL & \textsuperscript{*}+0.44   & \textsuperscript{***}+0.87 & \textsuperscript{***}+1.33 \\
		\bottomrule
	\end{tabular}
	}
	\caption{Mean \emph{ImageNet}-minus-\emph{White} rating shifts on the 500-word CP2004B subset by majority image--word relevance label. Positive values indicate higher ratings under \emph{ImageNet}. Star superscripts mark paired significance versus the human shift after Benjamini--Hochberg correction within each relevance label (\textsuperscript{*} $q<0.05$, \textsuperscript{**} $q<0.01$, \textsuperscript{***} $q<0.001$).}
	\label{tab:relevance_conditioned_shift}
\end{table}

\FloatBarrier

\section{Additional Model Prediction Results}
\label{sec:appendix_model_prediction_results}

\subsection{Signed Prediction Errors}
\label{sec:appendix_signed_shifts}

Table~\ref{tab:appendix_signed_shift} reports the direction of prediction error relative to human ratings.
Positive values indicate overestimation, while negative values indicate underestimation.
The relevant pattern is comparative: matched real-image contexts often have the most positive signed errors within a model/subset row, indicating an upward shift relative to the other visual contexts.
For some rows, this shift produces positive overestimation. For others, predictions remain below the human ratings in every context, but the matched real-image contexts are closest to zero.
This pattern is strongest for imagery and is also visible for abstract words, supporting the main RMSE results: matched real-image contexts can push ratings upward when the requested judgment should remain lexical.

\subsection{Spearman Rank Correlations}
\label{sec:appendix_correlation_tables}

The main results use RMSE because our primary question is whether visual context moves predictions away from the human rating scale.
Table~\ref{tab:appendix_gen_corr} adds the corresponding Spearman rank correlations for the same models, datasets, and visual contexts.
This metric asks a different question: whether the relative ordering of words is preserved, even if the predicted scale shifts.
We therefore treat Spearman as complementary, especially because the rating scales are dense. Small swaps among similarly rated words can affect rank-based scores, and Table~\ref{tab:appendix_dataset_stats} documents this close spacing through the median gap between adjacent distinct ratings.

The Spearman results mostly support the RMSE analysis, but with one useful distinction.
\emph{None} and \emph{White} remain strong overall, consistent with the main finding that removing or minimizing visual content generally preserves lexical judgments.
At the same time, \emph{Noise} is more damaging under Spearman than under RMSE for several models, suggesting that predictions can stay near the correct scale while losing some item-level ordering.
This separation between calibration and rank order also appears in the human validation study (Appendix~\ref{sec:appendix_human_validation}), where rank agreement can improve even when RMSE and mean signed error show a larger real-image rating shift.
In sum, the Spearman results refine the RMSE picture: they show when item-level ordering and scale calibration diverge, while leaving intact the main conclusion that visual context is associated with substantial shifts in model predictions for lexical judgments.

\section{Additional Probing Results}
\label{sec:appendix_additional_results}

\begin{figure*}[t]
	\centering
	\begin{minipage}{0.98\textwidth}
		\centering
		\includegraphics[width=\linewidth,height=0.18\textheight,keepaspectratio]{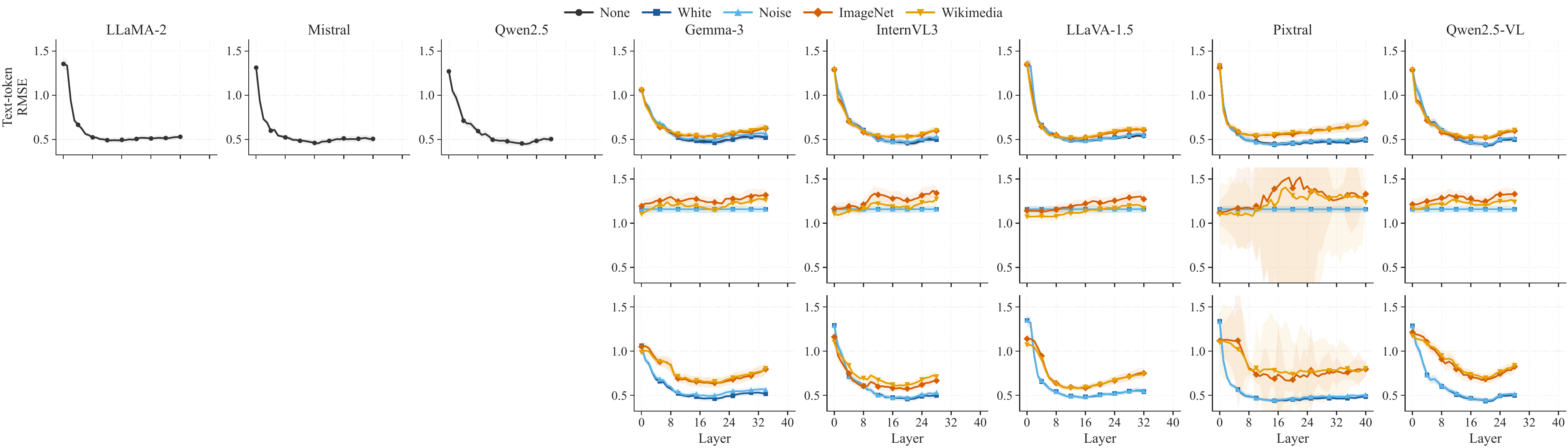}
		\caption*{(a) RMSE}
	\end{minipage}

	\vspace{0.4em}
	\begin{minipage}{0.98\textwidth}
		\centering
		\includegraphics[width=\linewidth,height=0.18\textheight,keepaspectratio]{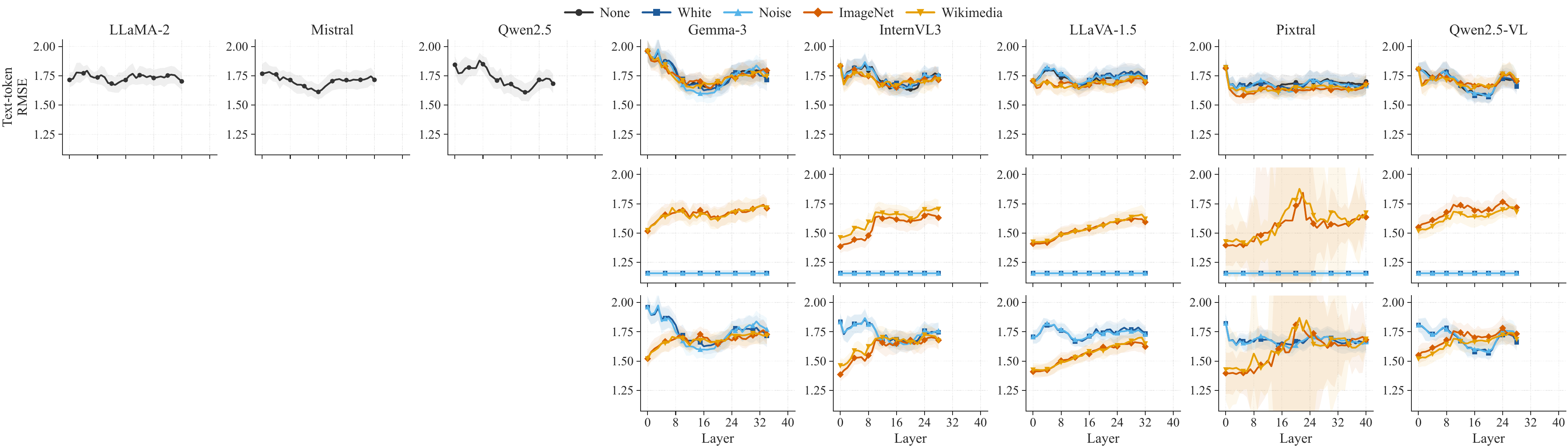}
		\caption*{(b) RMSE, permuted labels}
	\end{minipage}

	\vspace{0.4em}
	\begin{minipage}{0.98\textwidth}
		\centering
		\includegraphics[width=\linewidth,height=0.18\textheight,keepaspectratio]{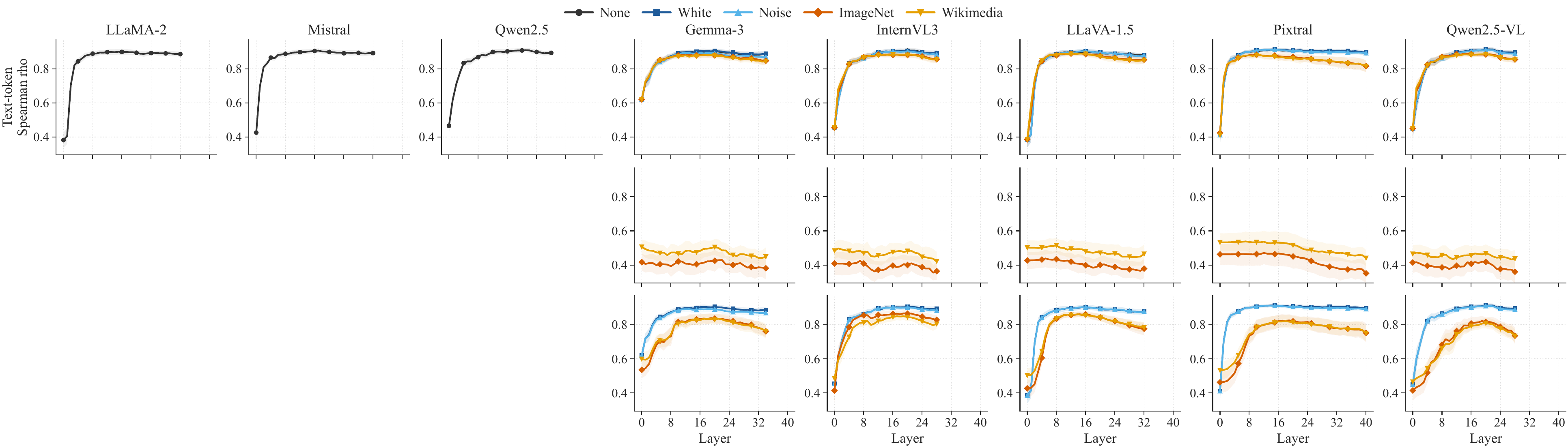}
		\caption*{(c) Spearman}
	\end{minipage}

	\vspace{0.4em}
	\begin{minipage}{0.98\textwidth}
		\centering
		\includegraphics[width=\linewidth,height=0.18\textheight,keepaspectratio]{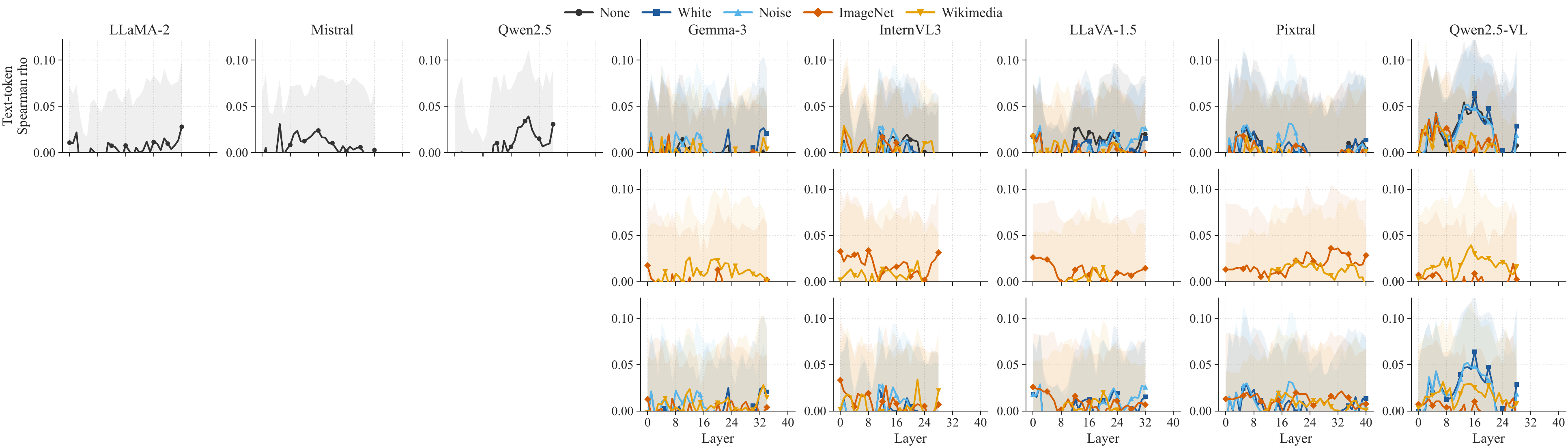}
		\caption*{(d) Spearman, permuted labels}
	\end{minipage}
	\caption{Layer-wise ridge probing on MT40k concreteness. Panels show RMSE and Spearman, each with a permuted-label baseline. VLM probes use text-token, image-token, and combined representations. Bands show standard deviations across folds and prompt templates.}
	\label{fig:appendix_probing_mt40k_ridge_grouped}
\end{figure*}

\begin{figure*}[t]
	\centering
	\begin{minipage}{0.98\textwidth}
		\centering
		\includegraphics[width=\linewidth,height=0.18\textheight,keepaspectratio]{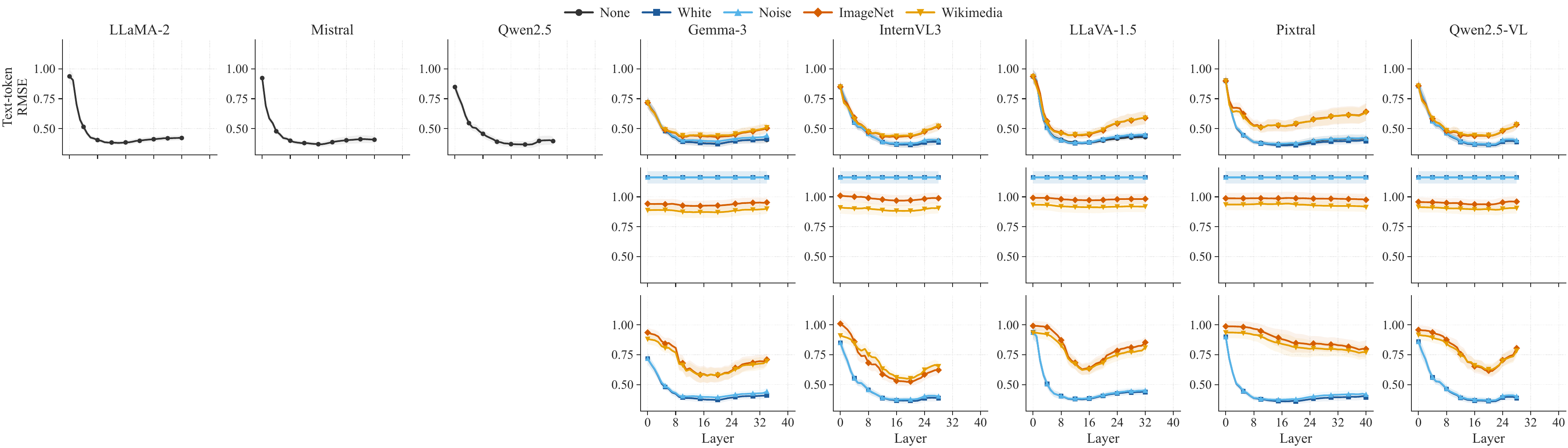}
		\caption*{(a) RMSE}
	\end{minipage}

	\vspace{0.4em}
	\begin{minipage}{0.98\textwidth}
		\centering
		\includegraphics[width=\linewidth,height=0.18\textheight,keepaspectratio]{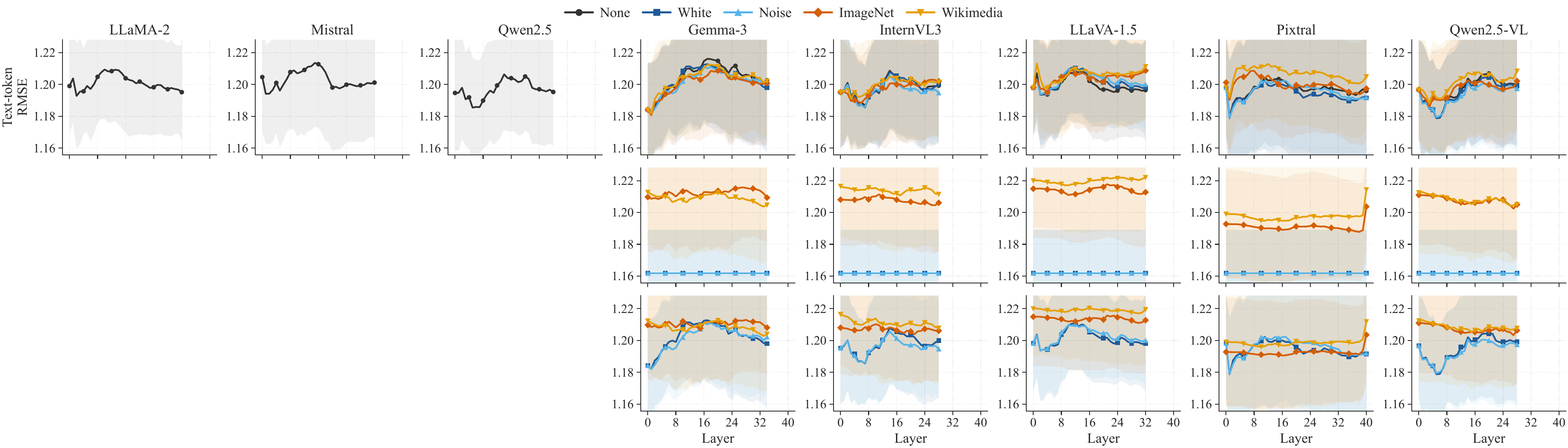}
		\caption*{(b) RMSE, permuted labels}
	\end{minipage}

	\vspace{0.4em}
	\begin{minipage}{0.98\textwidth}
		\centering
		\includegraphics[width=\linewidth,height=0.18\textheight,keepaspectratio]{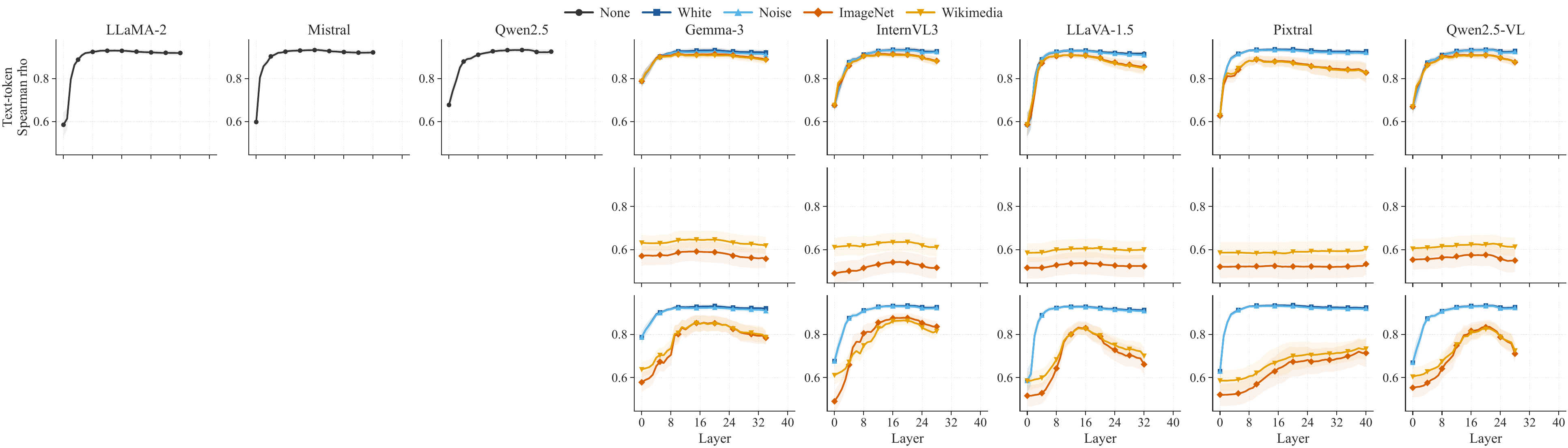}
		\caption*{(c) Spearman}
	\end{minipage}

	\vspace{0.4em}
	\begin{minipage}{0.98\textwidth}
		\centering
		\includegraphics[width=\linewidth,height=0.18\textheight,keepaspectratio]{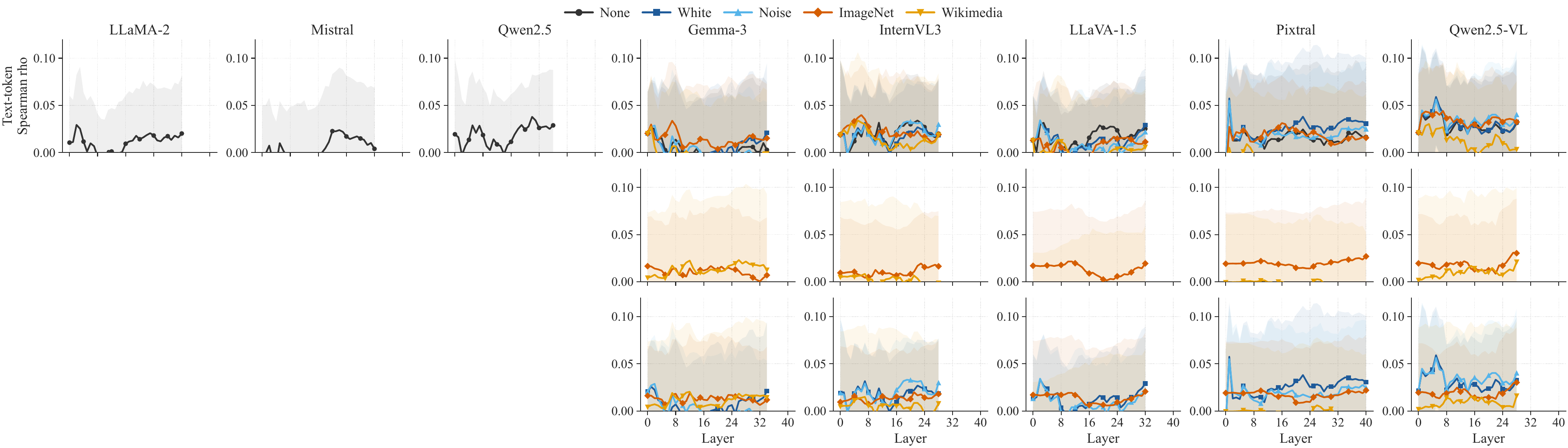}
		\caption*{(d) Spearman, permuted labels}
	\end{minipage}
	\caption{Layer-wise SVM probing on MT40k concreteness. Panels show RMSE and Spearman, each with a permuted-label baseline. VLM probes use text-token, image-token, and combined representations. Bands show standard deviations across folds and prompt templates.}
	\label{fig:appendix_probing_mt40k_svm_grouped}
\end{figure*}

\begin{figure*}[t]
	\centering
	\begin{minipage}{0.98\textwidth}
		\centering
		\includegraphics[width=\linewidth,height=0.18\textheight,keepaspectratio]{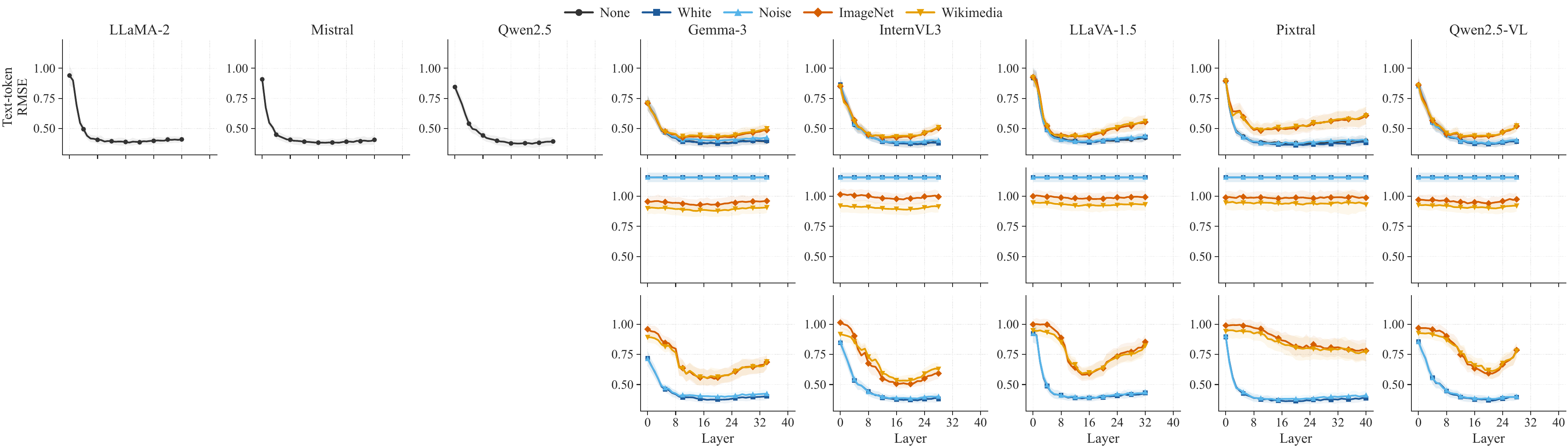}
		\caption*{(a) RMSE}
	\end{minipage}

	\vspace{0.4em}
	\begin{minipage}{0.98\textwidth}
		\centering
		\includegraphics[width=\linewidth,height=0.18\textheight,keepaspectratio]{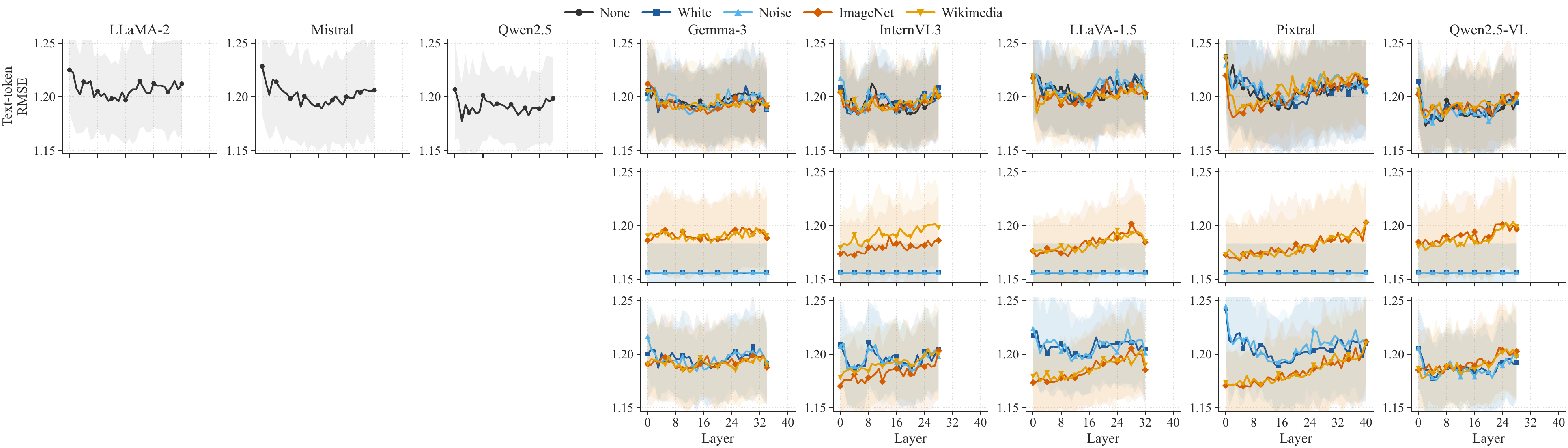}
		\caption*{(b) RMSE, permuted labels}
	\end{minipage}

	\vspace{0.4em}
	\begin{minipage}{0.98\textwidth}
		\centering
		\includegraphics[width=\linewidth,height=0.18\textheight,keepaspectratio]{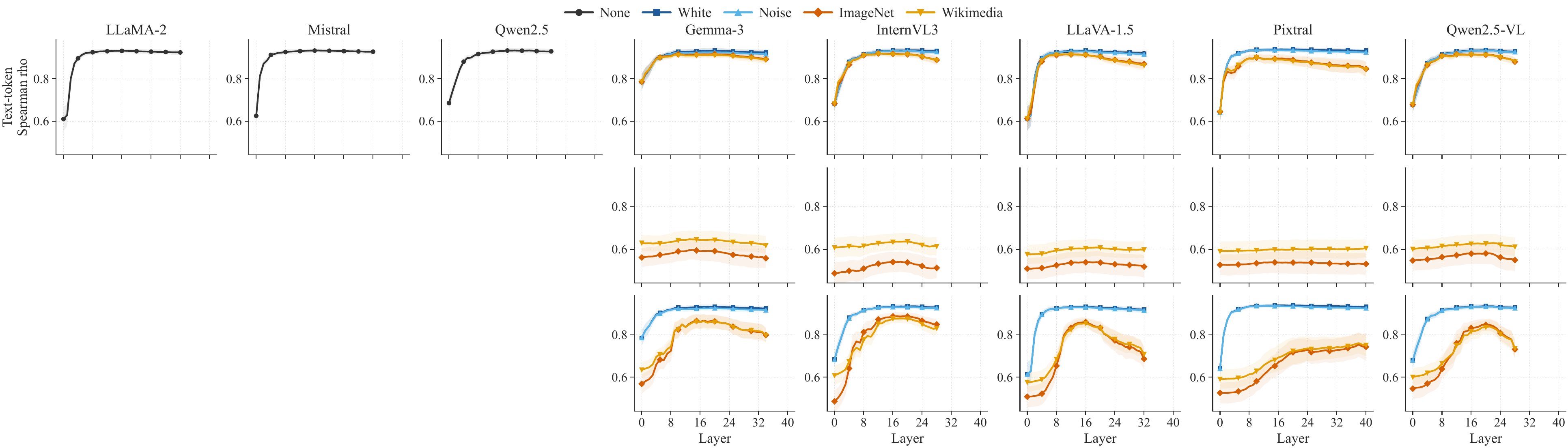}
		\caption*{(c) Spearman}
	\end{minipage}

	\vspace{0.4em}
	\begin{minipage}{0.98\textwidth}
		\centering
		\includegraphics[width=\linewidth,height=0.18\textheight,keepaspectratio]{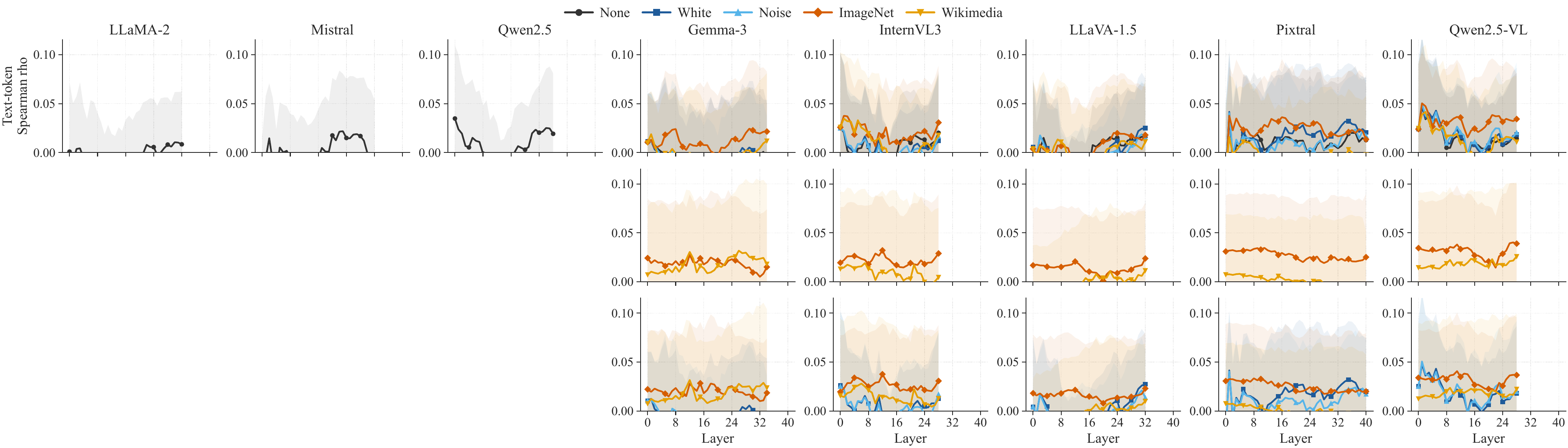}
		\caption*{(d) Spearman, permuted labels}
	\end{minipage}
	\caption{Layer-wise MLP probing on MT40k concreteness. Panels show RMSE and Spearman, each with a permuted-label baseline. VLM probes use text-token, image-token, and combined representations. Bands show standard deviations across folds and prompt templates.}
	\label{fig:appendix_probing_mt40k_mlp_grouped}
\end{figure*}

\begin{figure*}[t]
	\centering
	\begin{minipage}{0.98\textwidth}
		\centering
		\includegraphics[width=\linewidth,height=0.18\textheight,keepaspectratio]{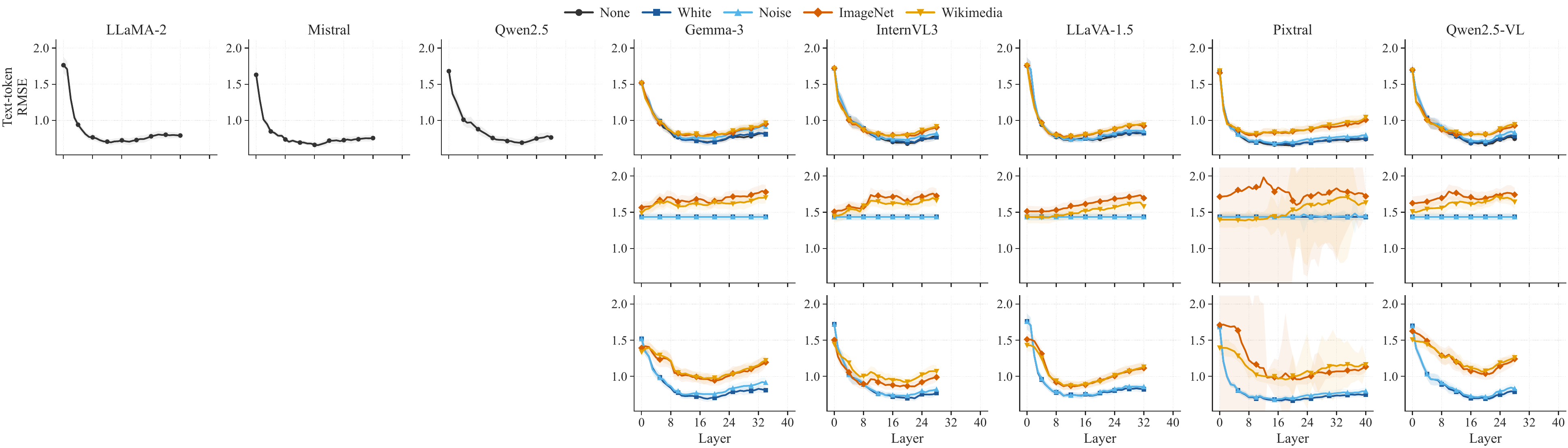}
		\caption*{(a) RMSE}
	\end{minipage}

	\vspace{0.4em}
	\begin{minipage}{0.98\textwidth}
		\centering
		\includegraphics[width=\linewidth,height=0.18\textheight,keepaspectratio]{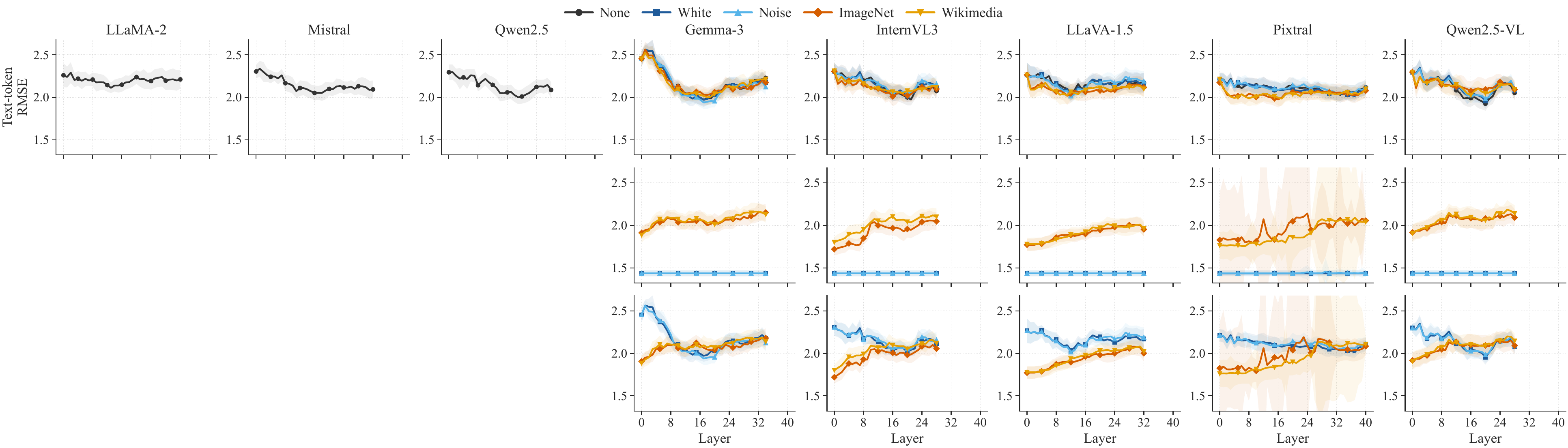}
		\caption*{(b) RMSE, permuted labels}
	\end{minipage}

	\vspace{0.4em}
	\begin{minipage}{0.98\textwidth}
		\centering
		\includegraphics[width=\linewidth,height=0.18\textheight,keepaspectratio]{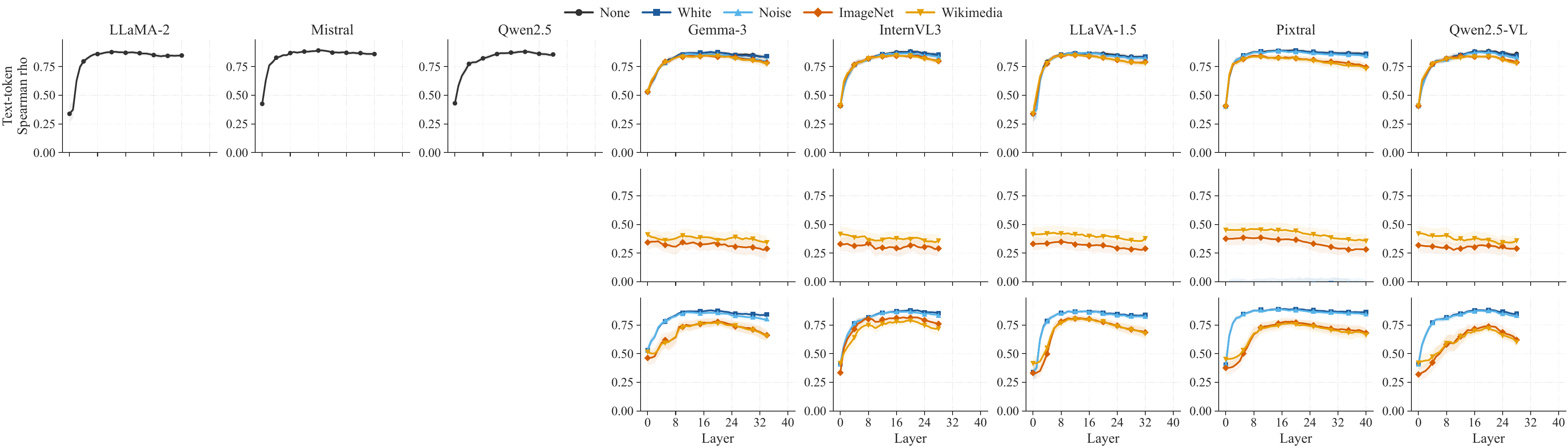}
		\caption*{(c) Spearman}
	\end{minipage}

	\vspace{0.4em}
	\begin{minipage}{0.98\textwidth}
		\centering
		\includegraphics[width=\linewidth,height=0.18\textheight,keepaspectratio]{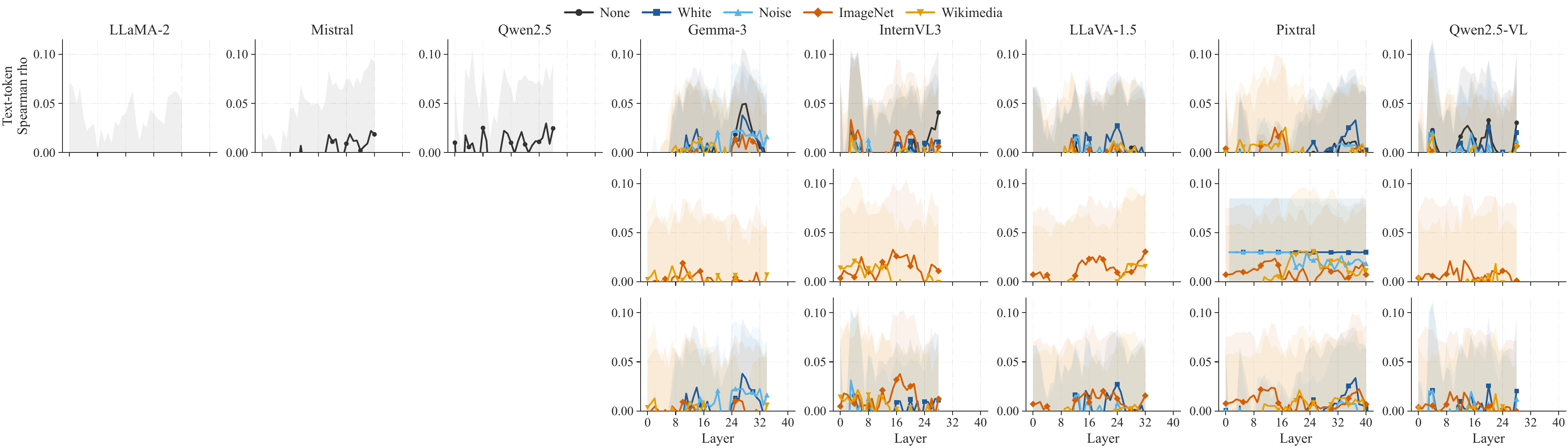}
		\caption*{(d) Spearman, permuted labels}
	\end{minipage}
	\caption{Layer-wise ridge probing on CP2004B imagery. Panels show RMSE and Spearman, each with a permuted-label baseline. VLM probes use text-token, image-token, and combined representations. Bands show standard deviations across folds and prompt templates.}
	\label{fig:appendix_probing_cp2004b_ridge_grouped}
\end{figure*}

\begin{figure*}[t]
	\centering
	\begin{minipage}{0.98\textwidth}
		\centering
		\includegraphics[width=\linewidth,height=0.18\textheight,keepaspectratio]{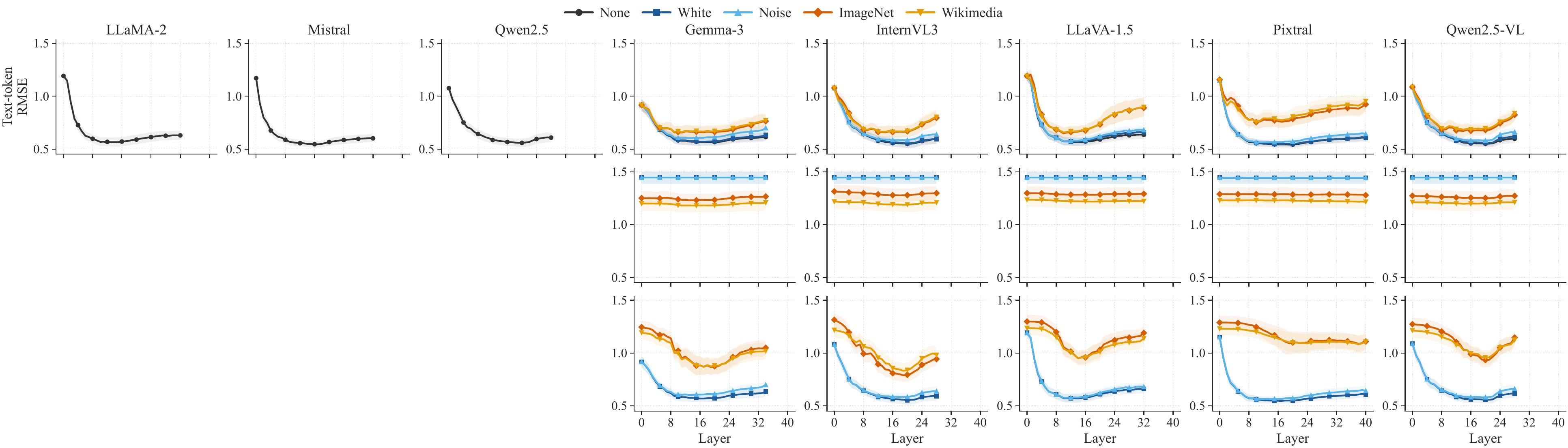}
		\caption*{(a) RMSE}
	\end{minipage}

	\vspace{0.4em}
	\begin{minipage}{0.98\textwidth}
		\centering
		\includegraphics[width=\linewidth,height=0.18\textheight,keepaspectratio]{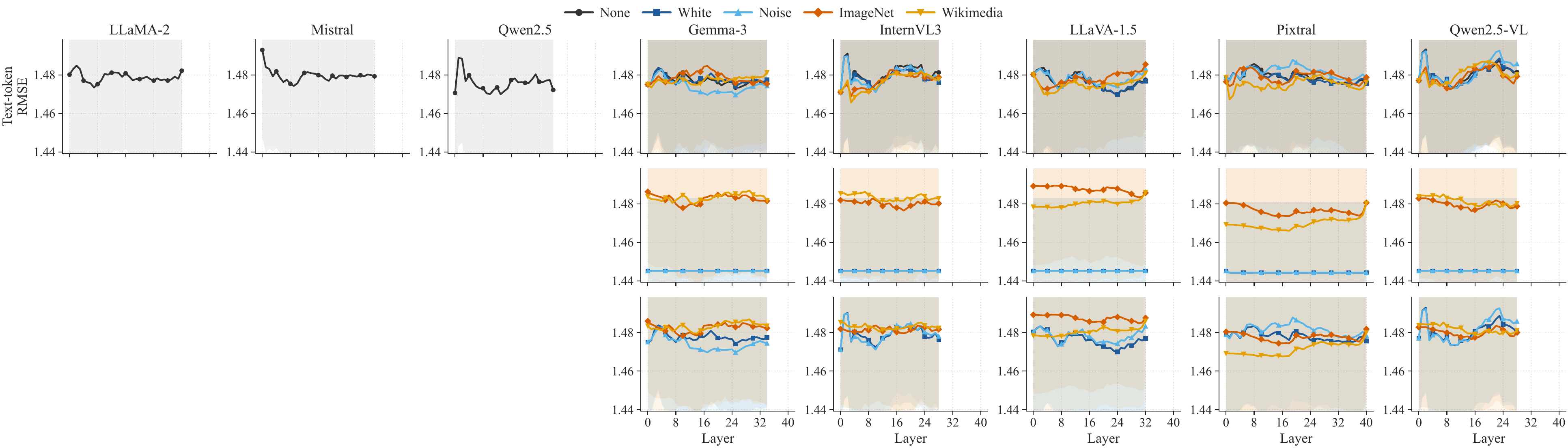}
		\caption*{(b) RMSE, permuted labels}
	\end{minipage}

	\vspace{0.4em}
	\begin{minipage}{0.98\textwidth}
		\centering
		\includegraphics[width=\linewidth,height=0.18\textheight,keepaspectratio]{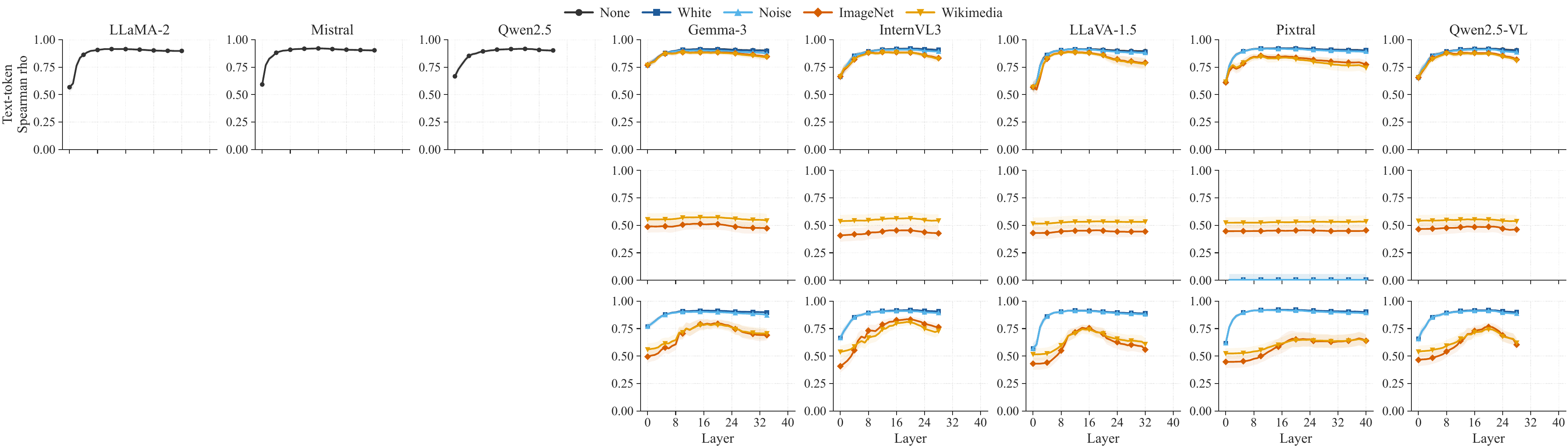}
		\caption*{(c) Spearman}
	\end{minipage}

	\vspace{0.4em}
	\begin{minipage}{0.98\textwidth}
		\centering
		\includegraphics[width=\linewidth,height=0.18\textheight,keepaspectratio]{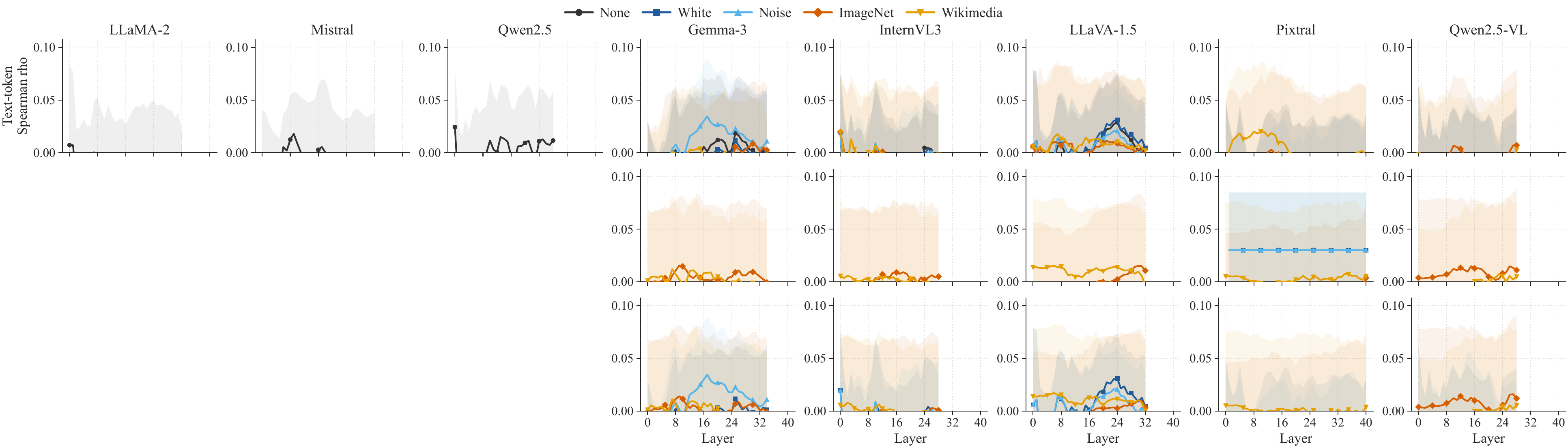}
		\caption*{(d) Spearman, permuted labels}
	\end{minipage}
	\caption{Layer-wise SVM probing on CP2004B imagery. Panels show RMSE and Spearman, each with a permuted-label baseline. VLM probes use text-token, image-token, and combined representations. Bands show standard deviations across folds and prompt templates.}
	\label{fig:appendix_probing_cp2004b_svm_grouped}
\end{figure*}

\begin{figure*}[t]
	\centering
	\begin{minipage}{0.98\textwidth}
		\centering
		\includegraphics[width=\linewidth,height=0.18\textheight,keepaspectratio]{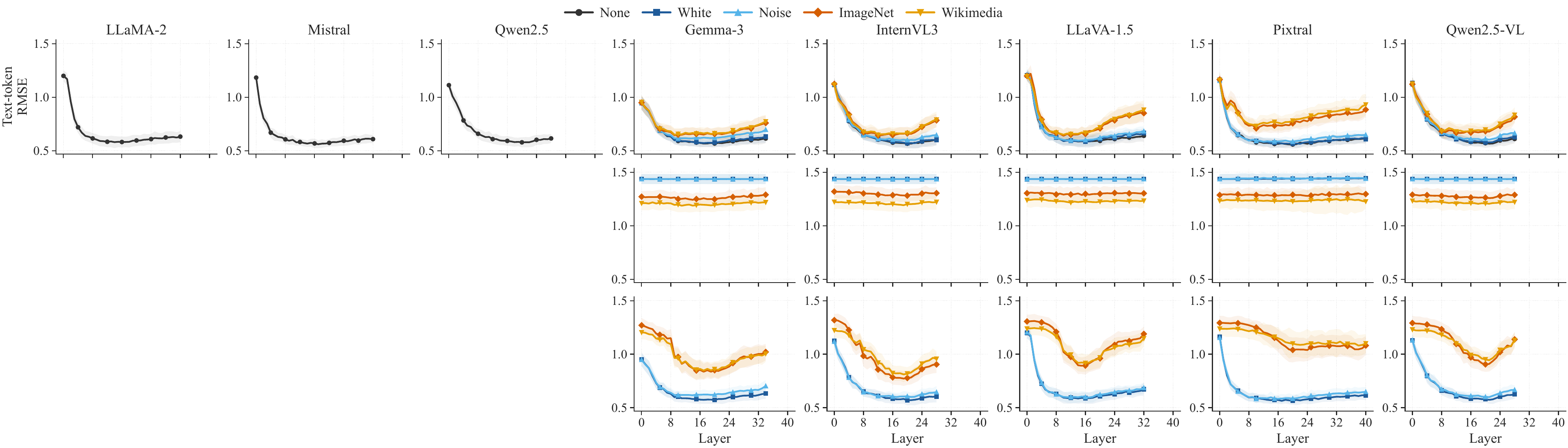}
		\caption*{(a) RMSE}
	\end{minipage}

	\vspace{0.4em}
	\begin{minipage}{0.98\textwidth}
		\centering
		\includegraphics[width=\linewidth,height=0.18\textheight,keepaspectratio]{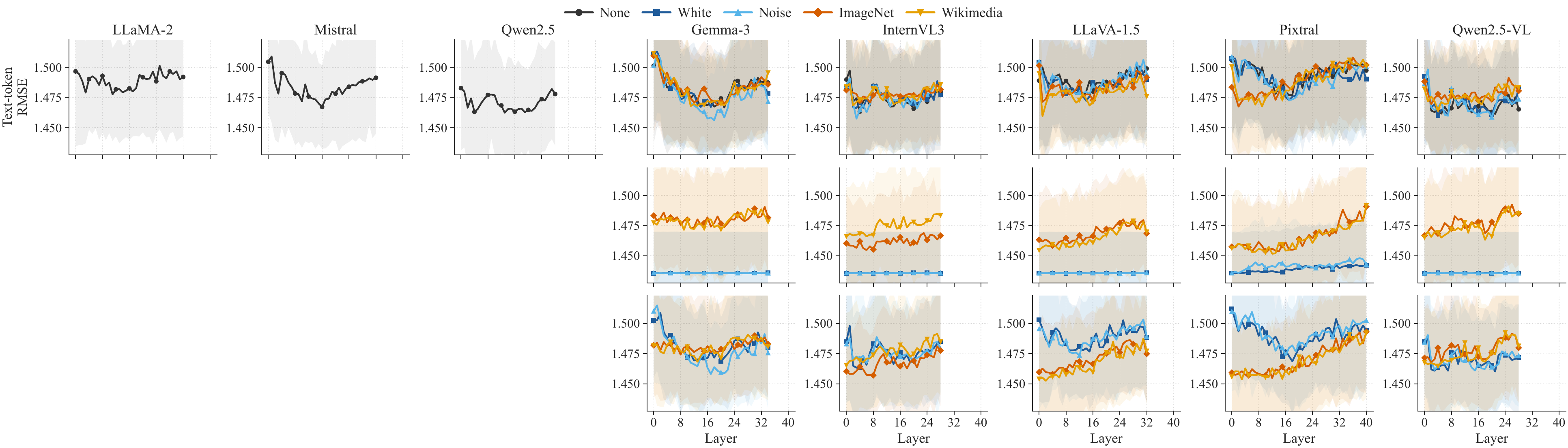}
		\caption*{(b) RMSE, permuted labels}
	\end{minipage}

	\vspace{0.4em}
	\begin{minipage}{0.98\textwidth}
		\centering
		\includegraphics[width=\linewidth,height=0.18\textheight,keepaspectratio]{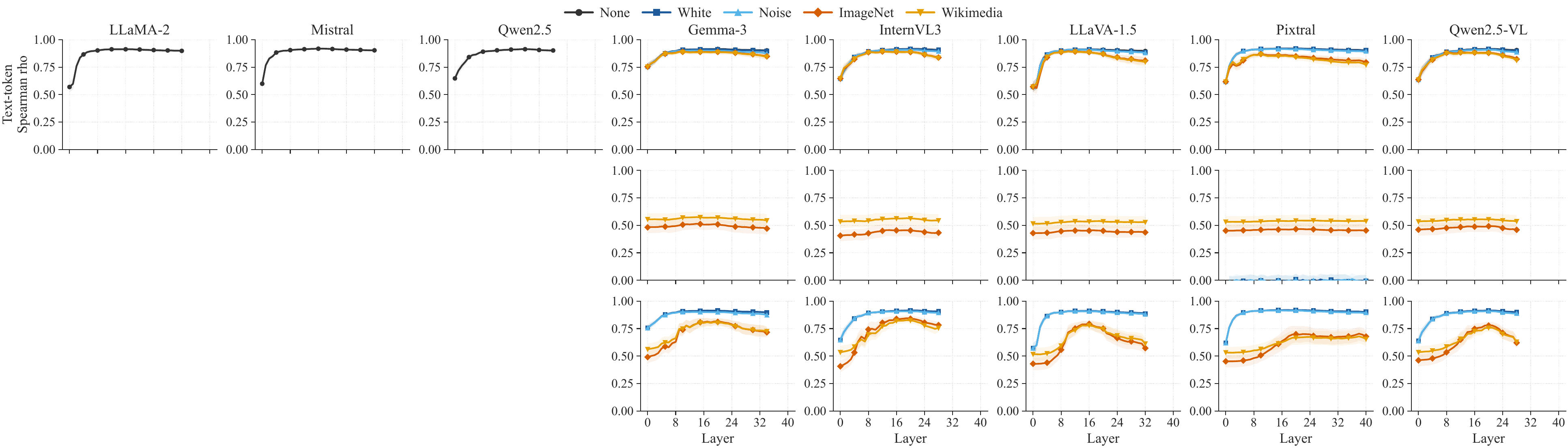}
		\caption*{(c) Spearman}
	\end{minipage}

	\vspace{0.4em}
	\begin{minipage}{0.98\textwidth}
		\centering
		\includegraphics[width=\linewidth,height=0.18\textheight,keepaspectratio]{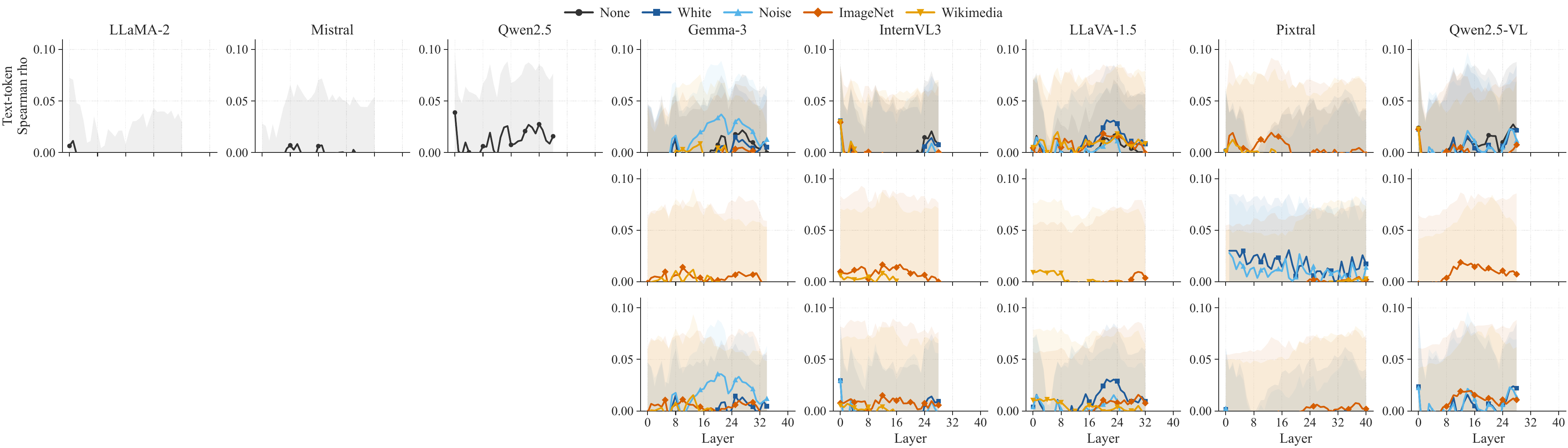}
		\caption*{(d) Spearman, permuted labels}
	\end{minipage}
	\caption{Layer-wise MLP probing on CP2004B imagery. Panels show RMSE and Spearman, each with a permuted-label baseline. VLM probes use text-token, image-token, and combined representations. Bands show standard deviations across folds and prompt templates.}
	\label{fig:appendix_probing_cp2004b_mlp_grouped}
\end{figure*}

This section provides additional probing results corresponding to the analyses summarized in the main text.
For each dataset (MT40k and CP2004B), we report layer-wise probing curves for all three regressors (ridge, SVM, and MLP) under RMSE and Spearman rank correlation, together with matched permuted-label baselines.
Figures~\ref{fig:appendix_probing_mt40k_ridge_grouped}--\ref{fig:appendix_probing_mt40k_mlp_grouped} show the MT40k results, and Figures~\ref{fig:appendix_probing_cp2004b_ridge_grouped}--\ref{fig:appendix_probing_cp2004b_mlp_grouped} show the CP2004B results.

\section{Additional Mitigation Analyses}
\label{sec:appendix_additional_mitigation}

\subsection{Additional Model Results}
\label{sec:appendix_additional_mitigation_models}

\begin{figure*}[t]
	\centering
	\begin{minipage}{0.48\textwidth}
		\centering
		\includegraphics[width=\linewidth]{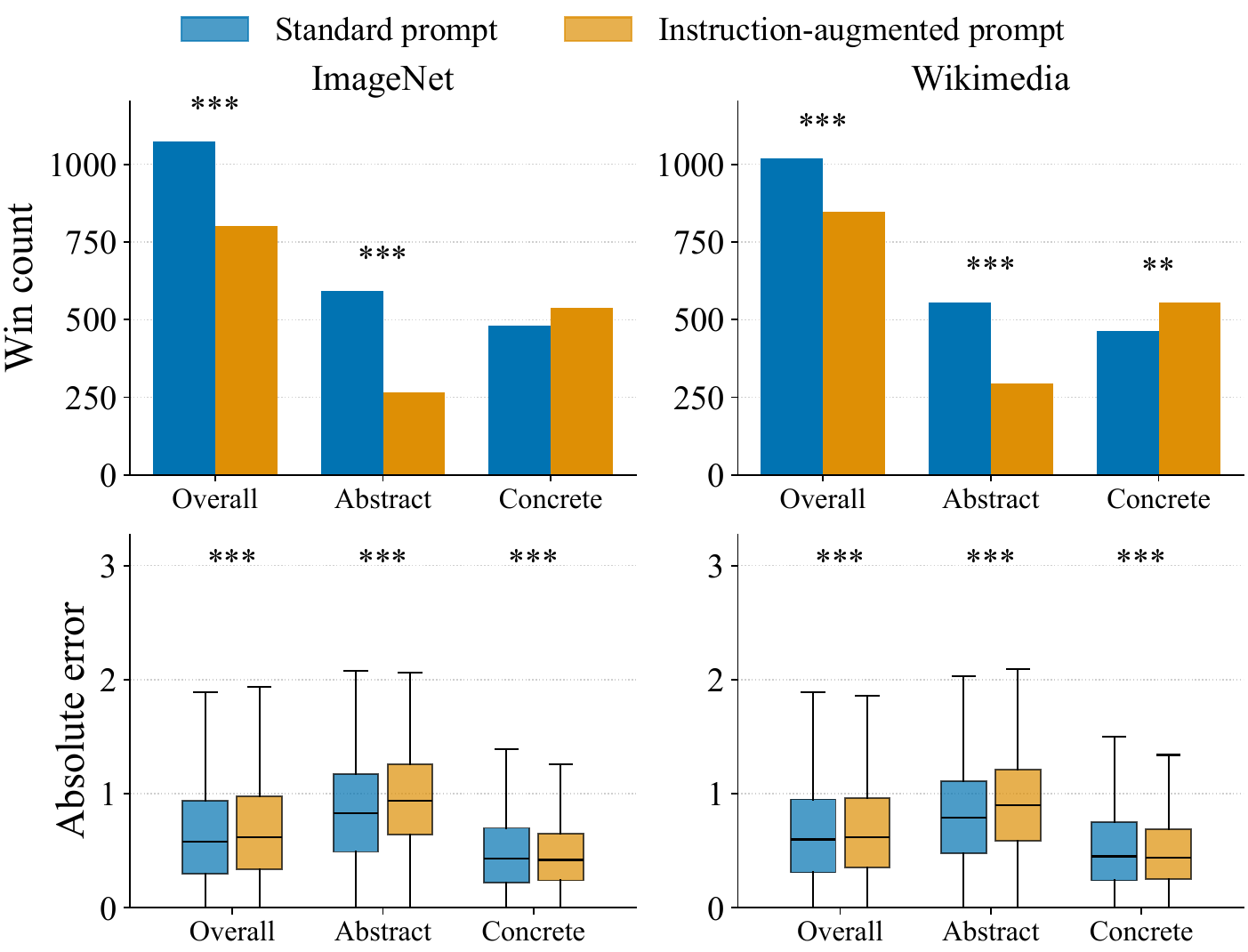}
		\caption*{(a) Gemma 3}
	\end{minipage}
	\hfill
	\begin{minipage}{0.48\textwidth}
		\centering
		\includegraphics[width=\linewidth]{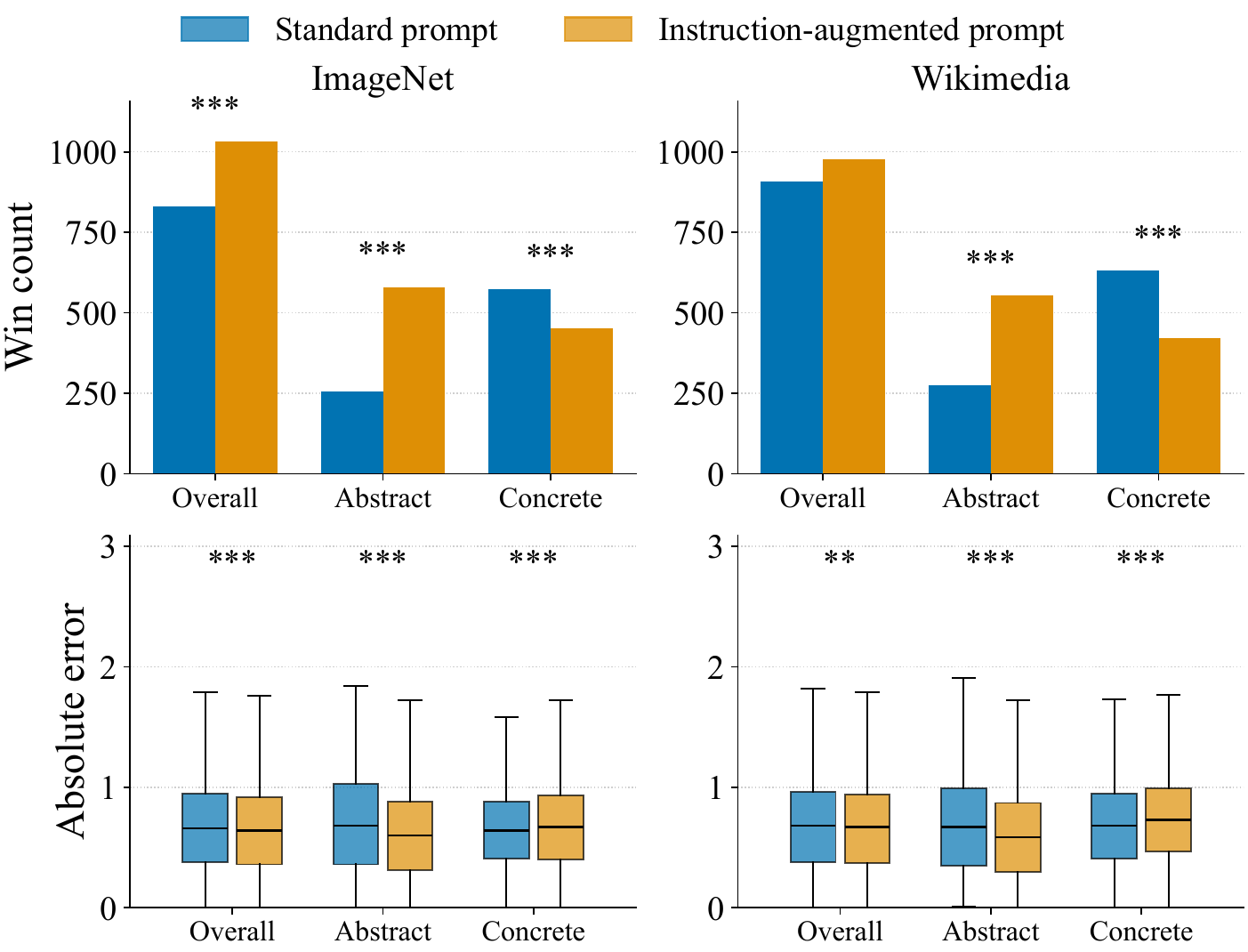}
		\caption*{(b) InternVL3}
	\end{minipage}

	\vspace{0.8em}
	\begin{minipage}{0.48\textwidth}
		\centering
		\includegraphics[width=\linewidth]{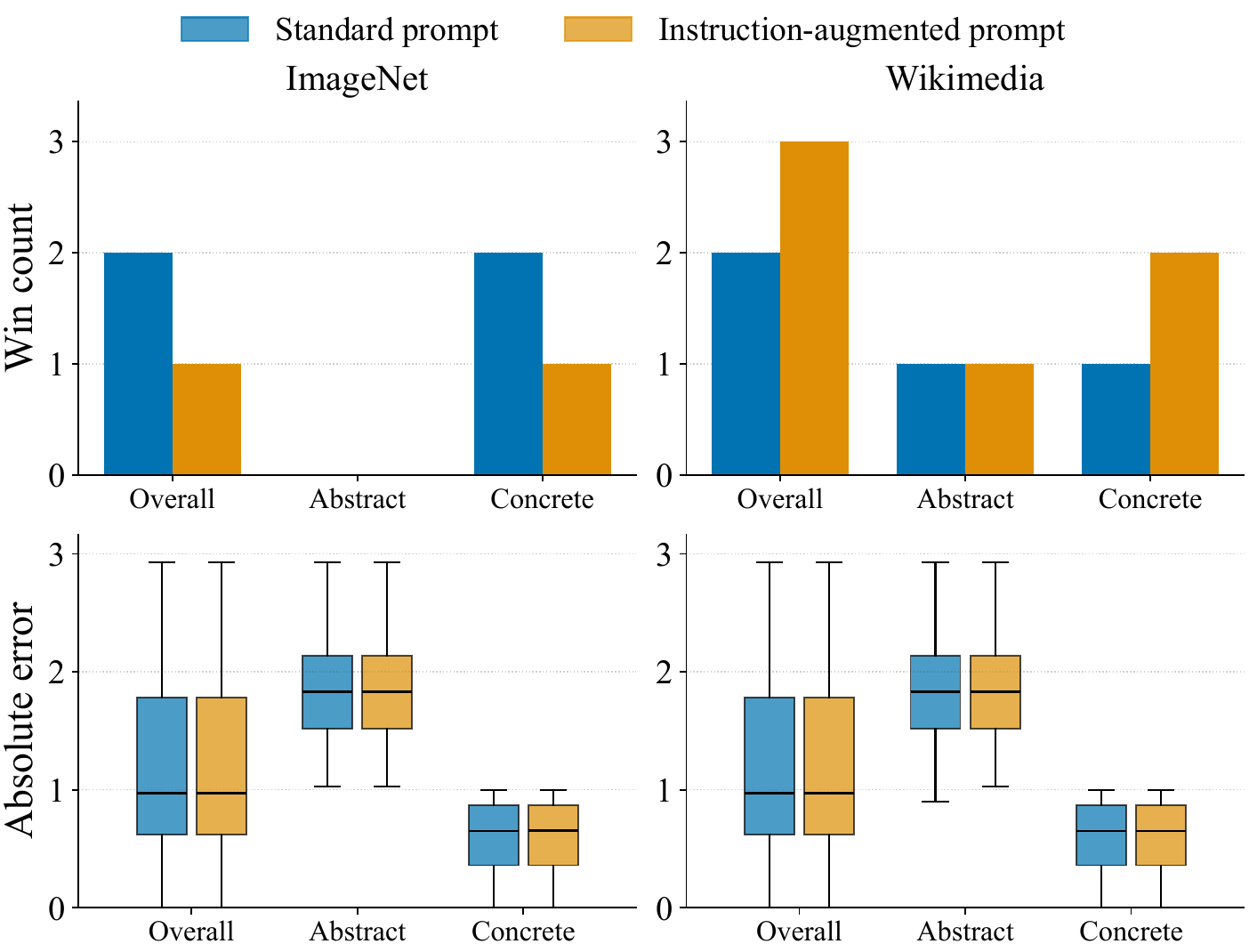}
		\caption*{(c) LLaVA 1.5}
	\end{minipage}
	\hfill
	\begin{minipage}{0.48\textwidth}
		\centering
		\includegraphics[width=\linewidth]{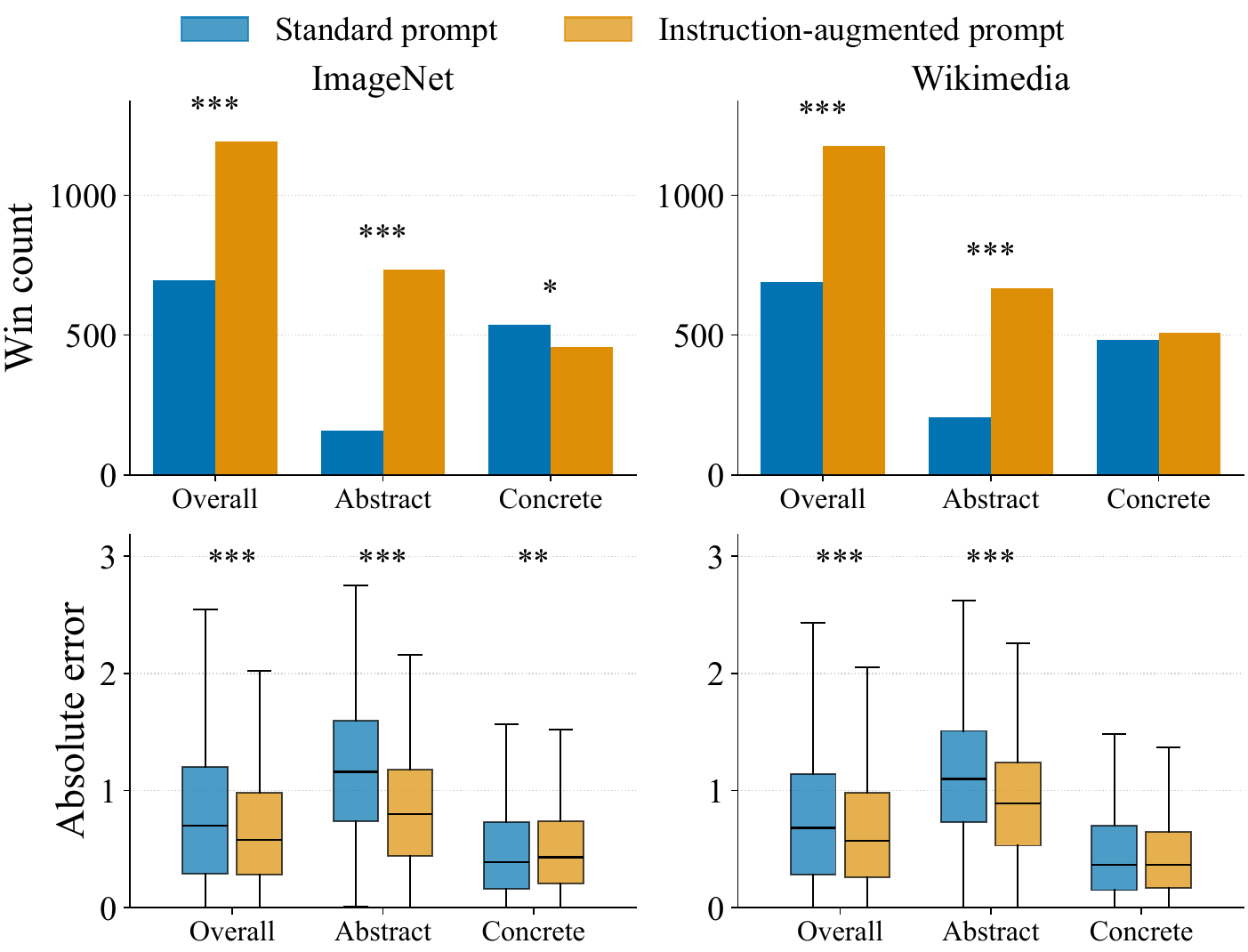}
		\caption*{(d) Pixtral}
	\end{minipage}
	\caption{Additional MT40k mitigation results for real-image contexts. Each subfigure compares standard and instruction-augmented prompting for one VLM. Top rows show non-tied win counts. Bottom rows show absolute-error distributions, overall and by concreteness subset. Superscripts mark significance as in Figure~\ref{fig:mitigation}.}
	\label{fig:appendix_mitigation_extra_mt40k}
\end{figure*}

\begin{figure*}[t]
	\centering
	\begin{minipage}{0.48\textwidth}
		\centering
		\includegraphics[width=\linewidth]{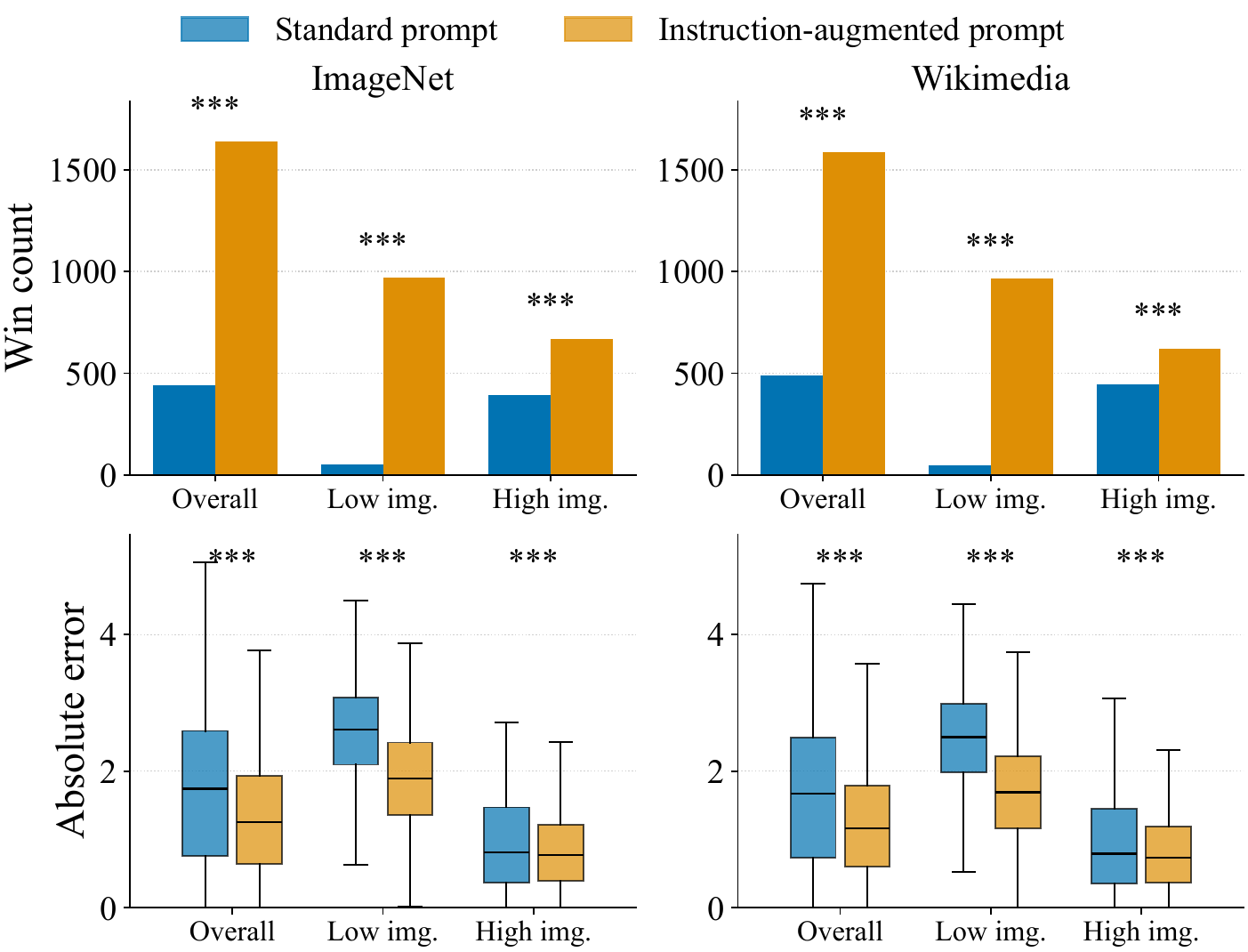}
		\caption*{(a) Gemma 3}
	\end{minipage}
	\hfill
	\begin{minipage}{0.48\textwidth}
		\centering
		\includegraphics[width=\linewidth]{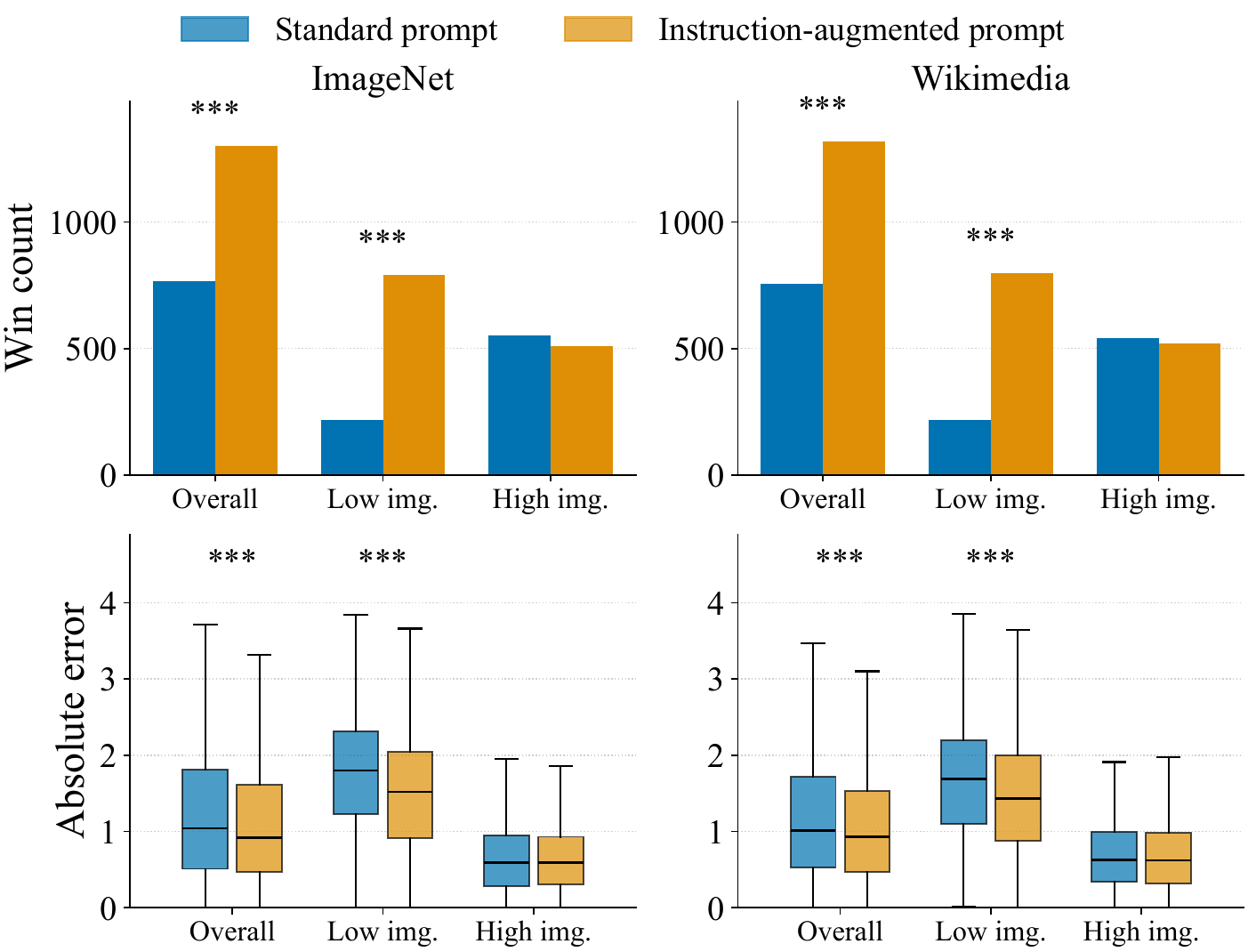}
		\caption*{(b) InternVL3}
	\end{minipage}

	\vspace{0.8em}
	\begin{minipage}{0.48\textwidth}
		\centering
		\includegraphics[width=\linewidth]{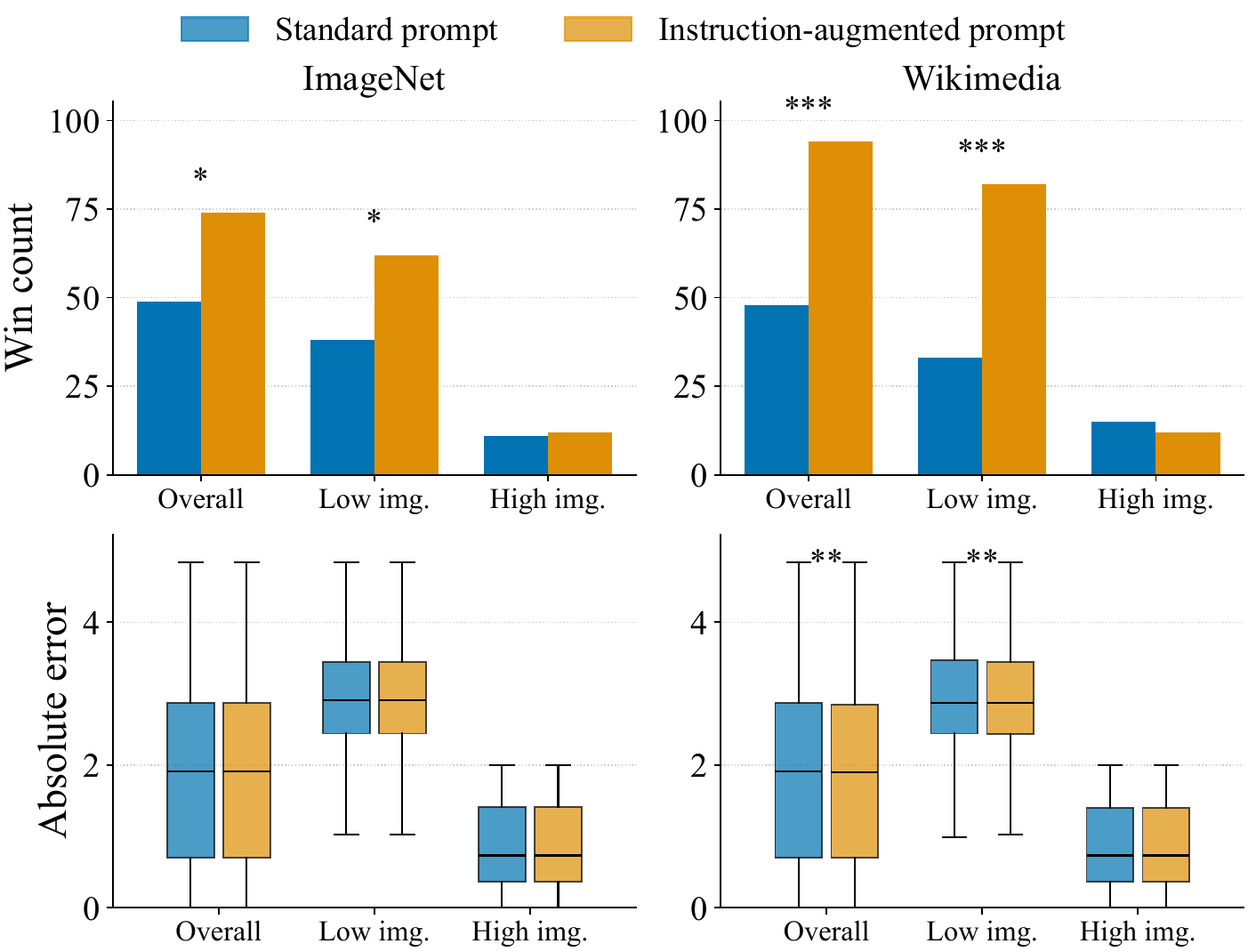}
		\caption*{(c) LLaVA 1.5}
	\end{minipage}
	\hfill
	\begin{minipage}{0.48\textwidth}
		\centering
		\includegraphics[width=\linewidth]{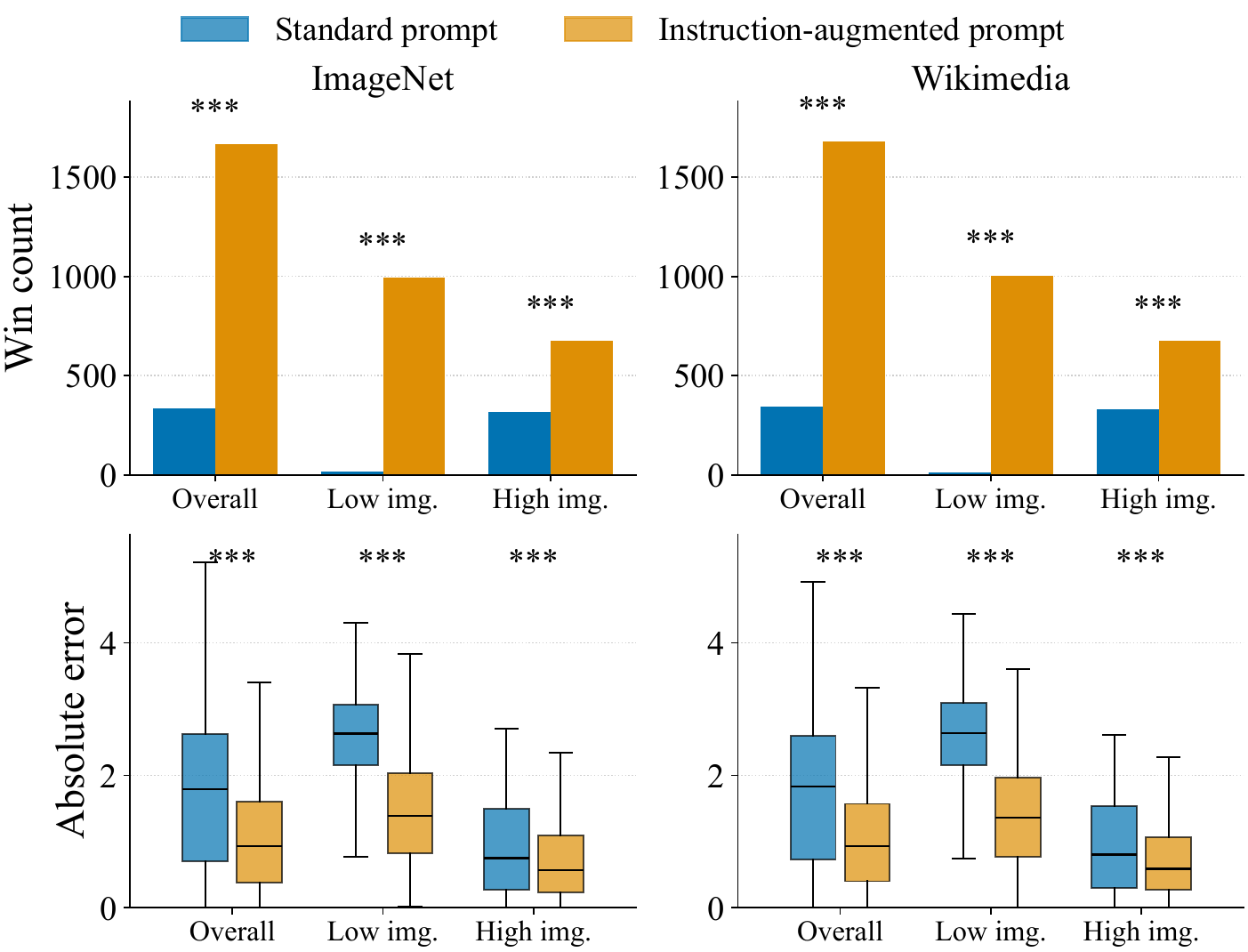}
		\caption*{(d) Pixtral}
	\end{minipage}
	\caption{Additional CP2004B mitigation results for real-image contexts. Each subfigure compares standard and instruction-augmented prompting for one VLM. Top rows show non-tied win counts. Bottom rows show absolute-error distributions, overall and by imagery subset. Superscripts mark significance as in Figure~\ref{fig:mitigation}.}
	\label{fig:appendix_mitigation_extra_cp2004b}
\end{figure*}

This section provides additional mitigation results for the other evaluated VLMs under the two real-image contexts.
As in the main experiment, the intervention prepends the same sentence to each of the five original rating prompt variants.
As in Figure~\ref{fig:mitigation}, non-tied win counts exclude tied absolute errors, while the lower panels show absolute-error distributions over all valid paired predictions. Superscripts use exact sign tests for non-tied win counts and paired sign-flip tests for absolute-error differences.
Figure~\ref{fig:appendix_mitigation_extra_mt40k} reports MT40k results, and Figure~\ref{fig:appendix_mitigation_extra_cp2004b} reports CP2004B results.

Overall, the additional models support the main mitigation result: making image irrelevance explicit can reduce errors, most consistently on CP2004B low-imagery words.
The effect is less uniform on MT40k, where gains depend more on the model and subset, so the intervention is best viewed as a lightweight, selective mitigation rather than a universal fix.

\subsection{Mitigation Prompt Wording Robustness}
\label{sec:appendix_mitigation_prompt_variants}

To test whether the Qwen2.5-VL mitigation results depend on instruction wording, we keep the model, visual contexts, and original rating prompt variants fixed while varying only the added instruction.
Predictions are averaged across the five original rating prompt variants, as in the main experiments.
Alongside the instruction used in the main experiment, we test four related variants:
\begin{enumerate}[leftmargin=*,itemsep=0pt,topsep=2pt]
	\item \textit{Ignore visual details that are unrelated to the target word, and rate the word based on its meaning.}
	\item \textit{Use the image only if it helps with the target word; otherwise, rely on the word itself.}
	\item \textit{Do not let irrelevant image content influence your rating of the word.}
	\item \textit{If the image and word seem mismatched, prioritize the word when assigning the rating.}
\end{enumerate}
Table~\ref{tab:appendix_mitigation_prompt_variants_qwen_vl} summarizes the results.
\emph{Win margin} is the mitigation win count minus the standard-prompt win count, divided by the number of non-tied pairs and reported in percentage points.
\emph{MAE reduction} is the mean item-level decrease in absolute error, \(|\hat{y}_{\text{standard}}-y|-|\hat{y}_{\text{mitigation}}-y|\), so positive values favor the mitigation prompt.

On MT40k, the variants reproduce the subset-level tradeoff seen in the main mitigation result.
The average rows show consistent gains for abstract words in both real-image contexts, while concrete words move in the opposite direction.
The overall MT40k averages are therefore negative, indicating that the wording variants improve the abstract subset but do not provide a uniform benefit across the dataset.

CP2004B shows broader support for mitigation.
The average rows are positive overall, with the clearest improvements on low-imagery words.
High-imagery words show little average change, suggesting that the mitigation mainly helps where irrelevant visual evidence is most disruptive.
Overall, these results suggest that the gains are robust across prompt wordings for the cases most vulnerable to irrelevant visual evidence, but the benefit remains selective rather than universal.

\begin{table*}[!t]
	\centering
	{
		\scriptsize
		\setlength{\ctxcolwidth}{3.9em}
		\setlength{\tabcolsep}{2pt}
		\renewcommand{\arraystretch}{0.86}
		\begin{tabular*}{\textwidth}{@{\extracolsep{\fill}}l P R r r P R r r P R r r@{}}
			\toprule
			& \multicolumn{4}{c}{All} & \multicolumn{4}{c}{Low imagery} & \multicolumn{4}{c}{High imagery} \\
			\cmidrule(lr){2-5} \cmidrule(lr){6-9} \cmidrule(lr){10-13}
			Source & \placeholderhead{White} & \realhead{ImageNet} & $\Delta$ & $\Delta-\Delta_H$ & \placeholderhead{White} & \realhead{ImageNet} & $\Delta$ & $\Delta-\Delta_H$ & \placeholderhead{White} & \realhead{ImageNet} & $\Delta$ & $\Delta-\Delta_H$ \\
			\midrule
			Human & \ctxmark{human-rmse-all-white-t}{\textbf{1.15}} & \ctxmark{human-rmse-all-image-t}{\textbf{1.20}} & \textbf{+0.04} & -- & \ctxmark{human-rmse-low-white-t}{1.24} & \ctxmark{human-rmse-low-image-t}{\textbf{1.43}} & \textsuperscript{**}\textbf{+0.19} & -- & \ctxmark{human-rmse-high-white-t}{1.07} & \ctxmark{human-rmse-high-image-t}{0.93} & \textsuperscript{*}-0.14 & -- \\
			\midrule
			Gemma 3 & 1.41 & 2.03 & \textsuperscript{***}+0.62 & \textsuperscript{***}+0.58 & 1.58 & 2.65 & \textsuperscript{***}+1.08 & \textsuperscript{***}+0.89 & 1.23 & \underline{1.17} & -0.06 & \textbf{+0.08} \\
			InternVL3 & 1.20 & 1.43 & \textsuperscript{***}+0.23 & \textsuperscript{***}\textbf{+0.19} & 1.33 & 1.89 & \textsuperscript{***}+0.55 & \textsuperscript{***}+0.36 & 1.06 & \textbf{0.78} & \textsuperscript{***}-0.28 & -0.14 \\
			LLaVA 1.5 & \underline{1.93} & \underline{2.26} & \textsuperscript{***}+0.33 & \textsuperscript{***}+0.29 & \underline{2.61} & \underline{3.05} & \textsuperscript{***}+0.44 & \textsuperscript{***}\textbf{+0.25} & \textbf{0.87} & 1.05 & \textsuperscript{***}\underline{+0.17} & \textsuperscript{***}\underline{+0.32} \\
			Pixtral & 1.36 & 2.02 & \textsuperscript{***}\underline{+0.65} & \textsuperscript{***}\underline{+0.61} & \textbf{1.19} & 2.63 & \textsuperscript{***}\underline{+1.44} & \textsuperscript{***}\underline{+1.25} & \underline{1.51} & 1.16 & \textsuperscript{***}\textbf{-0.35} & \textsuperscript{*}-0.21 \\
			Qwen2.5-VL & \ctxmark{human-rmse-all-white-b}{1.42} & \ctxmark{human-rmse-all-image-b}{1.76} & \textsuperscript{***}+0.34 & \textsuperscript{***}+0.30 & \ctxmark{human-rmse-low-white-b}{1.62} & \ctxmark{human-rmse-low-image-b}{2.32} & \textsuperscript{***}+0.70 & \textsuperscript{***}+0.51 & \ctxmark{human-rmse-high-white-b}{1.19} & \ctxmark{human-rmse-high-image-b}{0.97} & \textsuperscript{**}-0.23 & \textbf{-0.08} \\
			\noalign{\ctxoutline{placeholderctx}{human-rmse-all-white-t}{human-rmse-all-white-b}\ctxoutline{realctx}{human-rmse-all-image-t}{human-rmse-all-image-b}\ctxoutline{placeholderctx}{human-rmse-low-white-t}{human-rmse-low-white-b}\ctxoutline{realctx}{human-rmse-low-image-t}{human-rmse-low-image-b}\ctxoutline{placeholderctx}{human-rmse-high-white-t}{human-rmse-high-white-b}\ctxoutline{realctx}{human-rmse-high-image-t}{human-rmse-high-image-b}}
			\bottomrule
		\end{tabular*}
		\caption*{(a) RMSE}
		\vspace{0.15em}
		\begin{tabular*}{\textwidth}{@{\extracolsep{\fill}}l P R r r P R r r P R r r@{}}
			\toprule
			& \multicolumn{4}{c}{All} & \multicolumn{4}{c}{Low imagery} & \multicolumn{4}{c}{High imagery} \\
			\cmidrule(lr){2-5} \cmidrule(lr){6-9} \cmidrule(lr){10-13}
			Source & \placeholderhead{White} & \realhead{ImageNet} & $\Delta$ & $\Delta-\Delta_H$ & \placeholderhead{White} & \realhead{ImageNet} & $\Delta$ & $\Delta-\Delta_H$ & \placeholderhead{White} & \realhead{ImageNet} & $\Delta$ & $\Delta-\Delta_H$ \\
			\midrule
			Human & \ctxmark{human-signed-all-white-t}{+0.25} & \ctxmark{human-signed-all-image-t}{\textbf{+0.64}} & \textsuperscript{***}\textbf{+0.39} & -- & \ctxmark{human-signed-low-white-t}{\textbf{+0.73}} & \ctxmark{human-signed-low-image-t}{\textbf{+1.02}} & \textsuperscript{***}\textbf{+0.29} & -- & \ctxmark{human-signed-high-white-t}{-0.21} & \ctxmark{human-signed-high-image-t}{+0.28} & \textsuperscript{***}\textbf{+0.49} & -- \\
			\midrule
			Gemma 3 & +0.45 & +1.59 & \textsuperscript{***}+1.15 & \textsuperscript{***}+0.75 & +1.40 & +2.54 & \textsuperscript{***}+1.14 & \textsuperscript{***}+0.85 & -0.46 & +0.69 & \textsuperscript{***}+1.16 & \textsuperscript{***}+0.66 \\
			InternVL3 & +0.19 & +0.89 & \textsuperscript{***}+0.70 & \textsuperscript{***}+0.31 & +1.18 & +1.72 & \textsuperscript{***}+0.54 & \textsuperscript{**}+0.25 & -0.76 & \textbf{+0.09} & \textsuperscript{***}+0.85 & \textsuperscript{***}+0.36 \\
			LLaVA 1.5 & \underline{+1.26} & \underline{+1.76} & \textsuperscript{***}+0.51 & \textsuperscript{*}\textbf{+0.11} & \underline{+2.49} & \underline{+2.98} & \textsuperscript{***}+0.48 & \textsuperscript{*}\textbf{+0.19} & \textbf{+0.08} & +0.61 & \textsuperscript{***}+0.53 & \textbf{+0.04} \\
			Pixtral & \textbf{-0.13} & +1.65 & \textsuperscript{***}\underline{+1.77} & \textsuperscript{***}\underline{+1.38} & +0.83 & +2.53 & \textsuperscript{***}\underline{+1.70} & \textsuperscript{***}\underline{+1.41} & \underline{-1.03} & \underline{+0.80} & \textsuperscript{***}\underline{+1.84} & \textsuperscript{***}\underline{+1.34} \\
			Qwen2.5-VL & \ctxmark{human-signed-all-white-b}{+0.30} & \ctxmark{human-signed-all-image-b}{+1.33} & \textsuperscript{***}+1.02 & \textsuperscript{***}+0.63 & \ctxmark{human-signed-low-white-b}{+1.48} & \ctxmark{human-signed-low-image-b}{+2.19} & \textsuperscript{***}+0.71 & \textsuperscript{***}+0.42 & \ctxmark{human-signed-high-white-b}{-0.82} & \ctxmark{human-signed-high-image-b}{+0.50} & \textsuperscript{***}+1.32 & \textsuperscript{***}+0.83 \\
			\noalign{\ctxoutline{placeholderctx}{human-signed-all-white-t}{human-signed-all-white-b}\ctxoutline{realctx}{human-signed-all-image-t}{human-signed-all-image-b}\ctxoutline{placeholderctx}{human-signed-low-white-t}{human-signed-low-white-b}\ctxoutline{realctx}{human-signed-low-image-t}{human-signed-low-image-b}\ctxoutline{placeholderctx}{human-signed-high-white-t}{human-signed-high-white-b}\ctxoutline{realctx}{human-signed-high-image-t}{human-signed-high-image-b}}
			\bottomrule
		\end{tabular*}
		\caption*{(b) Mean signed error}
		\vspace{0.15em}
		\begin{tabular*}{\textwidth}{@{\extracolsep{\fill}}l P R r r P R r r P R r r@{}}
			\toprule
			& \multicolumn{4}{c}{All} & \multicolumn{4}{c}{Low imagery} & \multicolumn{4}{c}{High imagery} \\
			\cmidrule(lr){2-5} \cmidrule(lr){6-9} \cmidrule(lr){10-13}
			Source & \placeholderhead{White} & \realhead{ImageNet} & $\Delta$ & $\Delta-\Delta_H$ & \placeholderhead{White} & \realhead{ImageNet} & $\Delta$ & $\Delta-\Delta_H$ & \placeholderhead{White} & \realhead{ImageNet} & $\Delta$ & $\Delta-\Delta_H$ \\
			\midrule
			Human & \ctxmark{human-spearman-all-white-t}{\textbf{0.67}} & \ctxmark{human-spearman-all-image-t}{\textbf{0.75}} & \textsuperscript{**}+0.08 & -- & \ctxmark{human-spearman-low-white-t}{0.40} & \ctxmark{human-spearman-low-image-t}{0.44} & +0.03 & -- & \ctxmark{human-spearman-high-white-t}{\textbf{0.51}} & \ctxmark{human-spearman-high-image-t}{\textbf{0.60}} & +0.09 & -- \\
			\midrule
			Gemma 3 & 0.41 & 0.44 & \underline{+0.03} & -0.05 & \textbf{0.53} & \textbf{0.49} & \underline{-0.04} & -0.07 & \underline{0.08} & 0.10 & +0.02 & -0.07 \\
			InternVL3 & 0.59 & 0.64 & +0.05 & \textbf{-0.04} & 0.33 & 0.32 & -0.02 & \textbf{-0.05} & 0.48 & 0.48 & -0.01 & -0.10 \\
			LLaVA 1.5 & \underline{-0.01} & \underline{0.13} & +0.14 & +0.06 & \underline{-0.09} & \underline{0.07} & +0.17 & \textsuperscript{*}+0.13 & 0.08 & \underline{0.03} & \underline{-0.05} & \underline{-0.14} \\
			Pixtral & 0.39 & 0.59 & \textsuperscript{**}+0.21 & +0.12 & 0.40 & 0.48 & +0.08 & \textbf{+0.05} & 0.12 & 0.25 & +0.14 & \textbf{+0.04} \\
			Qwen2.5-VL & \ctxmark{human-spearman-all-white-b}{0.18} & \ctxmark{human-spearman-all-image-b}{0.58} & \textsuperscript{**}\textbf{+0.40} & \textsuperscript{***}\underline{+0.32} & \ctxmark{human-spearman-low-white-b}{0.13} & \ctxmark{human-spearman-low-image-b}{0.42} & \textsuperscript{**}\textbf{+0.29} & \textsuperscript{*}\underline{+0.26} & \ctxmark{human-spearman-high-white-b}{0.10} & \ctxmark{human-spearman-high-image-b}{0.26} & \textbf{+0.15} & +0.06 \\
			\noalign{\ctxoutline{placeholderctx}{human-spearman-all-white-t}{human-spearman-all-white-b}\ctxoutline{realctx}{human-spearman-all-image-t}{human-spearman-all-image-b}\ctxoutline{placeholderctx}{human-spearman-low-white-t}{human-spearman-low-white-b}\ctxoutline{realctx}{human-spearman-low-image-t}{human-spearman-low-image-b}\ctxoutline{placeholderctx}{human-spearman-high-white-t}{human-spearman-high-white-b}\ctxoutline{realctx}{human-spearman-high-image-t}{human-spearman-high-image-b}}
			\bottomrule
		\end{tabular*}
		\caption*{(c) Spearman rank correlation}
	}
	\caption{Human validation metrics on the CP2004B subset under \emph{White} and retrieved \emph{ImageNet} contexts, with $\Delta=\emph{ImageNet}-\emph{White}$.
		Positive $\Delta$ means worse RMSE, upward signed-error shift, or improved Spearman rank agreement.
		\(\Delta-\Delta_H\) is the source-specific $\Delta$ minus the human $\Delta$ for the same split and metric.
		\textbf{Boldface}/\underline{underlining} mark best/worst values per metric column.
		Star superscripts on $\Delta$ mark paired significance for \emph{ImageNet} versus \emph{White} within a source. Star superscripts on \(\Delta-\Delta_H\) mark paired significance versus the human $\Delta$, after Benjamini--Hochberg correction (\textsuperscript{*} $q<0.05$, \textsuperscript{**} $q<0.01$, \textsuperscript{***} $q<0.001$).}
	\label{tab:human_vlm_metrics_subset}
\end{table*}

\begin{table*}[!t]
\centering
\footnotesize
\begin{minipage}{0.48\textwidth}
\centering
\setlength{\tabcolsep}{3pt}
\resizebox{\linewidth}{!}{
\begin{tabular}{l r P P R R R}
\hline
& \emph{None} & \placeholderhead{White} & \placeholderhead{Noise} & \placeholderhead{Shuffled} & \realhead{ImageNet} & \realhead{Wikimedia} \\
\hline
\multicolumn{7}{c}{\textbf{All}} \\
\hline
Gemma 3 & \shiftcell{85}{\textbf{-0.02}} & \ctxmark{tab-appendix-signed-shift-mt-all-p-t}{\shiftcell{37}{-0.42}} & \shiftcell{21}{-0.54} & \shiftcell{5}{\underline{-0.85}} & \ctxmark{tab-appendix-signed-shift-mt-all-r-t}{\shiftcell{53}{+0.21}} & \shiftcell{69}{+0.17} \\
InternVL3 & \shiftcell{53}{-0.31} & \shiftcell{37}{-0.49} & \shiftcell{5}{\underline{-1.17}} & \shiftcell{21}{-0.90} & \shiftcell{85}{\textbf{-0.01}} & \shiftcell{69}{-0.05} \\
LLaVA 1.5 & \shiftcell{5}{\underline{+0.75}} & \shiftcell{85}{\textbf{+0.71}} & \shiftcell{53}{+0.71} & \shiftcell{69}{\textbf{+0.71}} & \shiftcell{37}{+0.71} & \shiftcell{21}{+0.71} \\
Pixtral & \shiftcell{5}{\underline{+0.59}} & \shiftcell{85}{\textbf{-0.30}} & \shiftcell{69}{+0.31} & \shiftcell{21}{-0.54} & \shiftcell{37}{+0.46} & \shiftcell{53}{+0.46} \\
Qwen2.5-VL & \shiftcell{37}{-0.28} & \shiftcell{5}{\underline{-0.33}} & \shiftcell{21}{-0.31} & \ctxmark{tab-appendix-signed-shift-mt-all-p-b}{\shiftcell{53}{-0.26}} & \shiftcell{69}{+0.10} & \ctxmark{tab-appendix-signed-shift-mt-all-r-b}{\shiftcell{85}{\textbf{+0.05}}} \\
\noalign{\ctxoutline{placeholderctx}{tab-appendix-signed-shift-mt-all-p-t}{tab-appendix-signed-shift-mt-all-p-b}\ctxoutline{realctx}{tab-appendix-signed-shift-mt-all-r-t}{tab-appendix-signed-shift-mt-all-r-b}}
\hline
\multicolumn{7}{c}{\textbf{Abstract}} \\
\hline
Gemma 3 & \shiftcell{37}{+0.62} & \ctxmark{tab-appendix-signed-shift-mt-lo-p-t}{\shiftcell{69}{+0.30}} & \shiftcell{53}{+0.41} & \shiftcell{85}{\textbf{+0.24}} & \ctxmark{tab-appendix-signed-shift-mt-lo-r-t}{\shiftcell{5}{\underline{+0.80}}} & \shiftcell{21}{+0.78} \\
InternVL3 & \shiftcell{37}{+0.45} & \shiftcell{53}{+0.32} & \shiftcell{85}{\textbf{-0.04}} & \shiftcell{69}{+0.17} & \shiftcell{5}{\underline{+0.68}} & \shiftcell{21}{+0.65} \\
LLaVA 1.5 & \shiftcell{5}{\underline{+1.86}} & \shiftcell{85}{\textbf{+1.82}} & \shiftcell{37}{+1.82} & \shiftcell{69}{+1.82} & \shiftcell{21}{+1.82} & \shiftcell{53}{+1.82} \\
Pixtral & \shiftcell{5}{\underline{+1.50}} & \shiftcell{85}{\textbf{+0.37}} & \shiftcell{21}{+1.27} & \shiftcell{69}{+0.58} & \shiftcell{37}{+1.14} & \shiftcell{53}{+1.12} \\
Qwen2.5-VL & \shiftcell{85}{\textbf{+0.60}} & \shiftcell{53}{+0.66} & \shiftcell{69}{+0.61} & \ctxmark{tab-appendix-signed-shift-mt-lo-p-b}{\shiftcell{37}{+0.72}} & \shiftcell{5}{\underline{+0.95}} & \ctxmark{tab-appendix-signed-shift-mt-lo-r-b}{\shiftcell{21}{+0.89}} \\
\noalign{\ctxoutline{placeholderctx}{tab-appendix-signed-shift-mt-lo-p-t}{tab-appendix-signed-shift-mt-lo-p-b}\ctxoutline{realctx}{tab-appendix-signed-shift-mt-lo-r-t}{tab-appendix-signed-shift-mt-lo-r-b}}
\hline
\multicolumn{7}{c}{\textbf{Concrete}} \\
\hline
Gemma 3 & \shiftcell{53}{-0.56} & \ctxmark{tab-appendix-signed-shift-mt-hi-p-t}{\shiftcell{37}{-1.03}} & \shiftcell{21}{-1.35} & \shiftcell{5}{\underline{-1.77}} & \ctxmark{tab-appendix-signed-shift-mt-hi-r-t}{\shiftcell{85}{\textbf{-0.29}}} & \shiftcell{69}{-0.35} \\
InternVL3 & \shiftcell{53}{-0.96} & \shiftcell{37}{-1.18} & \shiftcell{5}{\underline{-2.13}} & \shiftcell{21}{-1.81} & \shiftcell{85}{\textbf{-0.59}} & \shiftcell{69}{-0.64} \\
LLaVA 1.5 & \shiftcell{85}{\textbf{-0.19}} & \shiftcell{37}{-0.23} & \shiftcell{5}{\underline{-0.23}} & \shiftcell{21}{-0.23} & \shiftcell{53}{-0.23} & \shiftcell{69}{-0.23} \\
Pixtral & \shiftcell{53}{-0.18} & \shiftcell{21}{-0.87} & \shiftcell{37}{-0.50} & \shiftcell{5}{\underline{-1.48}} & \shiftcell{69}{-0.12} & \shiftcell{85}{\textbf{-0.10}} \\
Qwen2.5-VL & \shiftcell{53}{-1.01} & \shiftcell{21}{-1.16} & \shiftcell{37}{-1.08} & \ctxmark{tab-appendix-signed-shift-mt-hi-p-b}{\shiftcell{5}{\underline{-1.18}}} & \shiftcell{85}{\textbf{-0.61}} & \ctxmark{tab-appendix-signed-shift-mt-hi-r-b}{\shiftcell{69}{-0.66}} \\
\noalign{\ctxoutline{placeholderctx}{tab-appendix-signed-shift-mt-hi-p-t}{tab-appendix-signed-shift-mt-hi-p-b}\ctxoutline{realctx}{tab-appendix-signed-shift-mt-hi-r-t}{tab-appendix-signed-shift-mt-hi-r-b}}
\hline
\end{tabular}
}
\caption*{MT40k (Concreteness)}
\end{minipage}
\hfill
\begin{minipage}{0.48\textwidth}
\centering
\setlength{\tabcolsep}{3pt}
\resizebox{\linewidth}{!}{
\begin{tabular}{l r P P R R R}
\hline
& \emph{None} & \placeholderhead{White} & \placeholderhead{Noise} & \placeholderhead{Shuffled} & \realhead{ImageNet} & \realhead{Wikimedia} \\
\hline
\multicolumn{7}{c}{\textbf{All}} \\
\hline
Gemma 3 & \shiftcell{53}{+0.54} & \ctxmark{tab-appendix-signed-shift-cp-all-p-t}{\shiftcell{69}{+0.48}} & \shiftcell{37}{-1.11} & \shiftcell{85}{\textbf{+0.27}} & \ctxmark{tab-appendix-signed-shift-cp-all-r-t}{\shiftcell{5}{\underline{+1.59}}} & \shiftcell{21}{+1.49} \\
InternVL3 & \shiftcell{53}{+0.49} & \shiftcell{85}{\textbf{+0.20}} & \shiftcell{5}{\underline{-1.38}} & \shiftcell{69}{-0.37} & \shiftcell{21}{+0.91} & \shiftcell{37}{+0.81} \\
LLaVA 1.5 & \shiftcell{85}{\textbf{+0.65}} & \shiftcell{69}{+1.26} & \shiftcell{37}{+1.72} & \shiftcell{53}{+1.56} & \shiftcell{5}{\underline{+1.77}} & \shiftcell{21}{+1.76} \\
Pixtral & \shiftcell{37}{+1.55} & \shiftcell{85}{\textbf{-0.10}} & \shiftcell{53}{+0.41} & \shiftcell{69}{+0.35} & \shiftcell{21}{+1.65} & \shiftcell{5}{\underline{+1.69}} \\
Qwen2.5-VL & \shiftcell{37}{+0.39} & \shiftcell{53}{+0.33} & \shiftcell{69}{+0.31} & \ctxmark{tab-appendix-signed-shift-cp-all-p-b}{\shiftcell{85}{\textbf{-0.11}}} & \shiftcell{5}{\underline{+1.33}} & \ctxmark{tab-appendix-signed-shift-cp-all-r-b}{\shiftcell{21}{+1.21}} \\
\noalign{\ctxoutline{placeholderctx}{tab-appendix-signed-shift-cp-all-p-t}{tab-appendix-signed-shift-cp-all-p-b}\ctxoutline{realctx}{tab-appendix-signed-shift-cp-all-r-t}{tab-appendix-signed-shift-cp-all-r-b}}
\hline
\multicolumn{7}{c}{\textbf{Low imagery}} \\
\hline
Gemma 3 & \shiftcell{53}{+1.51} & \ctxmark{tab-appendix-signed-shift-cp-lo-p-t}{\shiftcell{69}{+1.42}} & \shiftcell{85}{\textbf{+0.18}} & \shiftcell{37}{+1.62} & \ctxmark{tab-appendix-signed-shift-cp-lo-r-t}{\shiftcell{5}{\underline{+2.56}}} & \shiftcell{21}{+2.45} \\
InternVL3 & \shiftcell{37}{+1.48} & \shiftcell{53}{+1.20} & \shiftcell{85}{\textbf{-0.05}} & \shiftcell{69}{+0.97} & \shiftcell{5}{\underline{+1.76}} & \shiftcell{21}{+1.67} \\
LLaVA 1.5 & \shiftcell{85}{\textbf{+1.88}} & \shiftcell{69}{+2.51} & \shiftcell{37}{+2.95} & \shiftcell{53}{+2.86} & \shiftcell{5}{\underline{+2.99}} & \shiftcell{21}{+2.99} \\
Pixtral & \shiftcell{5}{\underline{+2.62}} & \shiftcell{85}{\textbf{+0.86}} & \shiftcell{53}{+1.67} & \shiftcell{69}{+1.63} & \shiftcell{37}{+2.56} & \shiftcell{21}{+2.61} \\
Qwen2.5-VL & \shiftcell{53}{+1.52} & \shiftcell{69}{+1.51} & \shiftcell{37}{+1.55} & \ctxmark{tab-appendix-signed-shift-cp-lo-p-b}{\shiftcell{85}{\textbf{+1.20}}} & \shiftcell{5}{\underline{+2.20}} & \ctxmark{tab-appendix-signed-shift-cp-lo-r-b}{\shiftcell{21}{+2.11}} \\
\noalign{\ctxoutline{placeholderctx}{tab-appendix-signed-shift-cp-lo-p-t}{tab-appendix-signed-shift-cp-lo-p-b}\ctxoutline{realctx}{tab-appendix-signed-shift-cp-lo-r-t}{tab-appendix-signed-shift-cp-lo-r-b}}
\hline
\multicolumn{7}{c}{\textbf{High imagery}} \\
\hline
Gemma 3 & \shiftcell{85}{\textbf{-0.37}} & \ctxmark{tab-appendix-signed-shift-cp-hi-p-t}{\shiftcell{69}{-0.40}} & \shiftcell{5}{\underline{-2.32}} & \shiftcell{21}{-1.01} & \ctxmark{tab-appendix-signed-shift-cp-hi-r-t}{\shiftcell{37}{+0.68}} & \shiftcell{53}{+0.59} \\
InternVL3 & \shiftcell{53}{-0.45} & \shiftcell{37}{-0.74} & \shiftcell{5}{\underline{-2.64}} & \shiftcell{21}{-1.63} & \shiftcell{69}{+0.10} & \shiftcell{85}{\textbf{+0.01}} \\
LLaVA 1.5 & \shiftcell{53}{-0.51} & \shiftcell{85}{\textbf{+0.08}} & \shiftcell{37}{+0.57} & \shiftcell{69}{+0.32} & \shiftcell{5}{\underline{+0.61}} & \shiftcell{21}{+0.61} \\
Pixtral & \shiftcell{85}{\textbf{+0.55}} & \shiftcell{5}{\underline{-1.00}} & \shiftcell{69}{-0.77} & \shiftcell{21}{-0.86} & \shiftcell{53}{+0.79} & \shiftcell{37}{+0.82} \\
Qwen2.5-VL & \shiftcell{53}{-0.67} & \shiftcell{37}{-0.78} & \shiftcell{21}{-0.86} & \ctxmark{tab-appendix-signed-shift-cp-hi-p-b}{\shiftcell{5}{\underline{-1.36}}} & \shiftcell{69}{+0.51} & \ctxmark{tab-appendix-signed-shift-cp-hi-r-b}{\shiftcell{85}{\textbf{+0.37}}} \\
\noalign{\ctxoutline{placeholderctx}{tab-appendix-signed-shift-cp-hi-p-t}{tab-appendix-signed-shift-cp-hi-p-b}\ctxoutline{realctx}{tab-appendix-signed-shift-cp-hi-r-t}{tab-appendix-signed-shift-cp-hi-r-b}}
\hline
\end{tabular}
}
\caption*{CP2004B (Imagery)}
\end{minipage}

\caption{Mean signed prediction error, $\hat{y}_{\mathrm{context}}-y_{\mathrm{human}}$, for concreteness (left) and imagery (right).
Positive/negative values indicate over-/underestimation.
\placeholderctxtext{\emph{White}/\emph{Noise}} are uninformative contexts, and \placeholderctxtext{\emph{Shuffled}} is a mismatched-image control. \realctxtext{\emph{ImageNet}/\emph{Wikimedia}} are matched real-image contexts.
Green/purple cells mark smaller/larger within-row absolute errors. \textbf{Boldface}/\underline{underlining} mark best/worst contexts.}
\label{tab:appendix_signed_shift}
\end{table*}

\begin{table*}[!t]
\centering
\footnotesize
\begin{minipage}{0.48\textwidth}
\centering
\setlength{\tabcolsep}{3pt}
\resizebox{\linewidth}{!}{
\begin{tabular}{l r P P R R R}
\hline
& \emph{None} & \placeholderhead{White} & \placeholderhead{Noise} & \placeholderhead{Shuffled} & \realhead{ImageNet} & \realhead{Wikimedia} \\
\hline
\multicolumn{7}{c}{\textbf{All}} \\
\hline
LLaMA 2 & \corrcell{85}{\underline{\textbf{0.426}}} & -- & -- & -- & -- & -- \\
Mistral & \corrcell{85}{\underline{\textbf{0.737}}} & -- & -- & -- & -- & -- \\
Qwen2.5 & \corrcell{85}{\underline{\textbf{0.829}}} & -- & -- & -- & -- & -- \\
Gemma 3 & \corrcell{85}{\textbf{0.797}} & \ctxmark{tab-appendix-gen-corr-mt-all-p-t}{\corrcell{37}{\textsuperscript{***}0.730}} & \corrcell{21}{\textsuperscript{***}0.430} & \corrcell{5}{\textsuperscript{***}\underline{0.015}} & \ctxmark{tab-appendix-gen-corr-mt-all-r-t}{\corrcell{69}{\textsuperscript{*}0.775}} & \corrcell{53}{\textsuperscript{**}0.767} \\
InternVL3 & \corrcell{85}{\textbf{0.775}} & \corrcell{69}{\textsuperscript{*}0.751} & \corrcell{5}{\textsuperscript{***}\underline{-0.111}} & \corrcell{21}{\textsuperscript{***}0.089} & \corrcell{53}{\textsuperscript{*}0.744} & \corrcell{37}{\textsuperscript{**}0.736} \\
LLaVA 1.5 & \corrcell{69}{0.039} & \corrcell{85}{\textbf{0.040}} & \corrcell{5}{\textsuperscript{**}\underline{-0.085}} & \corrcell{21}{-0.025} & \corrcell{37}{0.004} & \corrcell{53}{0.022} \\
Pixtral & \corrcell{69}{0.659} & \corrcell{37}{0.655} & \corrcell{21}{\textsuperscript{***}0.410} & \corrcell{5}{\textsuperscript{***}\underline{-0.013}} & \corrcell{53}{0.655} & \corrcell{85}{\textsuperscript{*}\textbf{0.695}} \\
Qwen2.5-VL & \corrcell{53}{0.600} & \corrcell{21}{\textsuperscript{***}0.497} & \corrcell{37}{\textsuperscript{*}0.564} & \ctxmark{tab-appendix-gen-corr-mt-all-p-b}{\corrcell{5}{\textsuperscript{***}\underline{0.206}}} & \corrcell{69}{0.629} & \ctxmark{tab-appendix-gen-corr-mt-all-r-b}{\corrcell{85}{\textsuperscript{*}\textbf{0.642}}} \\
\noalign{\ctxoutline{placeholderctx}{tab-appendix-gen-corr-mt-all-p-t}{tab-appendix-gen-corr-mt-all-p-b}\ctxoutline{realctx}{tab-appendix-gen-corr-mt-all-r-t}{tab-appendix-gen-corr-mt-all-r-b}}
\hline
\multicolumn{7}{c}{\textbf{Abstract}} \\
\hline
LLaMA 2 & \corrcell{85}{\underline{\textbf{0.146}}} & -- & -- & -- & -- & -- \\
Mistral & \corrcell{85}{\underline{\textbf{0.281}}} & -- & -- & -- & -- & -- \\
Qwen2.5 & \corrcell{85}{\underline{\textbf{0.358}}} & -- & -- & -- & -- & -- \\
Gemma 3 & \corrcell{85}{\textbf{0.413}} & \ctxmark{tab-appendix-gen-corr-mt-lo-p-t}{\corrcell{37}{0.346}} & \corrcell{21}{\textsuperscript{***}0.214} & \corrcell{5}{\textsuperscript{***}\underline{0.106}} & \ctxmark{tab-appendix-gen-corr-mt-lo-r-t}{\corrcell{69}{0.358}} & \corrcell{53}{0.356} \\
InternVL3 & \corrcell{85}{\textbf{0.274}} & \corrcell{69}{0.264} & \corrcell{21}{\textsuperscript{***}0.036} & \corrcell{5}{\textsuperscript{***}\underline{0.018}} & \corrcell{37}{0.238} & \corrcell{53}{0.244} \\
LLaVA 1.5 & \corrcell{5}{\underline{-0.039}} & \corrcell{37}{0.000} & \corrcell{69}{0.025} & \corrcell{53}{0.004} & \corrcell{85}{\textbf{0.036}} & \corrcell{21}{-0.004} \\
Pixtral & \corrcell{53}{0.229} & \corrcell{37}{0.224} & \corrcell{21}{0.206} & \corrcell{5}{\underline{0.145}} & \corrcell{69}{0.234} & \corrcell{85}{\textbf{0.287}} \\
Qwen2.5-VL & \corrcell{69}{0.172} & \corrcell{37}{0.117} & \corrcell{21}{\textsuperscript{*}0.070} & \ctxmark{tab-appendix-gen-corr-mt-lo-p-b}{\corrcell{5}{\textsuperscript{***}\underline{-0.024}}} & \corrcell{53}{0.154} & \ctxmark{tab-appendix-gen-corr-mt-lo-r-b}{\corrcell{85}{\textbf{0.199}}} \\
\noalign{\ctxoutline{placeholderctx}{tab-appendix-gen-corr-mt-lo-p-t}{tab-appendix-gen-corr-mt-lo-p-b}\ctxoutline{realctx}{tab-appendix-gen-corr-mt-lo-r-t}{tab-appendix-gen-corr-mt-lo-r-b}}
\hline
\multicolumn{7}{c}{\textbf{Concrete}} \\
\hline
LLaMA 2 & \corrcell{85}{\underline{\textbf{0.286}}} & -- & -- & -- & -- & -- \\
Mistral & \corrcell{85}{\underline{\textbf{0.640}}} & -- & -- & -- & -- & -- \\
Qwen2.5 & \corrcell{85}{\underline{\textbf{0.729}}} & -- & -- & -- & -- & -- \\
Gemma 3 & \corrcell{85}{\textbf{0.690}} & \ctxmark{tab-appendix-gen-corr-mt-hi-p-t}{\corrcell{37}{\textsuperscript{***}0.596}} & \corrcell{21}{\textsuperscript{***}0.201} & \corrcell{5}{\textsuperscript{***}\underline{-0.057}} & \ctxmark{tab-appendix-gen-corr-mt-hi-r-t}{\corrcell{69}{\textsuperscript{*}0.642}} & \corrcell{53}{\textsuperscript{**}0.633} \\
InternVL3 & \corrcell{69}{0.711} & \corrcell{85}{\textbf{0.711}} & \corrcell{5}{\textsuperscript{***}\underline{-0.231}} & \corrcell{21}{\textsuperscript{***}0.063} & \corrcell{53}{0.702} & \corrcell{37}{0.664} \\
LLaVA 1.5 & \corrcell{85}{\textbf{0.191}} & \corrcell{53}{\textsuperscript{***}0.031} & \corrcell{5}{\textsuperscript{***}\underline{-0.077}} & \corrcell{21}{\textsuperscript{***}-0.064} & \corrcell{37}{\textsuperscript{***}0.018} & \corrcell{69}{\textsuperscript{**}0.043} \\
Pixtral & \corrcell{37}{0.494} & \corrcell{85}{\textsuperscript{**}\textbf{0.582}} & \corrcell{21}{\textsuperscript{***}0.287} & \corrcell{5}{\textsuperscript{***}\underline{-0.160}} & \corrcell{53}{0.516} & \corrcell{69}{0.527} \\
Qwen2.5-VL & \corrcell{69}{0.628} & \corrcell{53}{\textsuperscript{**}0.568} & \corrcell{85}{\textbf{0.655}} & \ctxmark{tab-appendix-gen-corr-mt-hi-p-b}{\corrcell{5}{\textsuperscript{***}\underline{0.340}}} & \corrcell{21}{\textsuperscript{**}0.520} & \ctxmark{tab-appendix-gen-corr-mt-hi-r-b}{\corrcell{37}{\textsuperscript{*}0.549}} \\
\noalign{\ctxoutline{placeholderctx}{tab-appendix-gen-corr-mt-hi-p-t}{tab-appendix-gen-corr-mt-hi-p-b}\ctxoutline{realctx}{tab-appendix-gen-corr-mt-hi-r-t}{tab-appendix-gen-corr-mt-hi-r-b}}
\hline
\end{tabular}
}
\caption*{MT40k (Concreteness)}
\end{minipage}
\hfill
\begin{minipage}{0.48\textwidth}
\centering
\setlength{\tabcolsep}{3pt}
\resizebox{\linewidth}{!}{
\begin{tabular}{l r P P R R R}
\hline
& \emph{None} & \placeholderhead{White} & \placeholderhead{Noise} & \placeholderhead{Shuffled} & \realhead{ImageNet} & \realhead{Wikimedia} \\
\hline
\multicolumn{7}{c}{\textbf{All}} \\
\hline
LLaMA 2 & \corrcell{85}{\underline{\textbf{0.085}}} & -- & -- & -- & -- & -- \\
Mistral & \corrcell{85}{\underline{\textbf{0.638}}} & -- & -- & -- & -- & -- \\
Qwen2.5 & \corrcell{85}{\underline{\textbf{0.538}}} & -- & -- & -- & -- & -- \\
Gemma 3 & \corrcell{85}{\textbf{0.439}} & \ctxmark{tab-appendix-gen-corr-cp-all-p-t}{\corrcell{53}{0.414}} & \corrcell{21}{\textsuperscript{***}-0.106} & \corrcell{5}{\textsuperscript{***}\underline{-0.144}} & \ctxmark{tab-appendix-gen-corr-cp-all-r-t}{\corrcell{37}{0.413}} & \corrcell{69}{0.428} \\
InternVL3 & \corrcell{85}{\textbf{0.641}} & \corrcell{37}{\textsuperscript{***}0.561} & \corrcell{5}{\textsuperscript{***}\underline{-0.329}} & \corrcell{21}{\textsuperscript{***}-0.184} & \corrcell{69}{0.624} & \corrcell{53}{0.614} \\
LLaVA 1.5 & \corrcell{37}{0.047} & \corrcell{21}{-0.007} & \corrcell{53}{0.058} & \corrcell{5}{\textsuperscript{***}\underline{-0.229}} & \corrcell{69}{\textsuperscript{**}0.183} & \corrcell{85}{\textsuperscript{**}\textbf{0.187}} \\
Pixtral & \corrcell{53}{0.438} & \corrcell{37}{\textsuperscript{*}0.365} & \corrcell{21}{\textsuperscript{***}-0.021} & \corrcell{5}{\textsuperscript{***}\underline{-0.053}} & \corrcell{69}{\textsuperscript{***}0.578} & \corrcell{85}{\textsuperscript{***}\textbf{0.591}} \\
Qwen2.5-VL & \corrcell{53}{0.361} & \corrcell{37}{\textsuperscript{***}0.195} & \corrcell{21}{\textsuperscript{***}-0.019} & \ctxmark{tab-appendix-gen-corr-cp-all-p-b}{\corrcell{5}{\textsuperscript{***}\underline{-0.163}}} & \corrcell{85}{\textsuperscript{***}\textbf{0.576}} & \ctxmark{tab-appendix-gen-corr-cp-all-r-b}{\corrcell{69}{\textsuperscript{***}0.562}} \\
\noalign{\ctxoutline{placeholderctx}{tab-appendix-gen-corr-cp-all-p-t}{tab-appendix-gen-corr-cp-all-p-b}\ctxoutline{realctx}{tab-appendix-gen-corr-cp-all-r-t}{tab-appendix-gen-corr-cp-all-r-b}}
\hline
\multicolumn{7}{c}{\textbf{Low imagery}} \\
\hline
LLaMA 2 & \corrcell{85}{\underline{\textbf{0.178}}} & -- & -- & -- & -- & -- \\
Mistral & \corrcell{85}{\underline{\textbf{0.509}}} & -- & -- & -- & -- & -- \\
Qwen2.5 & \corrcell{85}{\underline{\textbf{0.426}}} & -- & -- & -- & -- & -- \\
Gemma 3 & \corrcell{85}{\textbf{0.474}} & \ctxmark{tab-appendix-gen-corr-cp-lo-p-t}{\corrcell{53}{0.448}} & \corrcell{21}{\textsuperscript{***}0.118} & \corrcell{5}{\textsuperscript{***}\underline{0.109}} & \ctxmark{tab-appendix-gen-corr-cp-lo-r-t}{\corrcell{69}{0.450}} & \corrcell{37}{0.441} \\
InternVL3 & \corrcell{85}{\textbf{0.463}} & \corrcell{69}{\textsuperscript{***}0.349} & \corrcell{21}{\textsuperscript{***}-0.036} & \corrcell{5}{\textsuperscript{***}\underline{-0.116}} & \corrcell{37}{\textsuperscript{***}0.298} & \corrcell{53}{\textsuperscript{***}0.325} \\
LLaVA 1.5 & \corrcell{85}{\textbf{0.207}} & \corrcell{21}{\textsuperscript{***}-0.040} & \corrcell{37}{\textsuperscript{**}0.006} & \corrcell{5}{\textsuperscript{***}\underline{-0.063}} & \corrcell{53}{0.097} & \corrcell{69}{0.099} \\
Pixtral & \corrcell{85}{\textbf{0.478}} & \corrcell{37}{\textsuperscript{*}0.373} & \corrcell{5}{\textsuperscript{***}\underline{-0.093}} & \corrcell{21}{\textsuperscript{***}0.044} & \corrcell{69}{0.430} & \corrcell{53}{\textsuperscript{*}0.404} \\
Qwen2.5-VL & \corrcell{53}{0.324} & \corrcell{37}{\textsuperscript{***}0.195} & \corrcell{21}{\textsuperscript{***}-0.038} & \ctxmark{tab-appendix-gen-corr-cp-lo-p-b}{\corrcell{5}{\textsuperscript{***}\underline{-0.092}}} & \corrcell{85}{\textbf{0.395}} & \ctxmark{tab-appendix-gen-corr-cp-lo-r-b}{\corrcell{69}{0.379}} \\
\noalign{\ctxoutline{placeholderctx}{tab-appendix-gen-corr-cp-lo-p-t}{tab-appendix-gen-corr-cp-lo-p-b}\ctxoutline{realctx}{tab-appendix-gen-corr-cp-lo-r-t}{tab-appendix-gen-corr-cp-lo-r-b}}
\hline
\multicolumn{7}{c}{\textbf{High imagery}} \\
\hline
LLaMA 2 & \corrcell{85}{\underline{\textbf{0.001}}} & -- & -- & -- & -- & -- \\
Mistral & \corrcell{85}{\underline{\textbf{0.429}}} & -- & -- & -- & -- & -- \\
Qwen2.5 & \corrcell{85}{\underline{\textbf{0.279}}} & -- & -- & -- & -- & -- \\
Gemma 3 & \corrcell{37}{0.009} & \ctxmark{tab-appendix-gen-corr-cp-hi-p-t}{\corrcell{69}{0.031}} & \corrcell{5}{\textsuperscript{***}\underline{-0.230}} & \corrcell{21}{\textsuperscript{***}-0.206} & \ctxmark{tab-appendix-gen-corr-cp-hi-r-t}{\corrcell{85}{\textbf{0.039}}} & \corrcell{53}{0.022} \\
InternVL3 & \corrcell{53}{0.416} & \corrcell{69}{0.423} & \corrcell{5}{\textsuperscript{***}\underline{-0.292}} & \corrcell{21}{\textsuperscript{***}-0.059} & \corrcell{85}{\textbf{0.431}} & \corrcell{37}{0.389} \\
LLaVA 1.5 & \corrcell{21}{-0.082} & \corrcell{53}{\textsuperscript{**}0.049} & \corrcell{69}{\textsuperscript{*}0.052} & \corrcell{5}{\underline{-0.140}} & \corrcell{37}{0.041} & \corrcell{85}{\textsuperscript{*}\textbf{0.077}} \\
Pixtral & \corrcell{21}{0.042} & \corrcell{37}{0.065} & \corrcell{53}{0.096} & \corrcell{5}{\underline{-0.070}} & \corrcell{69}{\textsuperscript{***}0.265} & \corrcell{85}{\textsuperscript{***}\textbf{0.285}} \\
Qwen2.5-VL & \corrcell{53}{0.161} & \corrcell{37}{\textsuperscript{***}0.029} & \corrcell{21}{\textsuperscript{***}-0.032} & \ctxmark{tab-appendix-gen-corr-cp-hi-p-b}{\corrcell{5}{\textsuperscript{***}\underline{-0.095}}} & \corrcell{85}{\textbf{0.250}} & \ctxmark{tab-appendix-gen-corr-cp-hi-r-b}{\corrcell{69}{0.234}} \\
\noalign{\ctxoutline{placeholderctx}{tab-appendix-gen-corr-cp-hi-p-t}{tab-appendix-gen-corr-cp-hi-p-b}\ctxoutline{realctx}{tab-appendix-gen-corr-cp-hi-r-t}{tab-appendix-gen-corr-cp-hi-r-b}}
\hline
\end{tabular}
}
\caption*{CP2004B (Imagery)}
\end{minipage}

\caption{Spearman correlations between model predictions and human ratings for concreteness (left) and imagery (right).
\placeholderctxtext{\emph{White}/\emph{Noise}} are uninformative contexts, and \placeholderctxtext{\emph{Shuffled}} is a mismatched-image control. \realctxtext{\emph{ImageNet}/\emph{Wikimedia}} are matched real-image contexts.
Green/purple cells mark higher/lower within-row correlations. \textbf{Boldface}/\underline{underlining} mark best/worst contexts.
Superscripts mark significance versus \emph{None} (\textsuperscript{*} $q<0.05$, \textsuperscript{**} $q<0.01$, \textsuperscript{***} $q<0.001$, paired swap tests with Benjamini--Hochberg correction).}
\label{tab:appendix_gen_corr}
\end{table*}

\begin{table*}[t]
	\centering
	\scriptsize
	\begin{minipage}[t]{0.49\textwidth}
		\centering
		\setlength{\tabcolsep}{3pt}
		\renewcommand{\arraystretch}{0.95}
		\begin{tabular}{l rr rr}
			\hline
			              & \multicolumn{2}{c}{Win margin}                    & \multicolumn{2}{c}{MAE reduction}                                                                                                                             \\
			Variant       & \realhead{ImageNet}                               & \realhead{Wikimedia}                              & \realhead{ImageNet}                                 & \realhead{Wikimedia}                                \\
			\hline
			\multicolumn{5}{c}{\textbf{All}}                                                                                                                                                                                                  \\
			\hline
			V1            & \cellcolor{heatlow!29}{\textsuperscript{***}-23}  & \cellcolor{heatlow!41}{\textsuperscript{***}-32}  & \cellcolor{heatlow!28}{\textsuperscript{***}-0.08}  & \cellcolor{heatlow!35}{\textsuperscript{***}-0.10}  \\
			V2            & \cellcolor{heatlow!40}{\textsuperscript{***}-29}  & \cellcolor{heatlow!51}{\textsuperscript{***}-37}  & \cellcolor{heatlow!39}{\textsuperscript{***}-0.12}  & \cellcolor{heatlow!42}{\textsuperscript{***}-0.13}  \\
			V3            & \cellcolor{heatlow!37}{\textsuperscript{***}-20}  & \cellcolor{heatlow!46}{\textsuperscript{***}-25}  & \cellcolor{heatlow!39}{\textsuperscript{***}-0.06}  & \cellcolor{heatlow!39}{\textsuperscript{***}-0.06}  \\
			V4            & \cellcolor{heatlow!39}{\textsuperscript{***}-23}  & \cellcolor{heatlow!49}{\textsuperscript{***}-29}  & \cellcolor{heatlow!34}{\textsuperscript{***}-0.06}  & \cellcolor{heatlow!45}{\textsuperscript{***}-0.08}  \\
			V5            & \cellcolor{heatlow!6}{\textsuperscript{*}-6}      & \cellcolor{heatlow!11}{\textsuperscript{***}-11}  & \cellcolor{heatlow!10}{\textsuperscript{***}-0.05}  & \cellcolor{heatlow!17}{\textsuperscript{***}-0.08}  \\
			\hdashline
			\textbf{Avg.} & \cellcolor{heatlow!27}{\textbf{-20}}              & \cellcolor{heatlow!37}{\textbf{-27}}              & \cellcolor{heatlow!26}{\textbf{-0.07}}              & \cellcolor{heatlow!33}{\textbf{-0.09}}              \\
			\hline
			\multicolumn{5}{c}{\textbf{Abstract}}                                                                                                                                                                                             \\
			\hline
			V1            & \cellcolor{heathigh!85}{\textsuperscript{***}+44} & \cellcolor{heathigh!48}{\textsuperscript{***}+25} & \cellcolor{heathigh!85}{\textsuperscript{***}+0.10} & \cellcolor{heathigh!51}{\textsuperscript{***}+0.06} \\
			V2            & \cellcolor{heathigh!85}{\textsuperscript{***}+23} & \cellcolor{white}{-0}                             & \cellcolor{heathigh!85}{\textsuperscript{***}+0.04} & \cellcolor{white}{-0.00}                            \\
			V3            & \cellcolor{heathigh!85}{\textsuperscript{***}+19} & \cellcolor{heathigh!31}{+7}                       & \cellcolor{heathigh!85}{\textsuperscript{***}+0.03} & \cellcolor{heathigh!28}{+0.01}                      \\
			V4            & \cellcolor{heathigh!85}{\textsuperscript{***}+22} & \cellcolor{heathigh!8}{+2}                        & \cellcolor{heathigh!85}{\textsuperscript{***}+0.04} & \cellcolor{white}{-0.00}                            \\
			V5            & \cellcolor{heathigh!85}{\textsuperscript{***}+84} & \cellcolor{heathigh!79}{\textsuperscript{***}+78} & \cellcolor{heathigh!85}{\textsuperscript{***}+0.35} & \cellcolor{heathigh!75}{\textsuperscript{***}+0.31} \\
			\hdashline
			\textbf{Avg.} & \cellcolor{heathigh!85}{\textbf{+38}}             & \cellcolor{heathigh!49}{\textbf{+22}}             & \cellcolor{heathigh!85}{\textbf{+0.11}}             & \cellcolor{heathigh!54}{\textbf{+0.07}}             \\
			\hline
			\multicolumn{5}{c}{\textbf{Concrete}}                                                                                                                                                                                             \\
			\hline
			V1            & \cellcolor{heatlow!81}{\textsuperscript{***}-64}  & \cellcolor{heatlow!85}{\textsuperscript{***}-67}  & \cellcolor{heatlow!85}{\textsuperscript{***}-0.24}  & \cellcolor{heatlow!81}{\textsuperscript{***}-0.23}  \\
			V2            & \cellcolor{heatlow!85}{\textsuperscript{***}-62}  & \cellcolor{heatlow!85}{\textsuperscript{***}-62}  & \cellcolor{heatlow!85}{\textsuperscript{***}-0.26}  & \cellcolor{heatlow!78}{\textsuperscript{***}-0.24}  \\
			V3            & \cellcolor{heatlow!85}{\textsuperscript{***}-46}  & \cellcolor{heatlow!85}{\textsuperscript{***}-46}  & \cellcolor{heatlow!85}{\textsuperscript{***}-0.13}  & \cellcolor{heatlow!85}{\textsuperscript{***}-0.13}  \\
			V4            & \cellcolor{heatlow!85}{\textsuperscript{***}-50}  & \cellcolor{heatlow!83}{\textsuperscript{***}-49}  & \cellcolor{heatlow!85}{\textsuperscript{***}-0.15}  & \cellcolor{heatlow!79}{\textsuperscript{***}-0.14}  \\
			V5            & \cellcolor{heatlow!82}{\textsuperscript{***}-81}  & \cellcolor{heatlow!85}{\textsuperscript{***}-84}  & \cellcolor{heatlow!79}{\textsuperscript{***}-0.38}  & \cellcolor{heatlow!85}{\textsuperscript{***}-0.41}  \\
			\hdashline
			\textbf{Avg.} & \cellcolor{heatlow!84}{\textbf{-61}}              & \cellcolor{heatlow!85}{\textbf{-62}}              & \cellcolor{heatlow!85}{\textbf{-0.23}}              & \cellcolor{heatlow!85}{\textbf{-0.23}}              \\
			\hline
		\end{tabular}
		\caption*{MT40k (Concreteness)}
	\end{minipage}
	\hfill
	\begin{minipage}[t]{0.49\textwidth}
		\centering
		\setlength{\tabcolsep}{3pt}
		\renewcommand{\arraystretch}{0.95}
		\begin{tabular}{l rr rr}
			\hline
			              & \multicolumn{2}{c}{Win margin}                    & \multicolumn{2}{c}{MAE reduction}                                                                                                                             \\
			Variant       & \realhead{ImageNet}                               & \realhead{Wikimedia}                              & \realhead{ImageNet}                                 & \realhead{Wikimedia}                                \\
			\hline
			\multicolumn{5}{c}{\textbf{All}}                                                                                                                                                                                                  \\
			\hline
			V1            & \cellcolor{heathigh!49}{\textsuperscript{***}+40} & \cellcolor{heathigh!39}{\textsuperscript{***}+32} & \cellcolor{heathigh!44}{\textsuperscript{***}+0.19} & \cellcolor{heathigh!32}{\textsuperscript{***}+0.14} \\
			V2            & \cellcolor{white}{+0}                             & \cellcolor{heatlow!45}{\textsuperscript{***}-9}   & \cellcolor{heatlow!17}{\textsuperscript{**}-0.02}   & \cellcolor{heatlow!42}{\textsuperscript{***}-0.05}  \\
			V3            & \cellcolor{heatlow!23}{-4}                        & \cellcolor{heatlow!51}{\textsuperscript{***}-9}   & \cellcolor{heatlow!28}{-0.01}                       & \cellcolor{heatlow!57}{\textsuperscript{**}-0.02}   \\
			V4            & \cellcolor{heatlow!55}{\textsuperscript{***}-28}  & \cellcolor{heatlow!55}{\textsuperscript{***}-28}  & \cellcolor{heatlow!47}{\textsuperscript{***}-0.11}  & \cellcolor{heatlow!51}{\textsuperscript{***}-0.12}  \\
			V5            & \cellcolor{heathigh!52}{\textsuperscript{***}+44} & \cellcolor{heathigh!45}{\textsuperscript{***}+38} & \cellcolor{heathigh!48}{\textsuperscript{***}+0.22} & \cellcolor{heathigh!39}{\textsuperscript{***}+0.18} \\
			\hdashline
			\textbf{Avg.} & \cellcolor{heathigh!45}{\textbf{+11}}             & \cellcolor{heathigh!20}{\textbf{+5}}              & \cellcolor{heathigh!39}{\textbf{+0.05}}             & \cellcolor{heathigh!23}{\textbf{+0.03}}             \\
			\hline
			\multicolumn{5}{c}{\textbf{Low imagery}}                                                                                                                                                                                          \\
			\hline
			V1            & \cellcolor{heathigh!85}{\textsuperscript{***}+70} & \cellcolor{heathigh!75}{\textsuperscript{***}+62} & \cellcolor{heathigh!85}{\textsuperscript{***}+0.37} & \cellcolor{heathigh!67}{\textsuperscript{***}+0.29} \\
			V2            & \cellcolor{heatlow!5}{-1}                         & \cellcolor{heatlow!85}{\textsuperscript{***}-17}  & \cellcolor{heatlow!42}{\textsuperscript{***}-0.05}  & \cellcolor{heatlow!85}{\textsuperscript{***}-0.10}  \\
			V3            & \cellcolor{heathigh!85}{+1}                       & \cellcolor{heatlow!11}{-2}                        & \cellcolor{white}{+0.00}                            & \cellcolor{heatlow!28}{-0.01}                       \\
			V4            & \cellcolor{heatlow!77}{\textsuperscript{***}-39}  & \cellcolor{heatlow!85}{\textsuperscript{***}-43}  & \cellcolor{heatlow!72}{\textsuperscript{***}-0.17}  & \cellcolor{heatlow!85}{\textsuperscript{***}-0.20}  \\
			V5            & \cellcolor{heathigh!85}{\textsuperscript{***}+72} & \cellcolor{heathigh!83}{\textsuperscript{***}+70} & \cellcolor{heathigh!85}{\textsuperscript{***}+0.39} & \cellcolor{heathigh!76}{\textsuperscript{***}+0.35} \\
			\hdashline
			\textbf{Avg.} & \cellcolor{heathigh!85}{\textbf{+21}}             & \cellcolor{heathigh!57}{\textbf{+14}}             & \cellcolor{heathigh!85}{\textbf{+0.11}}             & \cellcolor{heathigh!46}{\textbf{+0.06}}             \\
			\hline
			\multicolumn{5}{c}{\textbf{High imagery}}                                                                                                                                                                                         \\
			\hline
			V1            & \cellcolor{heathigh!15}{\textsuperscript{***}+12} & \cellcolor{heathigh!4}{+3}                        & \cellcolor{heathigh!7}{\textsuperscript{*}+0.03}    & \cellcolor{white}{-0.00}                            \\
			V2            & \cellcolor{heathigh!85}{+2}                       & \cellcolor{heatlow!5}{-1}                         & \cellcolor{white}{+0.00}                            & \cellcolor{heatlow!8}{-0.01}                        \\
			V3            & \cellcolor{heatlow!51}{\textsuperscript{**}-9}    & \cellcolor{heatlow!85}{\textsuperscript{***}-15}  & \cellcolor{heatlow!85}{\textsuperscript{**}-0.03}   & \cellcolor{heatlow!85}{\textsuperscript{***}-0.03}  \\
			V4            & \cellcolor{heatlow!32}{\textsuperscript{***}-16}  & \cellcolor{heatlow!26}{\textsuperscript{***}-13}  & \cellcolor{heatlow!21}{\textsuperscript{***}-0.05}  & \cellcolor{heatlow!17}{\textsuperscript{***}-0.04}  \\
			V5            & \cellcolor{heathigh!19}{\textsuperscript{***}+16} & \cellcolor{heathigh!7}{+6}                        & \cellcolor{heathigh!11}{\textsuperscript{***}+0.05} & \cellcolor{heathigh!7}{\textsuperscript{*}+0.03}    \\
			\hdashline
			\textbf{Avg.} & \cellcolor{heathigh!4}{\textbf{+1}}               & \cellcolor{heatlow!85}{\textbf{-4}}               & \cellcolor{white}{\textbf{-0.00}}                   & \cellcolor{heatlow!85}{\textbf{-0.01}}              \\
			\hline
		\end{tabular}
		\caption*{CP2004B (Imagery)}
	\end{minipage}
	\caption{Non-tied win margin and MAE reduction for Qwen2.5-VL mitigation prompt variants on concreteness (left) and imagery (right).
		\realctxtext{\emph{ImageNet}/\emph{Wikimedia}} are real-image contexts.
		Green/purple cells mark positive/negative changes, scaled within each variant and metric.
		Superscripts mark significance versus standard prompting after Benjamini--Hochberg correction, as in Figure~\ref{fig:mitigation}.}
	\label{tab:appendix_mitigation_prompt_variants_qwen_vl}
\end{table*}

\end{document}